\documentclass[runningheads]{llncs}
\usepackage{graphicx}
\usepackage{amsmath,amssymb} % define this before the line numbering.
\usepackage{color}
\usepackage[width=122mm,left=12mm,paperwidth=146mm,height=193mm,top=12mm,paperheight=217mm]{geometry}
\usepackage{epsfig}
\usepackage{graphicx}
\usepackage{breqn}
\usepackage{algorithm,algorithmic}
\usepackage{paralist, tabularx}
\usepackage{csquotes}
\usepackage{caption}
\usepackage{booktabs}       % professional-quality tables
\usepackage{amsfonts}       % blackboard math symbols
\usepackage{nicefrac}       % compact symbols for 1/2, etc.
\usepackage{microtype}      % microtypography
\usepackage{float}
\usepackage{afterpage}
\usepackage{float}
\newcommand{\reffig}[1]{Figure~\ref{#1}}
\newcommand{\argmin}{\arg\!\min}
\newcommand{\etal}{et al.}
\usepackage[pagebackref=true,breaklinks=true,colorlinks,citecolor=blue,linkcolor=blue,bookmarks=false]{hyperref}

\begin{document}
\pagestyle{headings}
\mainmatter

\title{A Closed-form Solution to \\Photorealistic Image Stylization} 
% Replace with your title

\titlerunning{A Closed-form Solution to Photorealistic Image Stylization}
% Replace with a meaningful short version of your title

\authorrunning{Y. Li, M.-Y. Liu, X. Li, M.-H. Yang, J. Kautz}
% Replace with shorter version of the author list. If there are more authors than fits a line, please use A. Author et al.

\author{Yijun Li$^1$, Ming-Yu Liu$^2$, Xueting Li$^1$, Ming-Hsuan Yang$^{1,2}$, Jan Kautz$^2$}

\institute{$^1$University of California, Merced~~~$^2$NVIDIA\\
	\email{\{yli62,xli75,mhyang\}@ucmerced.edu}~~~~\email{\{mingyul,jkautz\}@nvidia.com}
	\\
%\url{https://github.com/NVIDIA/FastPhotoStyle}
}

\maketitle

\begin{abstract}
Photorealistic image stylization concerns transferring style of a reference photo to a content photo with the constraint that the stylized photo should remain photorealistic. While several photorealistic image stylization methods exist, they tend to generate spatially inconsistent stylizations with noticeable artifacts. In this paper, we propose a method to address these issues. The proposed method consists of a stylization step and a smoothing step. While the stylization step transfers the style of the reference photo to the content photo, the smoothing step ensures spatially consistent stylizations. Each of the steps has a closed-form solution and can be computed efficiently. We conduct extensive experimental validations. The results show that the proposed method generates photorealistic stylization outputs that are more preferred by human subjects as compared to those by the competing methods while running much faster. Source code and additional results are available at \url{https://github.com/NVIDIA/FastPhotoStyle}.

\end{abstract}

\section{Introduction}

\begin{figure}[t]
\centering
	\begin{tabular}{ccccc}
    \includegraphics[height = .14\linewidth, width = .192\linewidth]{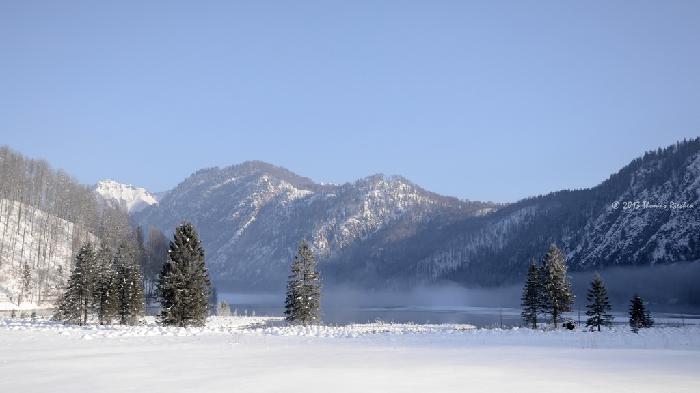} 
    \includegraphics[height = .14\linewidth, width = .192\linewidth]{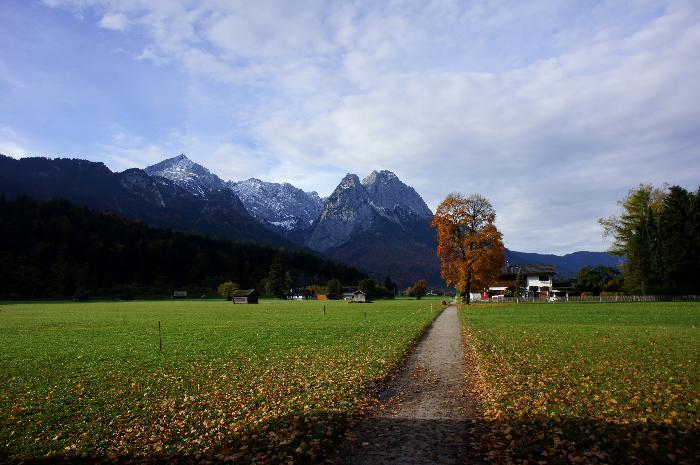} 
    \includegraphics[height = .14\linewidth, width = .192\linewidth]{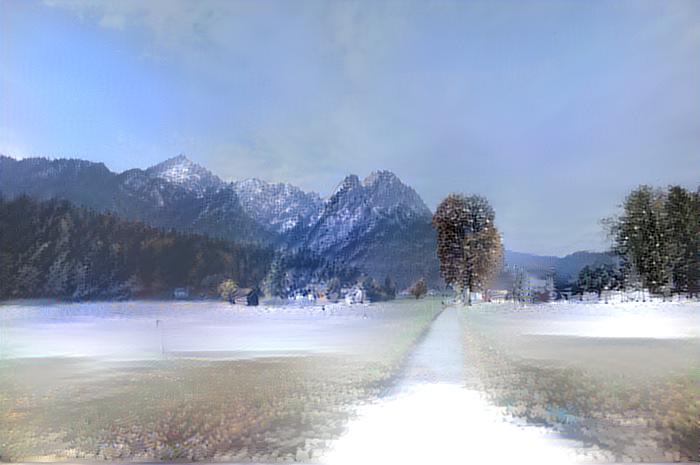} 
    \includegraphics[height = .14\linewidth, width = .192\linewidth]{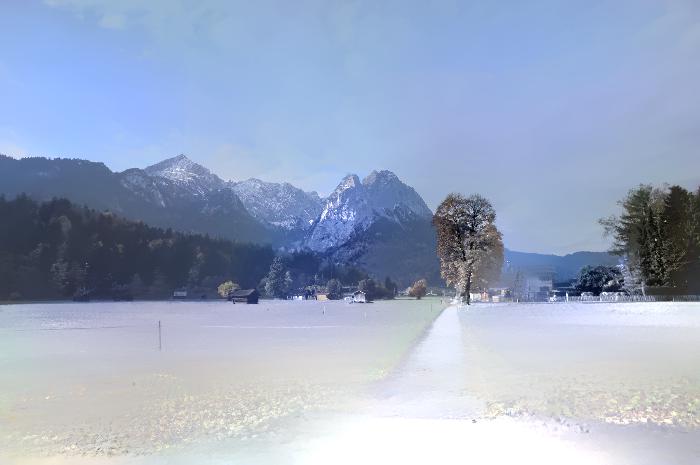} 
    \includegraphics[height = .14\linewidth, width = .192\linewidth]{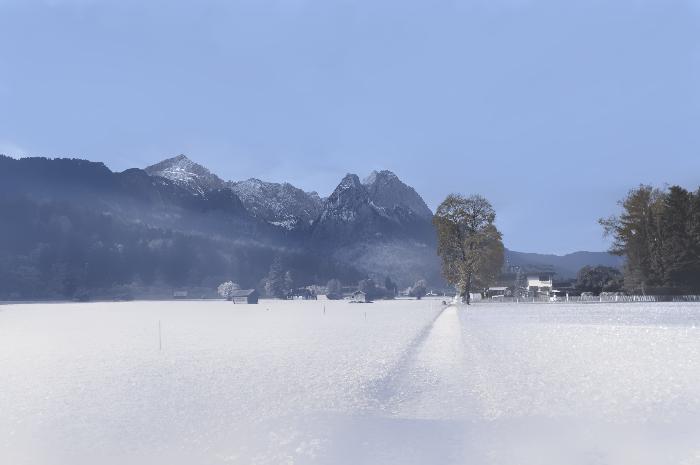}  \\
    \includegraphics[height = .14\linewidth, width = .192\linewidth]{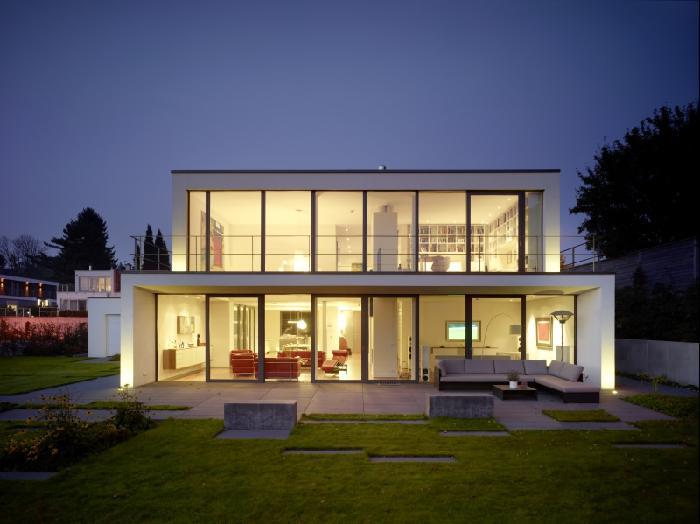} 
    \includegraphics[height = .14\linewidth, width = .192\linewidth]{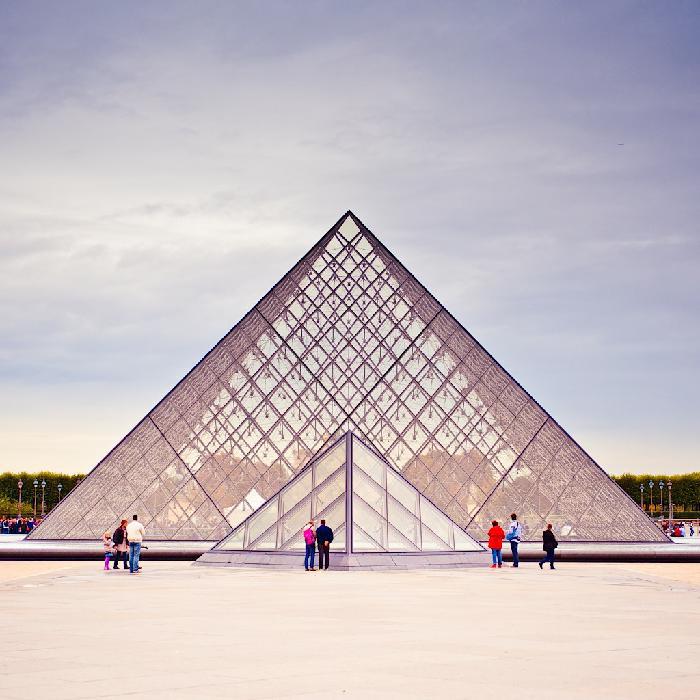} 
    \includegraphics[height = .14\linewidth, width = .192\linewidth]{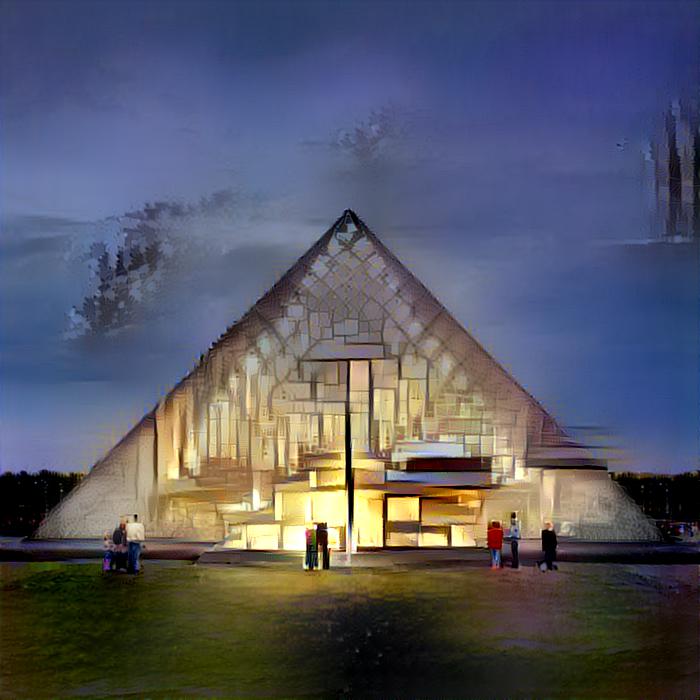} 
    \includegraphics[height = .14\linewidth, width = .192\linewidth]{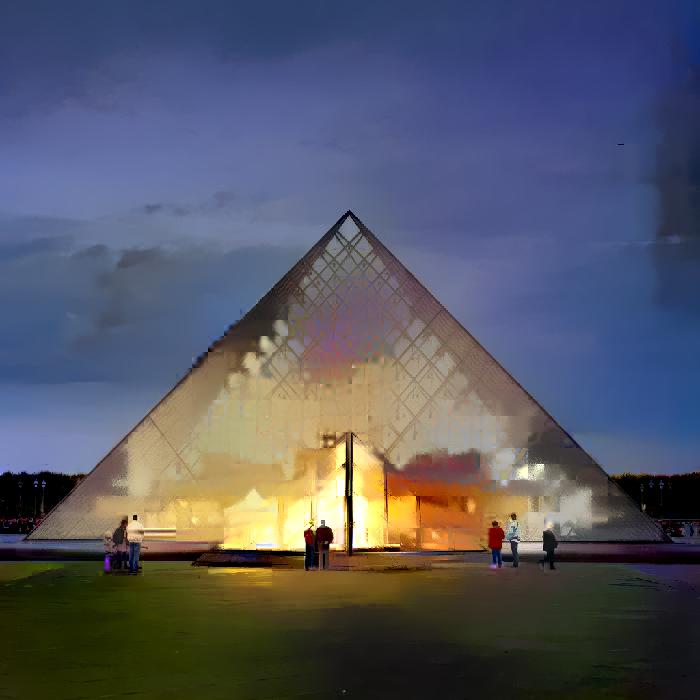} 
    \includegraphics[height = .14\linewidth, width = .192\linewidth]{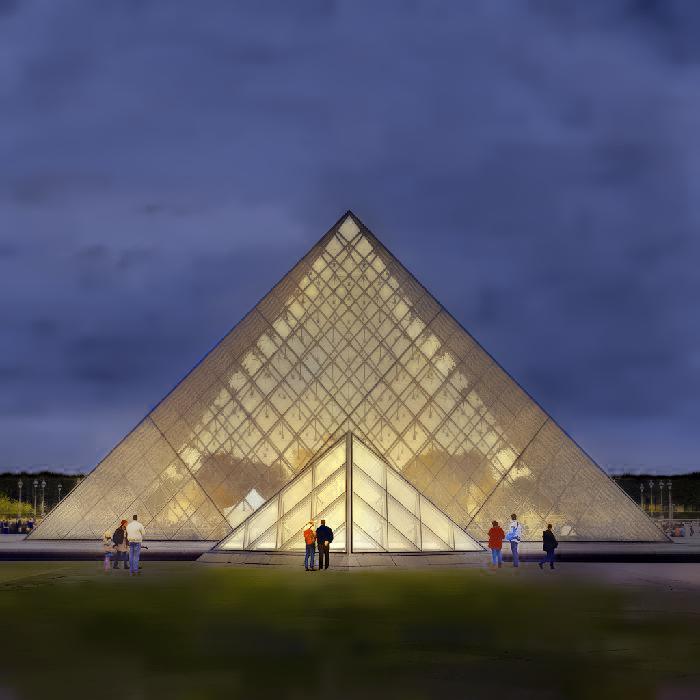}  
    \end{tabular}
	\begin{tabular}{ccccc}
    { ~~~(a)~Style~~~~~~~~}& {\hspace{-2mm} ~~(b) Content~~~~~~ } & { \hspace{-6mm}(c) Gatys \etal~\cite{GatysTransfer-CVPR2016} }& { \hspace{-2mm} (d) Luan \etal~\cite{Luan-2017-photorealism}}& {~~(e) Ours~~~~~}
    \end{tabular}
    \captionof{figure}{Given a style photo (a) and a content photo (b), photorealistic image stylization aims at transferring style of the style photo to the content photo as shown in (c), (d) and (e). Comparing with existing  methods~\cite{GatysTransfer-CVPR2016,Luan-2017-photorealism}, the output photos computed by our method are stylized more consistently and with fewer artifacts. Moreover, our method runs an order of magnitude faster.
    }
    \label{fig:teaser}
\end{figure}

Photorealistic image stylization aims at changing style of a photo to that of a reference photo. For a faithful stylization, content of the photo should remain the same. Furthermore, the output photo should look like a real photo as it were captured by a camera. \reffig{fig:teaser} shows two photorealistic image stylization examples. In one example, we transfer a summery photo to a snowy one, while in the other, we transfer a day-time photo to a night-time photo. %Photorealistic image stylization algorithms find use in numerous applications, ranging from image editing to content creation.

Classical photorealistic stylization methods are mostly based on color/tone matching~\cite{reinhard-2001color,Pitie-2005,sunkavalli2010multi,bae-2006two} and are often limited to specific scenarios (e.g., seasons~\cite{laffont-2014-transient} and headshot portraits~\cite{shih2-2014style}). Recently, Gatys \etal~\cite{GatysTexture-NIPS2015,GatysTransfer-CVPR2016} show that the correlations between deep features encode the visual style of an image and propose an optimization-based method, the neural style transfer algorithm, for image stylization. While the method shows impressive performance for \emph{artistic} stylization (converting images to paintings), it often introduces structural artifacts and distortions when applied to photorealistic image stylization as shown in \reffig{fig:teaser}(c). In a follow-up work, Luan \etal~\cite{Luan-2017-photorealism} propose adding a regularization term to the optimization objective function of the neural style transfer algorithm for avoiding distortions in the stylization output. However, this often results in inconsistent stylizations in semantically uniform regions as shown in \reffig{fig:teaser}(d). To address the issues, we propose a photorealistic image stylization method.

Our method consists of a stylization step and a smoothing step. Both have a closed-form solution\footnote{A closed-form solution means that the solution can be obtained in a fixed finite number of operations, including convolutions, max-pooling, whitening, etc.} and can be computed efficiently. The stylization step is based on the whitening and coloring transform (WCT)~\cite{WCT-2017-NIPS}, which stylizes images via feature projections. The WCT was designed for \emph{artistic} stylization. Similar to the neural style transfer algorithm, it suffers from structural artifacts when applied to photorealistic image stylization. Our WCT-based stylization step resolves the issue by utilizing a novel network design for feature transform. The WCT-based stylization step alone may generate spatially inconsistent stylizations. We resolve this issue by the proposed smoothing step, which is based on a manifold ranking algorithm. We conduct extensive experimental validation with comparison to the state-of-the-art methods. User study results show that our method generates outputs with better stylization effects and fewer artifacts.

\section{Related Work}

Existing stylization methods can be classified into two categories: global and local. Global methods~\cite{reinhard-2001color,Pitie-2005,freedman2010object} achieve stylization through matching the means and variances of pixel colors~\cite{reinhard-2001color} or their histograms~\cite{Pitie-2005}. Local methods~\cite{shih1-2013data,shih2-2014style,Wu-2013-content,laffont-2014-transient,tsai2016sky} stylize images through finding dense correspondences between the content and style photos based on either low-level or high-level features. These approaches are slow in practice. Also, they are often developed for specific scenarios (e.g., day-time or season change).

Gatys \etal~\cite{GatysTexture-NIPS2015,GatysTransfer-CVPR2016} propose the neural style transfer algorithm for \emph{artistic} stylization. The major step in the algorithm is to solve an optimization problem of matching the Gram matrices of deep features extracted from the content and style photos. A number of methods have been developed~\cite{MrfTransfer-CVPR2016,Texturenet-ICML2016,Perceptual-ECCV2016,Me-2017-diversified,chen2017stylebank,GoogleMultiTexture-2016,Ghiasi-2017-BMVC,Huang-2017-arbitrary,WCT-2017-NIPS,MSRA-2017-visual} to further improve its stylization performance and speed. However, these methods do not aim for preserving photorealism (see \reffig{fig:teaser}(c)). Post-processing techniques~\cite{Lapstyle-ACMMM,MatchGradientSPE-2017-BMVC} have been proposed to refine these results by matching the gradients between the input and output photos.

Photorealistic image stylization is related to the image-to-image translation problem~\cite{isola2016image,wang2017high,liu2016coupled,taigman2016unsupervised,shrivastava2016learning,liu2016unsupervised,zhu2017unpaired,MUNIT} where the goal is to learn to translate an image from one domain to another. However, photorealistic image stylization does not require a training dataset of content and style images for learning the translation function. Photorealistic image stylization can be considered as a special kind of image-to-image translation. Not only can it be used to translate a photo to a different domain (e.g., form day to night-time) but also transfer style (e.g., extent of darkness) of a specific reference image to the content image.

Closest to our work is the method of Luan \etal~\cite{Luan-2017-photorealism}. It improves photorealism of stylization outputs computed by the neural style transfer algorithm~\cite{GatysTexture-NIPS2015,GatysTransfer-CVPR2016} by incorporating a new loss term to the optimization objective, which has the effect of better preserving local structures in the content photo. However, it often generates inconsistent stylization with noticeable artifacts (Figure~\ref{fig:teaser}(d)). Moreover, the method is computationally expensive. Our proposed algorithm aims at efficient and effective photorealistic image stylization. We demonstrate that it performs favorably against Luan \etal~\cite{Luan-2017-photorealism} \mbox{in terms of both quality and speed.}

\begin{figure}[t]
	\centering
	\includegraphics[width=0.92\textwidth]{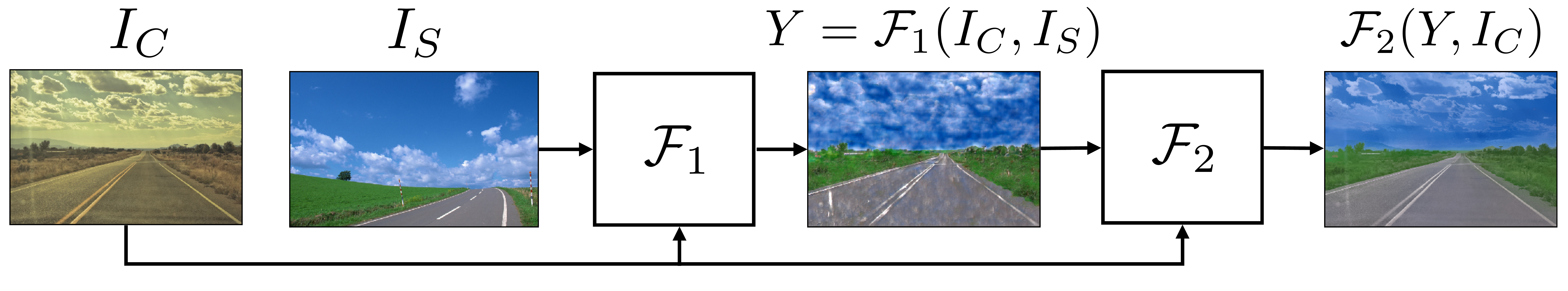}
	\caption{Our photorealistic image stylization method consists of two closed-form steps: $\mathcal{F}_1$ and $\mathcal{F}_2$. While $\mathcal{F}_1$ maps $I_C$ to an intermediate image carrying the style of $I_S$, $\mathcal{F}_2$ removes noticeable artifacts, which produces a photorealistic output.}
	\label{fig:closed-form}
\end{figure}

\section{Photorealistic Image Stylization}

Our photorealistic image stylization algorithm consists of two steps as illustrated in \reffig{fig:closed-form}. The first step is a stylization transform $\mathcal{F}_1$ called PhotoWCT. Given a style photo $I_S$, $\mathcal{F}_1$ transfer the style of $I_S$ to the content photo $I_C$ while minimizing structural artifacts in the output image. Although $\mathcal{F}_1$ can faithfully stylize $I_C$, it often generates inconsistent stylizations in semantically similar regions. Therefore, we use a photorealistic smoothing function $\mathcal{F}_2$, to eliminate these artifacts. \mbox{Our whole algorithm can be written as a two-step mapping function:}
\begin{equation}
    \mathcal{F}_2\Big{(}\mathcal{F}_1\big{(}I_C,I_S\big{)}, I_C\Big{)},
\end{equation}
In the following, we discuss the stylization and smoothing steps in details.

\subsection{Stylization}

The PhotoWCT is based on the WCT~\cite{WCT-2017-NIPS}. It utilizes a novel network design for achieving photorealistic image stylization. We briefly review the WCT below.

\begin{figure}[t]
	\centering
	\begin{tabular}{c@{\hspace{0.005\linewidth}}c@{\hspace{0.005\linewidth}}c}
		\includegraphics[width = .99\linewidth]{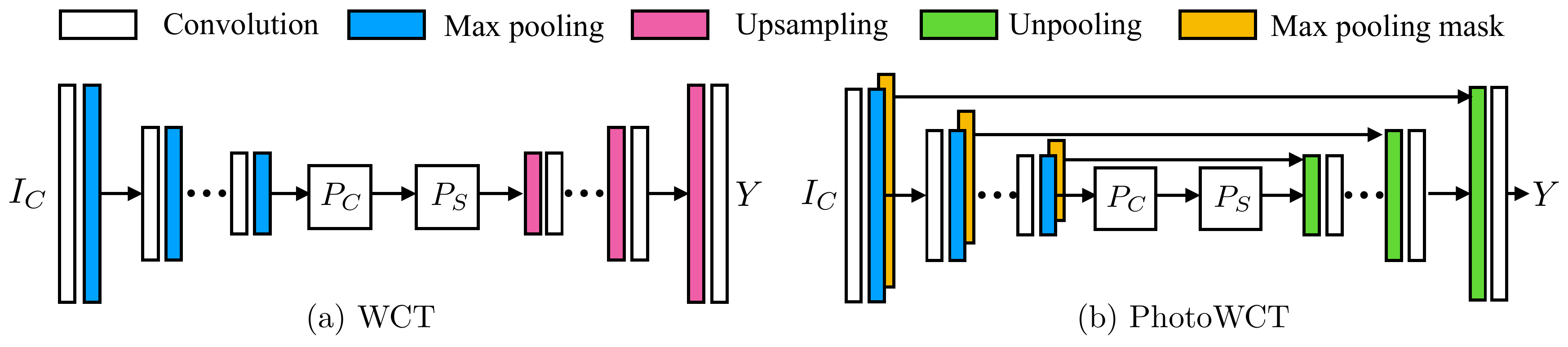} & \\
	\end{tabular}
	\caption{The PhotoWCT and WCT share the same encoder architecture and projection steps. In the PhotoWCT, we replace the upsampling layers (pink) with unpooling layers (green). Note that the unpooling layer is used together with the pooling mask (yellow) which records \emph{where} carries the \emph{maximum} over each max pooling region in the corresponding pooling layer~\cite{SWWAE-2016-ICLR}.}
	\label{fig:network_comparison}
\end{figure}

\paragraph{\bf WCT.} The WCT~\cite{WCT-2017-NIPS} formulates stylization as an image reconstruction problem with feature projections. To utilize WCT, an auto-encoder for general image reconstruction is first trained. Specifically, it uses the VGG-19 model~\cite{VGG-2014} as the encoder $\mathcal{E}$ (weights are kept fixed) and trains a decoder $\mathcal{D}$ for reconstructing the input image. The decoder is symmetrical to the encoder and uses upsampling layers (pink blocks in Figure~\ref{fig:network_comparison}(a)) to enlarge the spatial resolutions of the feature maps. Once the auto-encoder is trained, a pair of projection functions are inserted at the network bottleneck to perform stylization through the whitening ($P_C$) and coloring ($P_S$) transforms. The key idea behind the WCT is to directly match feature correlations of the content image to those of the style image via the two projections. Specifically, given a pair of content image $I_{C}$ and style image $I_{S}$, the WCT first extracts their vectorised VGG features $H_{C}=\mathcal{E}(I_C)$ and $H_{S}=\mathcal{E}(I_S)$, and then transform the content feature $H_C$ via
\begin{equation}\label{formula_whiten}
H_{CS} = P_S P_C H_C,
\end{equation}
where $P_C=E_{C}\Lambda^{-\frac{1}{2}}_{C}E^{\top}_{C}$, and  $P_S=E_{S}\Lambda^{\frac{1}{2}}_{S}E^{\top}_{S}$. Here $\Lambda_{C}$ and $\Lambda_{S}$ are the diagonal matrices with the eigenvalues of the covariance matrix $H_{C} H^{\top}_{C}$ and $H_{S} H_{S}^{\top}$ respectively. The matrices $E_{C}$ and $E_{S}$ are the corresponding orthonormal matrices of the eigenvectors, respectively. After the transformation, the correlations of transformed features match those of the style features, i.e., $H_{CS} H_{CS}^{\top}=H_{S} H^{\top}_{S}$. Finally, the stylized image is obtained by directly feeding the transformed feature map into the decoder: $Y = \mathcal{D}(H_{CS})$. For better stylization performance, Li \etal~\cite{WCT-2017-NIPS} use a multi-level stylization strategy, which performs the WCT on the VGG features at different layers.

\begin{figure}[t]
\centering
\begin{tabular}{c@{\hspace{0.005\linewidth}}c@{\hspace{0.005\linewidth}}c}
\includegraphics[height = .236\linewidth, width = .48\linewidth]{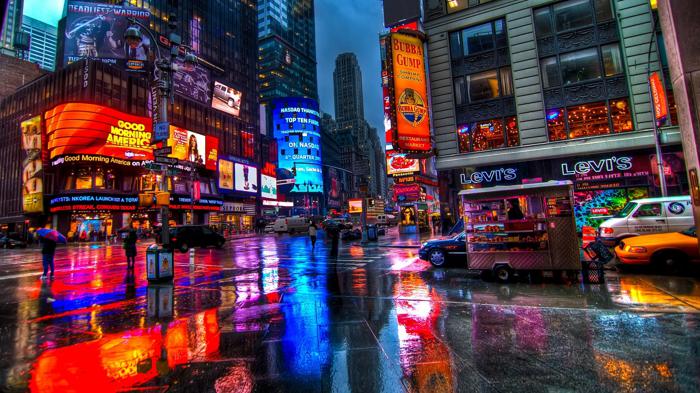} & 
\includegraphics[height = .24\linewidth, width = .48\linewidth]{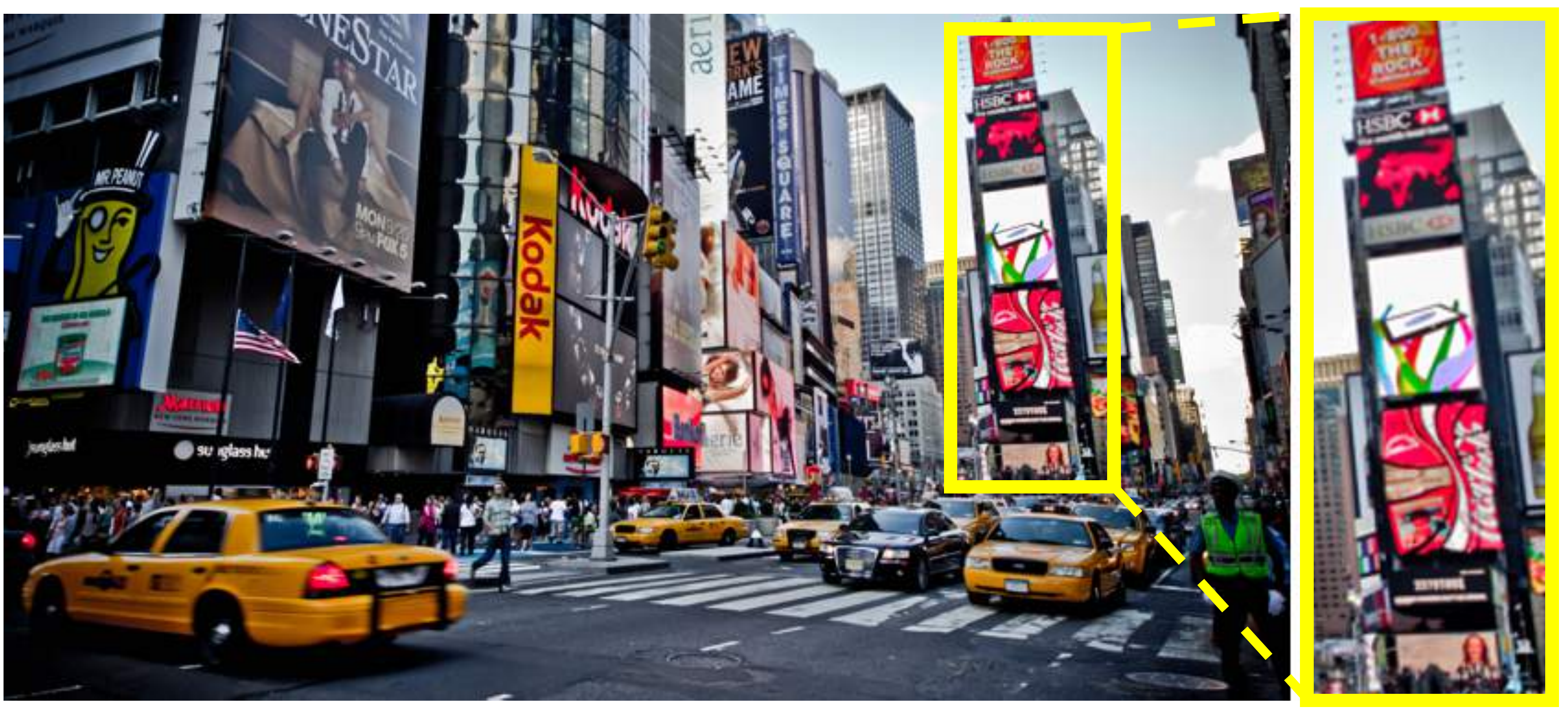} &\\
{(a) Style } & {(b) Content } & \\
\includegraphics[height = .24\linewidth, width = .48\linewidth]{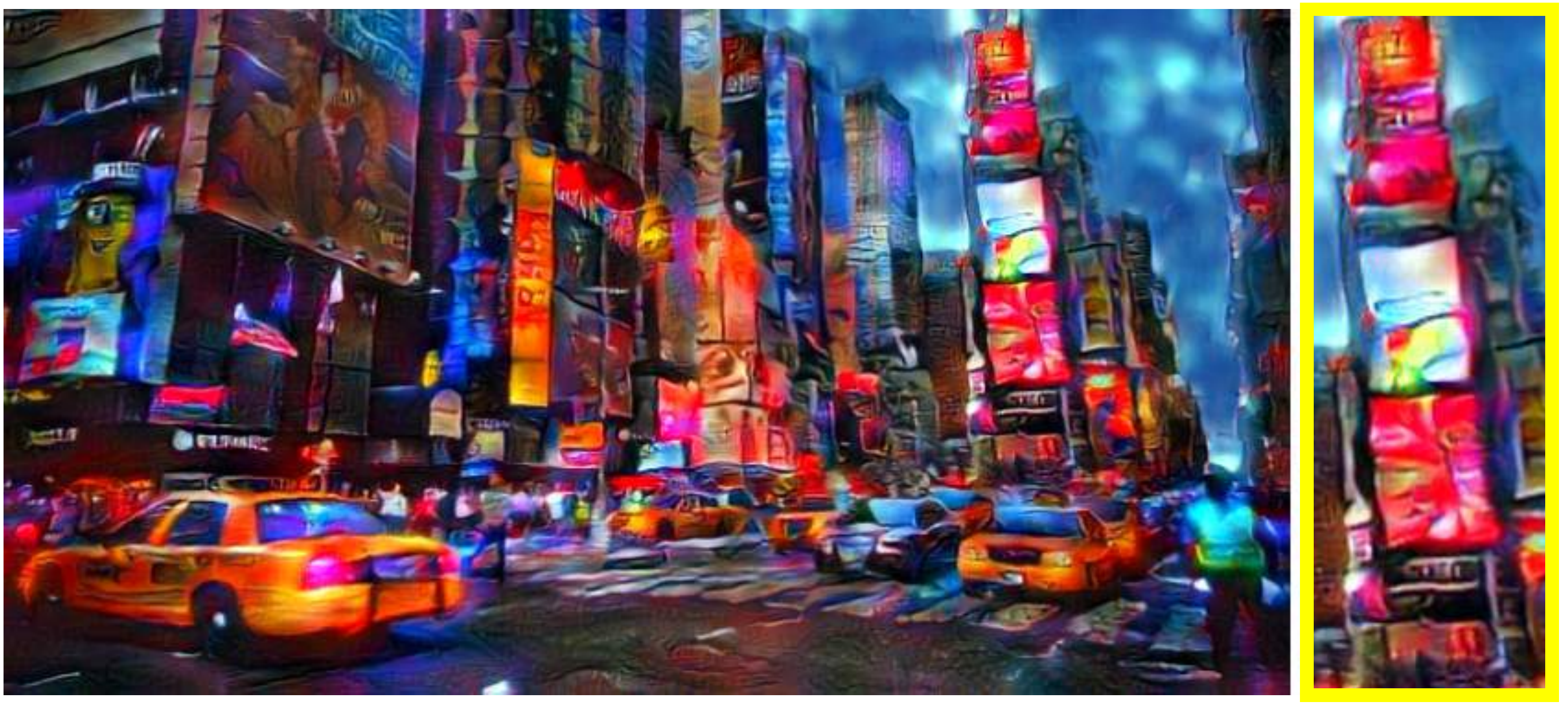} &
\includegraphics[height = .24\linewidth, width = .48\linewidth]{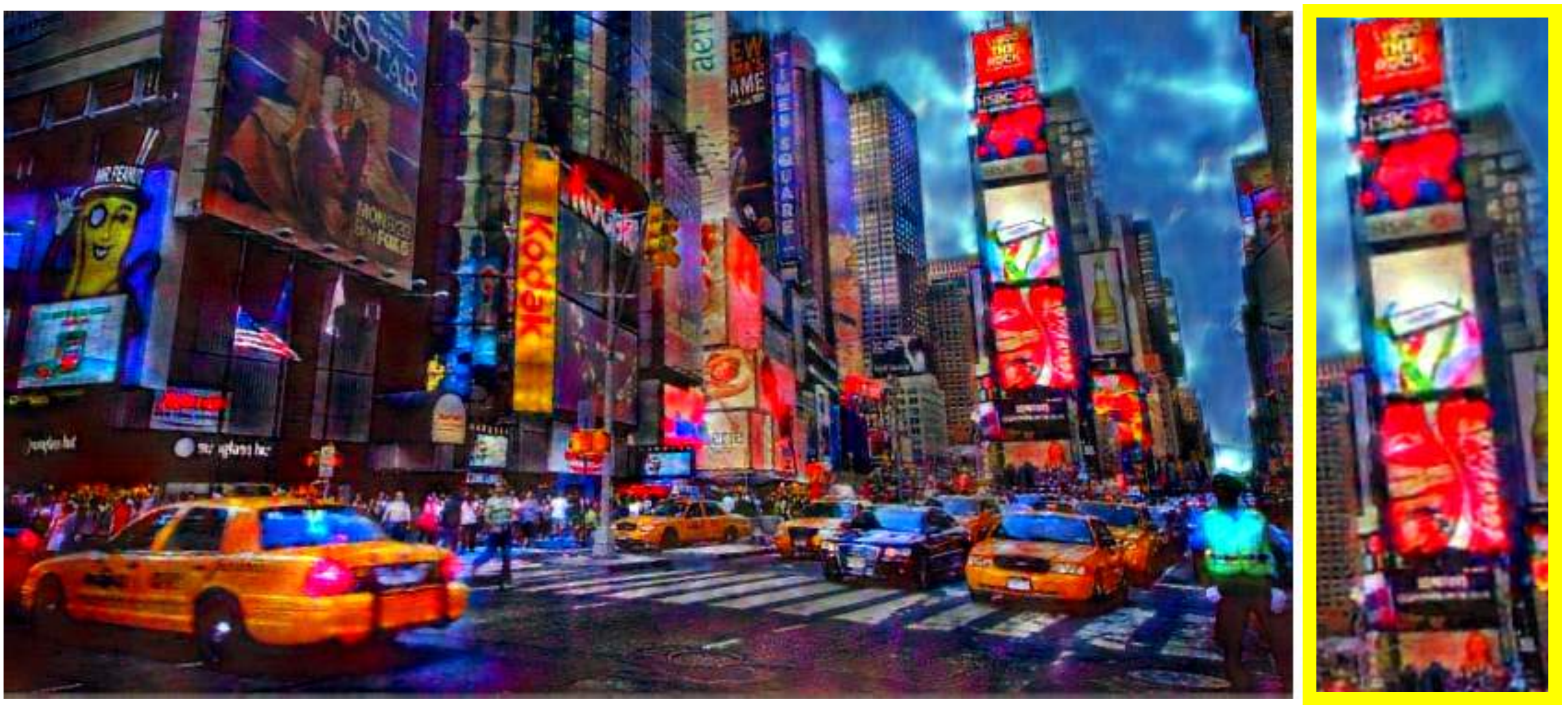} & \\
{(c) WCT~\cite{WCT-2017-NIPS} } & {(d) PhotoWCT } & \\
\includegraphics[height = .24\linewidth, width = .48\linewidth]{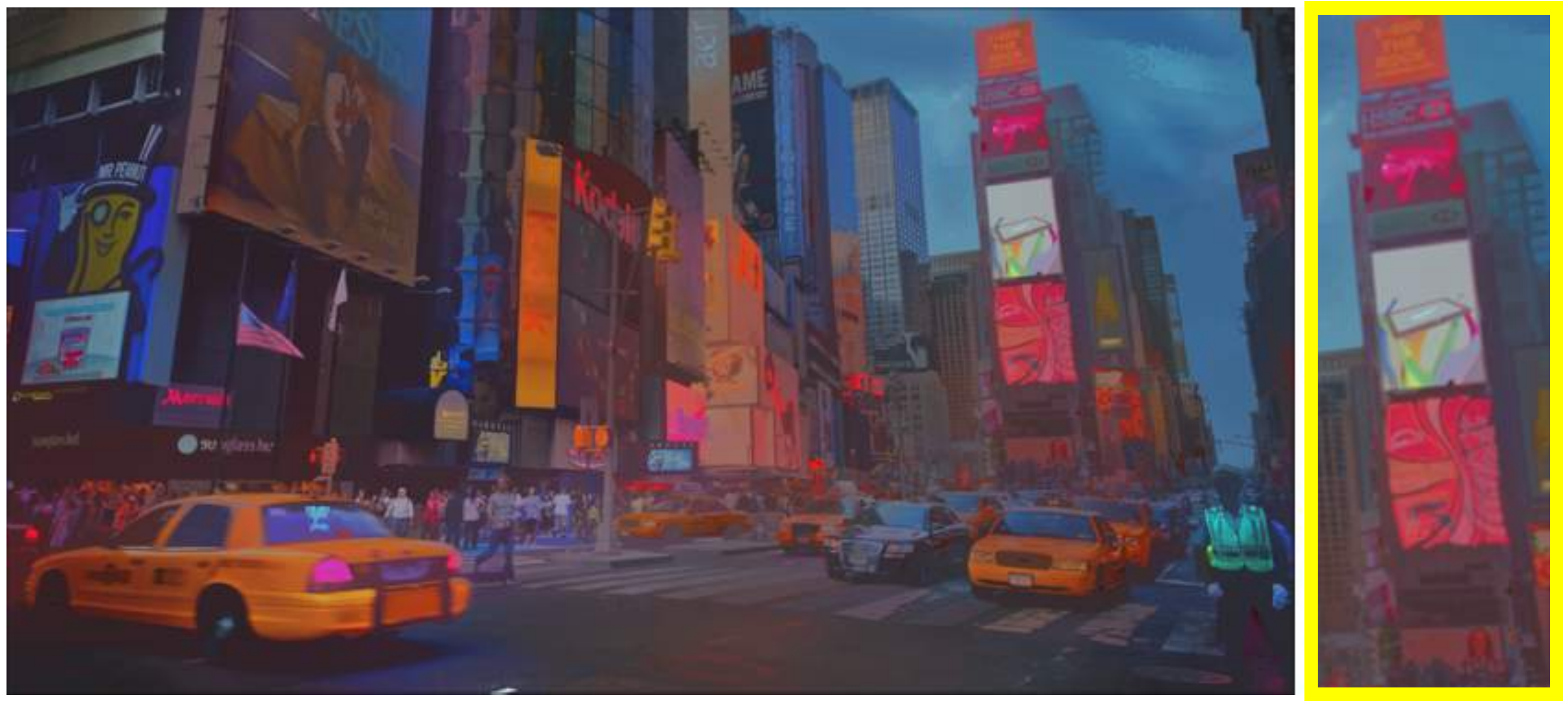} &
\includegraphics[height = .24\linewidth, width = .48\linewidth]{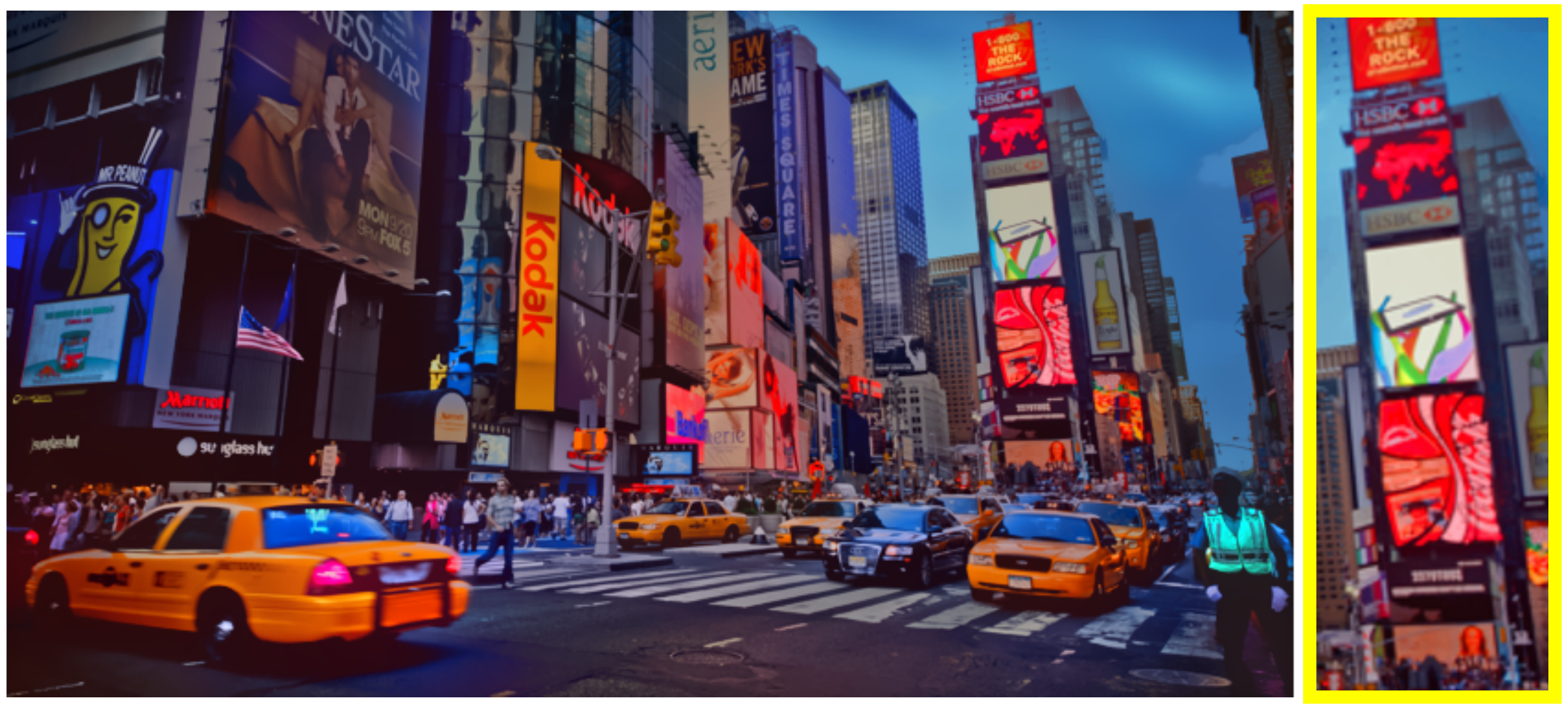} & \\
{(e) WCT + smoothing } & {(f) PhotoWCT + smoothing } & 
\end{tabular}
\caption{The stylization output generated by the PhotoWCT better preserves local structures in the content images, which is important for the image smoothing step as shown in (e) and (f).}
\label{fig:upsample_vs_unpool}
\end{figure}

The WCT performs well for artistic image stylization. However it generates structural artifacts (e.g., distortions on object boundaries) for photorealistic image stylization (Figure~\ref{fig:upsample_vs_unpool}(c)). The proposed PhotoWCT is designed to suppress these structural artifacts.

\paragraph{\bf PhotoWCT.} Our PhotoWCT design is motivated by the observation that the max-pooling operation in the WCT reduces spatial information in feature maps. Simply upsampling feature maps in the decoder fails to recover detailed structures of the input image. That is, we need to pass the lost spatial information to the decoder to facilitate reconstructing these fine details. Inspired by the success of the unpooling layer~\cite{SWWAE-2016-ICLR,zeiler-2014-visualizing,noh-2015-learning} in preserving spatial information, the PhotoWCT replaces the upsampling layers in the WCT with unpooling layers. The PhotoWCT function is formulated as
\begin{equation}
    Y=\mathcal{F}_1(I_C,I_S)=\overline{\mathcal{D}}(P_S P_C H_C),
\end{equation}
where $\overline{\mathcal{D}}$ is the decoder, which contains unpooling layers and is trained for image reconstruction. Figure~\ref{fig:network_comparison} illustrates the network architecture difference between the WCT and the proposed PhotoWCT.

\begin{figure}[t]
\centering
\begin{tabular}{c@{\hspace{0.005\linewidth}}c@{\hspace{0.005\linewidth}}c@{\hspace{0.005\linewidth}}c@{\hspace{0.005\linewidth}}c@{\hspace{0.005\linewidth}}c@{\hspace{0.005\linewidth}}c}
\includegraphics[height = .232\linewidth, width = .32\linewidth]{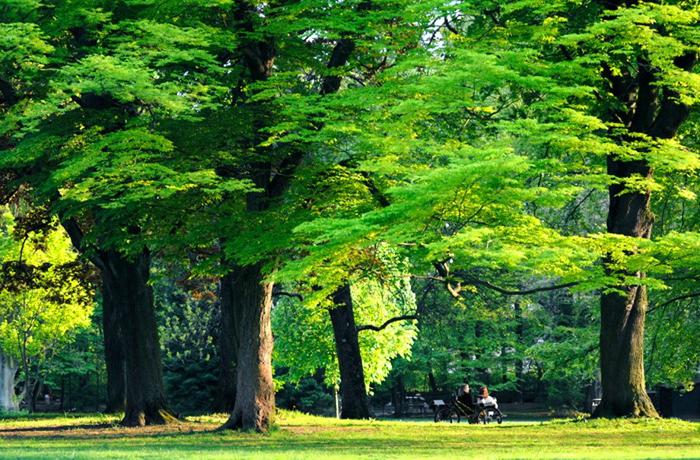} &
\includegraphics[height = .232\linewidth, width = .32\linewidth]{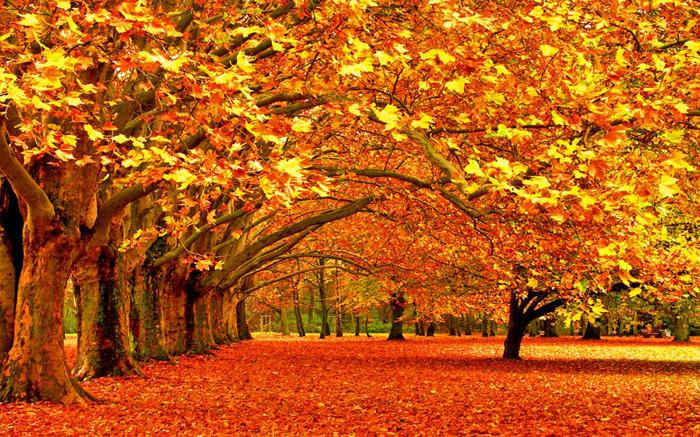} &
\includegraphics[height = .232\linewidth, width = .32\linewidth]{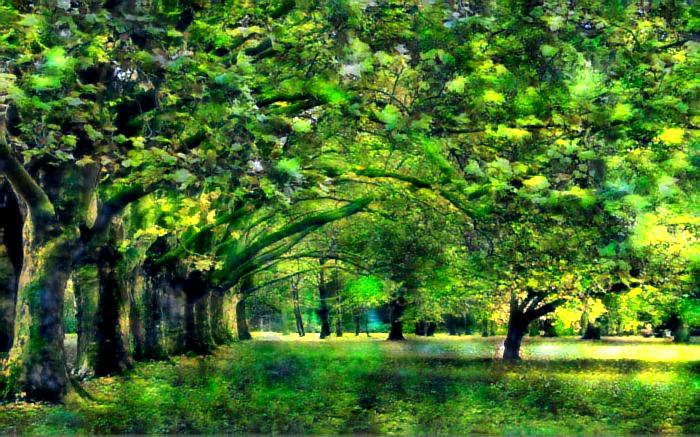} & \\
{(a) Style } & {(b) Content } & {(c) PhotoWCT (Ours) } \\
\includegraphics[height = .232\linewidth, width = .32\linewidth]{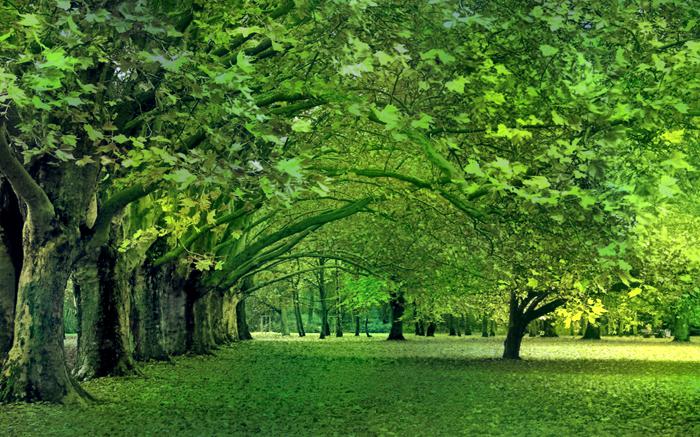} &
\includegraphics[height = .232\linewidth, width = .32\linewidth]{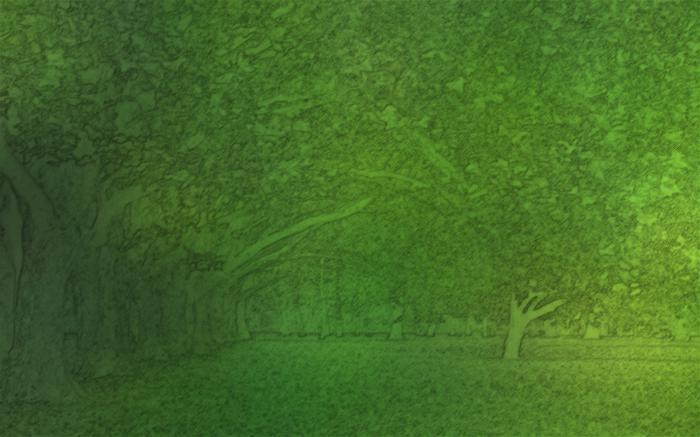} &
\includegraphics[height = .232\linewidth, width = .32\linewidth]{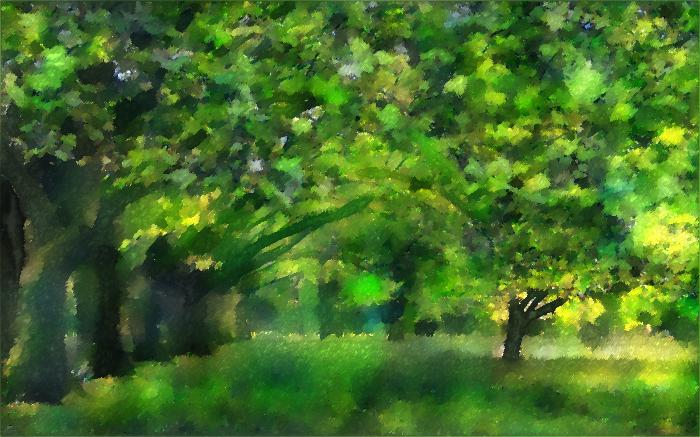} & \\
{(d) MattingAff} & {(e) GaussianAff $\sigma=1$} & {(f) GaussianAff $\sigma=0.1$}
\end{tabular}
\caption{
Smoothing with different affinities. To refine the PhotoWCT result in (c), it is hard to find an optimal $\sigma$ for the Gaussian Affinity that performs globally well as shown in (e)-(f). In contrast, using the Matting Affinity can simultaneously smooth different regions well as shown in (d).
}
\label{fig:affinity}
\end{figure}

Figure~\ref{fig:upsample_vs_unpool}(c) and (d) compare the stylization results of the WCT and PhotoWCT. As highlighted in close-ups, the straight lines along the building boundary in the content image becomes zigzagged in the WCT stylization result but remains straight in the PhotoWCT result. The PhotoWCT-stylized image has much fewer structural artifacts.
We also perform a user study in the experiment section to quantitatively verify that the PhotoWCT generally leads to better stylization effects than the WCT.

\subsection{Photorealistic Smoothing}\label{ref:smoothing}

The PhotoWCT-stylized result (Figure~\ref{fig:upsample_vs_unpool}(d)) still looks less like a photo since semantically similar regions are often stylized inconsistently. As shown in Figure~\ref{fig:upsample_vs_unpool}, when applying the PhotoWCT to stylize the day-time photo using the night-time photo, the stylized sky region would be more photorealistic if it were uniformly dark blue instead of partly dark and partly light blue. It is based on this observation, we employ the pixel affinities in the content photo to \mbox{smooth the PhotoWCT-stylized result.}

We aim to achieve two goals in the smoothing step. First, pixels with similar content in a local neighborhood should be stylized similarly. Second, the output should not deviate significantly from the PhotoWCT result in order to maintain the global stylization effects. We first represent all pixels as nodes in a graph and define an affinity matrix $W=\{w_{ij}\} \in \mathbb{R}^{N\times N}$ ($N$ is the number of pixels) to describe pixel similarities. We define a smoothness term and a fitting term that model these two goals in the following optimization problem:
\begin{equation}\label{formula_propagation}
\argmin_r \frac{1}{2} (\sum \limits_{i,j=1}^{N} w_{ij} \| \frac{r_{i}}{\sqrt{d_{ii}}} - \frac{r_{j}}{\sqrt{d_{jj}}} \|^{2} + \lambda \sum \limits_{i=1}^{N} \| r_{i} - y_{i} \|^{2}),
\end{equation}
where $y_i$ is the pixel color in the PhotoWCT-stylized result $Y$ and $r_i$ is the pixel color in the desired smoothed output $R$. The variable $d_{ii}=\sum_{j} w_{ij}$ is the diagonal element in the degree matrix $D$ of $W$, i.e., $D=\mbox{diag}\{d_{11}, d_{22}, ..., d_{NN}\}$. In~\eqref{formula_propagation}, $\lambda$ controls the balance of the two terms.

Our formulation is motivated by the graph-based ranking algorithms~\cite{zhou2004ranking,MR-2013-saliency}. In the ranking algorithms, $Y$ is a binary input where each element indicates if a specific item is a query ($y_{i}=1$ if $y_{i}$ is a query and $y_{i}=0$ otherwise). The optimal solution $R$ is the ranking values of all the items based on their pairwise affinities. In our method, we set $Y$ as the PhotoWCT-stylized result. The optimal solution $R$ is the smoothed version of $Y$ based on the pairwise pixel affinities, which encourages consistent stylization within semantically similar regions. The above optimization problem is a simple quadratic problem with a closed-form solution, which is given by
\begin{equation}\label{formula_closedform0}
R^{\ast} = (1-\alpha) (I - \alpha S)^{-1} Y,
\end{equation}
where $I$ is the identity matrix, $\alpha = \frac{1}{1+\lambda}$ and $S$ is the normalized Laplacian matrix computed from $I_{C}$, i.e., $S = D^{-\frac{1}{2}} W D^{-\frac{1}{2}} \in \mathbb{R}^{N\times N}$. As the constructed graph is often sparsely connected
(i.e., most elements in $W$ are zero), the inverse operation in (\ref{formula_closedform0}) can be computed efficiently. With the closed-form solution, the smoothing step can be written as a function mapping given by:
\begin{equation}\label{formula_closedform}
R^{\ast} = \mathcal{F}_2(Y,I_C) =(1-\alpha) (I - \alpha S)^{-1} Y.
\end{equation}

\paragraph{\bf Affinity.} The affinity matrix $W$ is computed using the content photo based on an 8-connected image graph assumption. While several choices of affinity metrics exist, a popular one is to define the affinity (denoted as GaussianAff) as
$w_{ij} = e^{-\| I_{i} - I_{j} \|^{2} / \sigma^{2}}$ where $I_{i}, I_{j}$ are the RGB values of adjacent pixels $i, j$ and $\sigma$ is a global scaling hyper-parameter~\cite{shi-2000-normalized}. However, it is difficult to determine the $\sigma$ value in practice. It often results in either over-smoothing the entire photo (Figure~\ref{fig:affinity}(e)) or stylizing the photo inconsistently (Figure~\ref{fig:affinity}(f)). To avoid selecting one global scaling hyper-parameter, we resort to the matting affinity~\cite{Levin-2008-closed,zelnik-2005-self} (denoted as MattingAff) where the affinity between two pixels is based on means and variances of pixels in a local window. Figure~\ref{fig:affinity}(d) shows that the matting affinity is able to simultaneously smooth different regions well.

\paragraph{\bf WCT plus Smoothing.} We note that the smoothing step can also remove structural artifacts in the WCT as shown in Figure~\ref{fig:upsample_vs_unpool}(e). However, it leads to unsatisfactory stylization. The main reason is that the content photo and the WCT result are severely misaligned due to spatial distortions. For example, a stylized pixel of the building in the WCT result may correspond to a pixel of the sky in the content photo. Consequently this causes wrong queries in $Y$ for the smoothing step. This shows why we need to use the PhotoWCT to remove distortions first. Figure~\ref{fig:upsample_vs_unpool}(f) shows that the combination of PhotoWCT and smoothing leads to better photorealism while still maintaining faithful stylization.

\afterpage{\clearpage}
\begin{figure}[!tbh]
	\centering
	\begin{tabular}{c@{\hspace{0.005\linewidth}}c@{\hspace{0.005\linewidth}}c@{\hspace{0.005\linewidth}}c@{\hspace{0.005\linewidth}}c@{\hspace{0.005\linewidth}}c}
		\includegraphics[height = .20\linewidth, width = .32\linewidth]{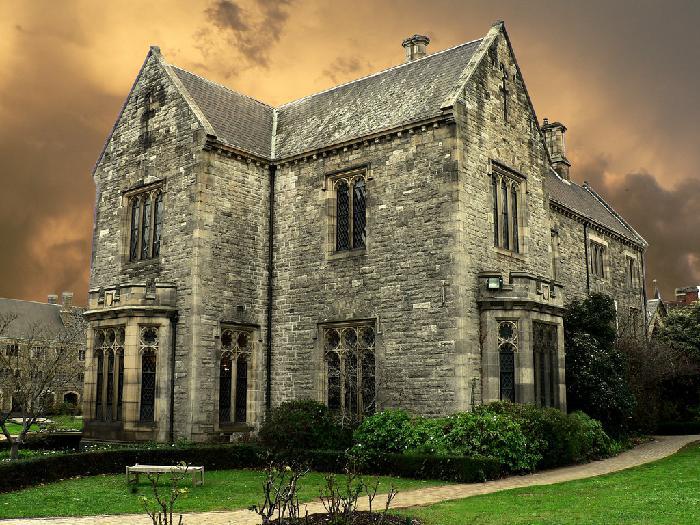} &
		\includegraphics[height = .20\linewidth, width = .32\linewidth]{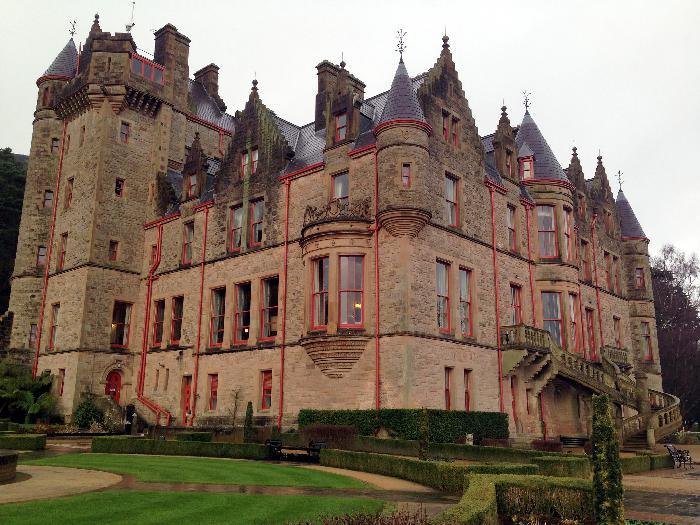} &
		\includegraphics[height = .20\linewidth, width = .32\linewidth]{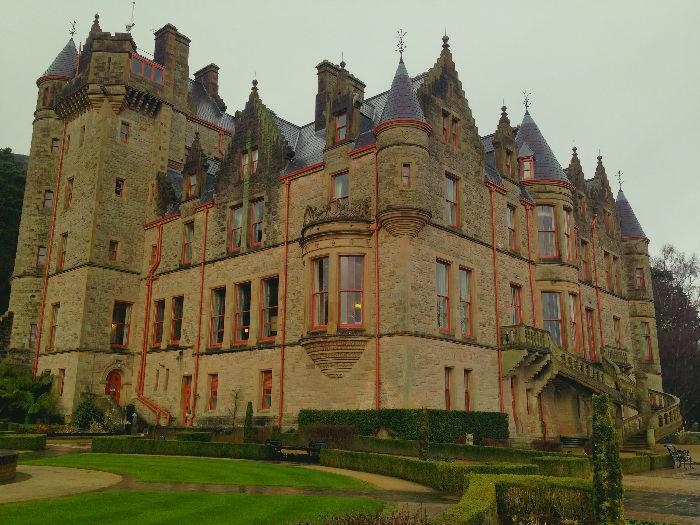} \\
		{Style }& { Content } & { Reinhard \etal~\cite{reinhard-2001color} } \\
		\includegraphics[height = .20\linewidth, width = .32\linewidth]{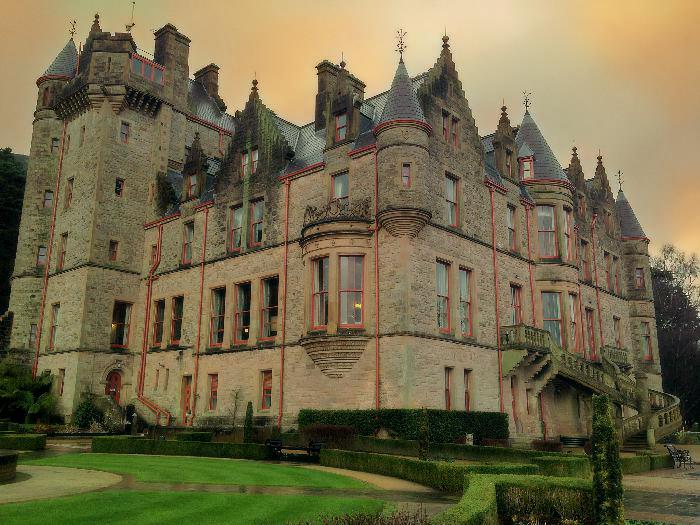} & 
		\includegraphics[height = .20\linewidth, width = .32\linewidth]{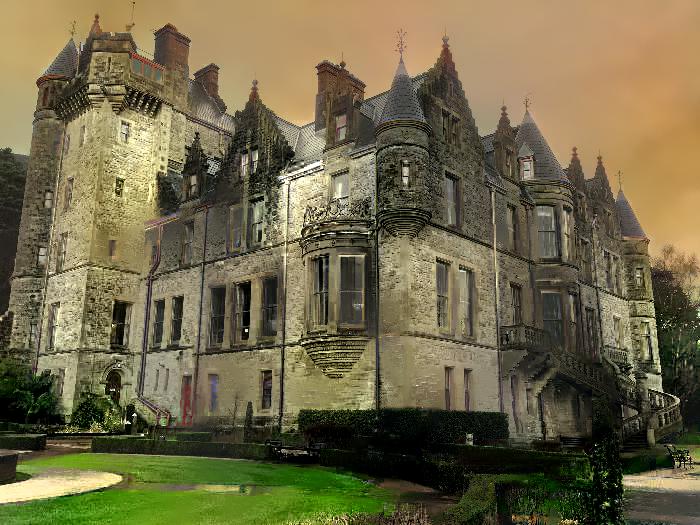} &
		\includegraphics[height = .20\linewidth, width = .32\linewidth]{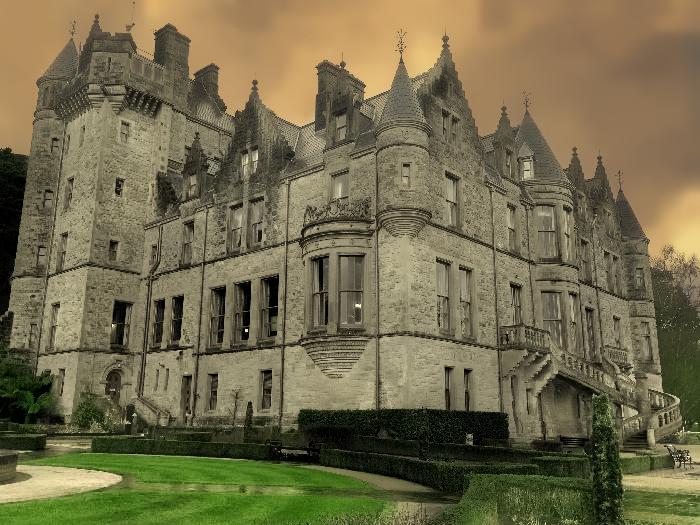} \\
		{Piti\'e \etal~\cite{Pitie-2005}} & {Luan \etal~\cite{Luan-2017-photorealism}}& {Ours }\\

		\includegraphics[height = .20\linewidth, width = .32\linewidth]{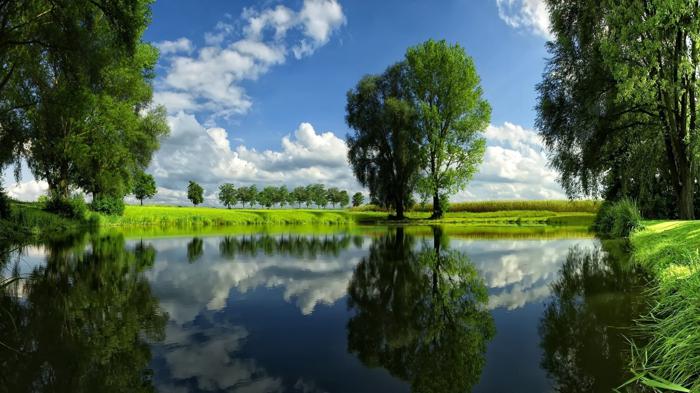} &
		\includegraphics[height = .20\linewidth, width = .32\linewidth]{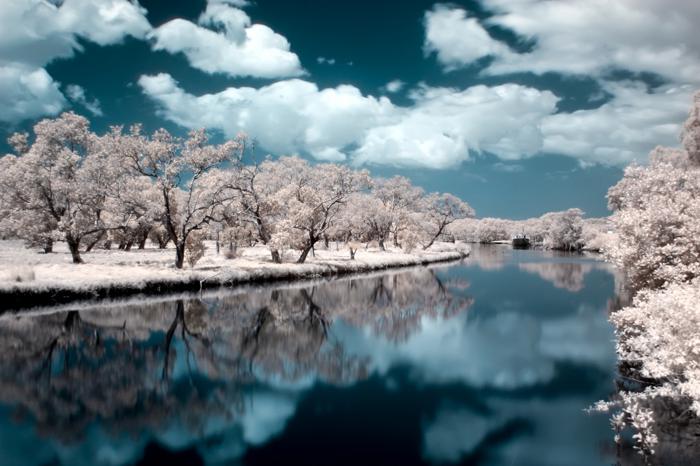} &
		\includegraphics[height = .20\linewidth, width = .32\linewidth]{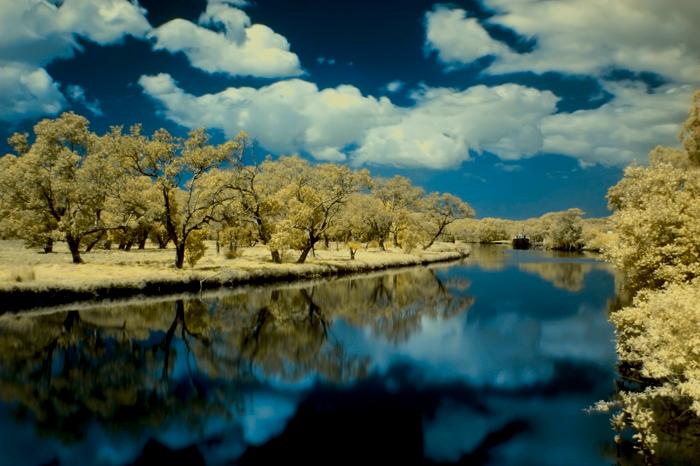} \\
		{Style }& { Content } & { Reinhard \etal~\cite{reinhard-2001color} } \\
		\includegraphics[height = .20\linewidth, width = .32\linewidth]{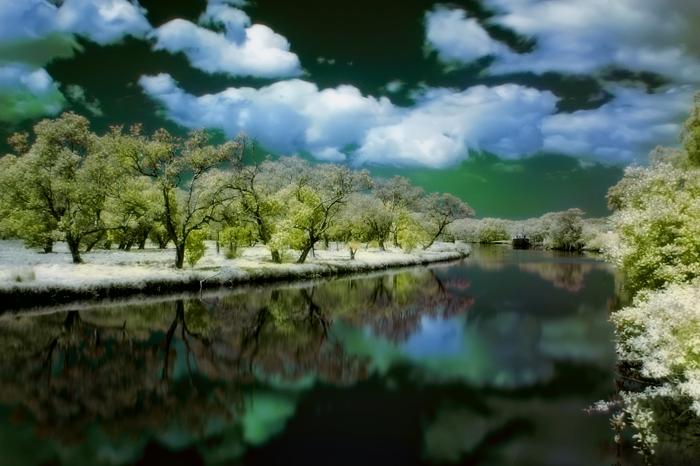} & 
		\includegraphics[height = .20\linewidth, width = .32\linewidth]{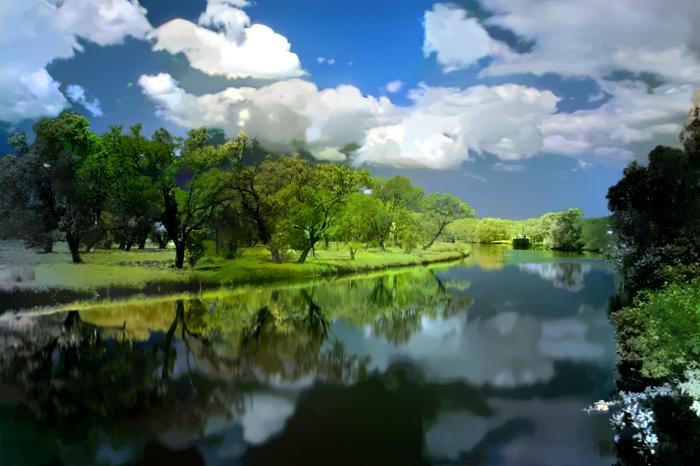} &
		\includegraphics[height = .20\linewidth, width = .32\linewidth]{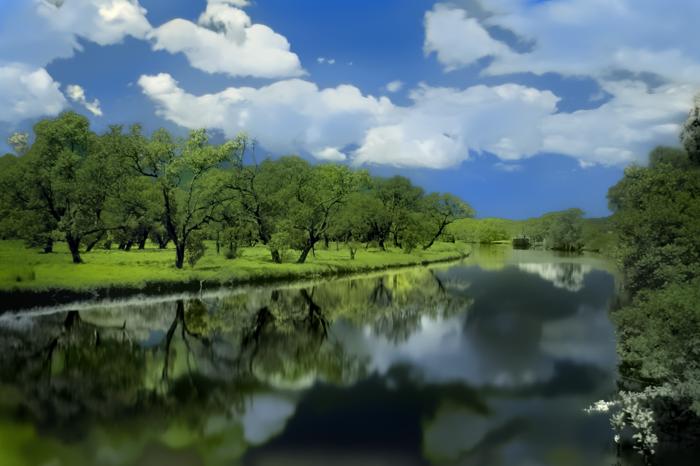} \\
		{Piti\'e \etal~\cite{Pitie-2005}} & {Luan \etal~\cite{Luan-2017-photorealism}}& {Ours }\\

		\includegraphics[height = .20\linewidth, width = .32\linewidth]{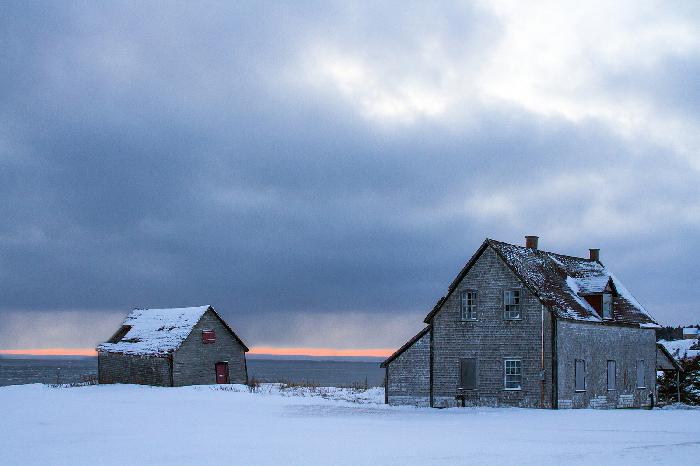} &
		\includegraphics[height = .20\linewidth, width = .32\linewidth]{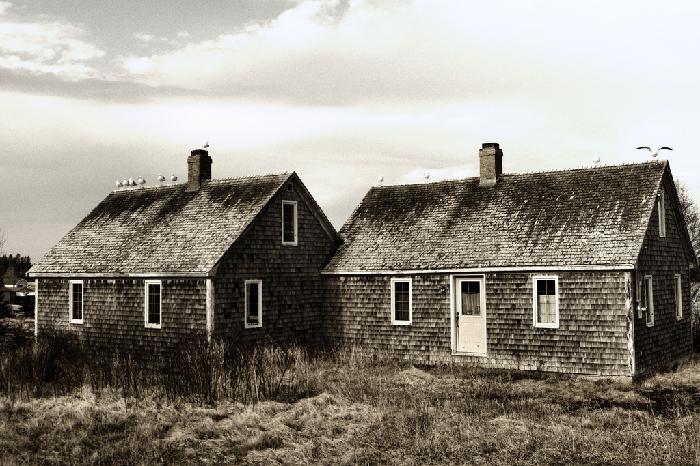} &
		\includegraphics[height = .20\linewidth, width = .32\linewidth]{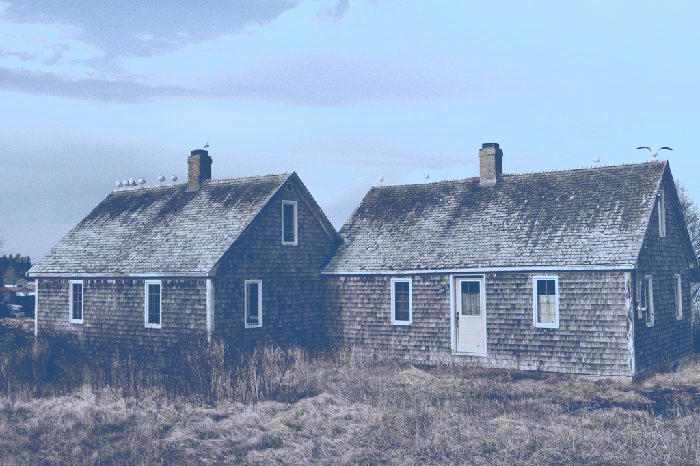} \\
		{Style }& { Content } & { Reinhard \etal~\cite{reinhard-2001color} } \\
		\includegraphics[height = .20\linewidth, width = .32\linewidth]{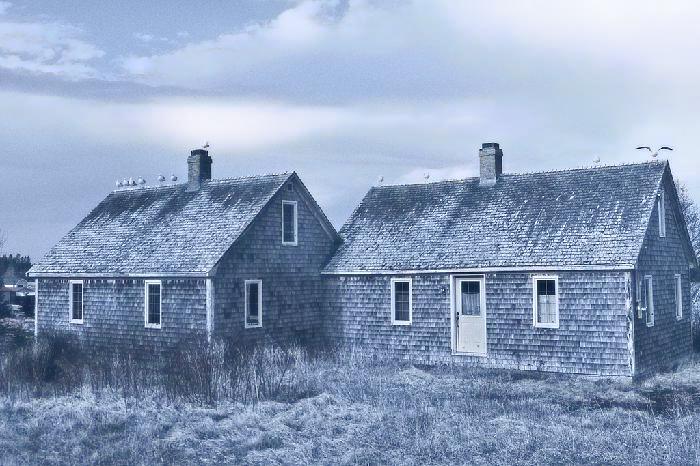} & 
		\includegraphics[height = .20\linewidth, width = .32\linewidth]{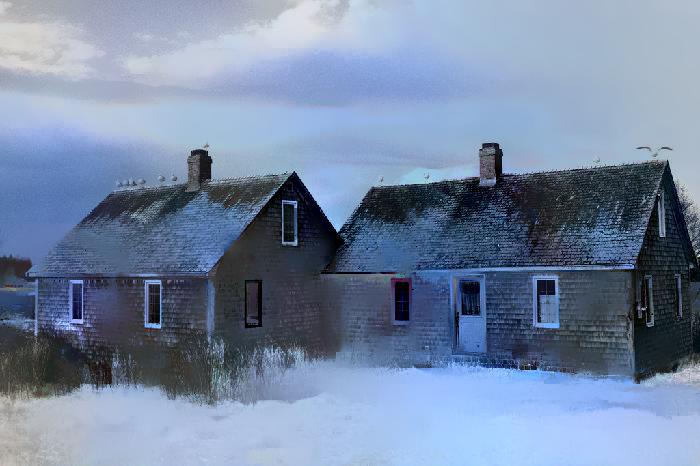} &
		\includegraphics[height = .20\linewidth, width = .32\linewidth]{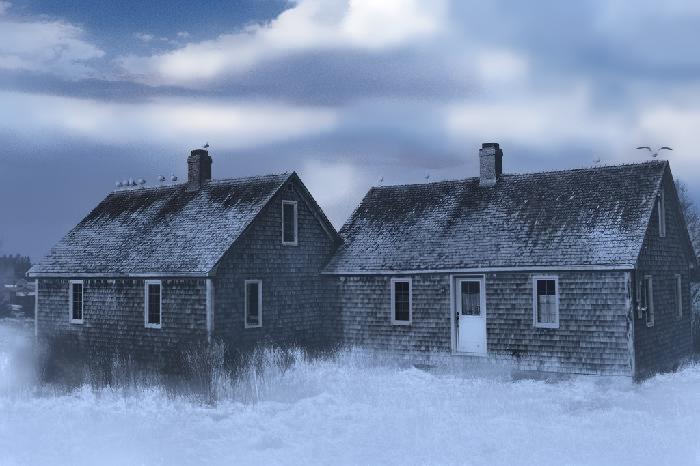} \\
		{Piti\'e \etal~\cite{Pitie-2005}} & {Luan \etal~\cite{Luan-2017-photorealism}}& {Ours }\\
		
	\end{tabular}
	\vspace{-0.5em}
	\caption{Visual comparisons with photorealistic stylization methods. In addition to color transfer, our method also synthesizes patterns in the style photos (e.g., the dark cloud in the top example, the snow at the bottom example).}
	\label{fig:compare_photorealistic}
	\vspace{-0.5em}
\end{figure}

\section{Experiments}\label{sec:experiments}

In the section, we will first discuss the implementation details. We will then present visual and user study evaluation results. Finally, we will analyze various design choices and run-time of the proposed algorithm. 
%Code and additional results are available at \url{https://github.com/NVIDIA/FastPhotoStyle}.

\paragraph{\bf Implementation details.} We use the layers from $conv1\_1$ to $conv4\_1$ in the VGG-19 network~\cite{VGG-2014} for the encoder $\mathcal{E}$. The encoder weights are given by ImageNet-pretrained weights. The decoder $\overline{\mathcal{D}}$ is the inverse of the encoder. We train the decoder by minimizing the sum of the $L_2$ reconstruction loss and perceptual loss~\cite{Perceptual-ECCV2016} using the Microsoft COCO dataset~\cite{COCO-lin2014-microsoft}. We adopt the multi-level stylization strategy proposed in the WCT~\cite{WCT-2017-NIPS} where we apply the PhotoWCT to VGG features in different layers.

Similar to the state-of-the-art methods~\cite{Gatys2016-control,Luan-2017-photorealism}, our algorithm can leverage semantic label maps for obtaining better stylization results when they are available. When performing PhotoWCT stylization, for each semantic label, we compute a pair of projection matrices $P_C$ and $P_S$ using the features from the image regions with the same label in the content and style photos, respectively. The pair is then used to stylize these image regions. With a semantic label map, content and style matching can be performed more accurately. We note that the proposed algorithm does not need precise semantic label maps for obtaining good stylization results. Finally, we also use the efficient filtering step described in Luan \etal~\cite{Luan-2017-photorealism} for post-processing.

\paragraph{\bf Visual comparison.} We compare the proposed algorithm to two categories of stylization algorithms: photorealistic and artistic. The evaluated photorealistic stylization algorithms include Reinhard \etal~\cite{reinhard-2001color}, Piti\'e \etal~\cite{Pitie-2005}, and Luan \etal~\cite{Luan-2017-photorealism}. Both Reinhard \etal~\cite{reinhard-2001color} and Piti\'e \etal~\cite{Pitie-2005} represent classical techniques that are based on color statistics matching, while Luan \etal~\cite{Luan-2017-photorealism} is based on neural style transfer~\cite{GatysTransfer-CVPR2016}. On the other hand, the set of evaluated artistic stylization algorithms include Gatys~\etal\cite{GatysTransfer-CVPR2016}, Huang~\etal\cite{Huang-2017-arbitrary}, and the WCT~\cite{WCT-2017-NIPS}. They all utilize deep networks. 

\begin{figure}[H]
	\centering
	\begin{tabular}{c@{\hspace{0.005\linewidth}}c@{\hspace{0.005\linewidth}}c@{\hspace{0.005\linewidth}}c@{\hspace{0.005\linewidth}}c@{\hspace{0.005\linewidth}}c}
		\includegraphics[height = .18\linewidth, width = .32\linewidth]{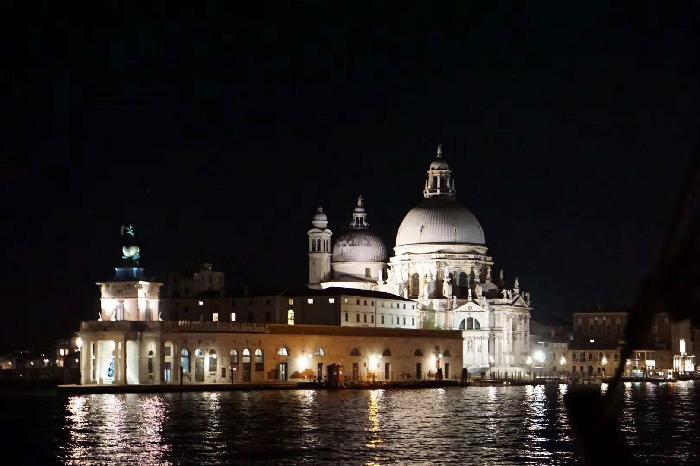} &
		\includegraphics[height = .18\linewidth, width = .32\linewidth]{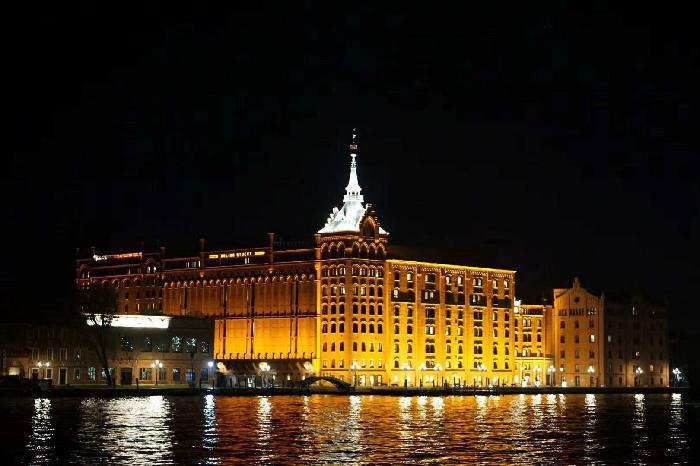} &
		\includegraphics[height = .18\linewidth, width = .32\linewidth]{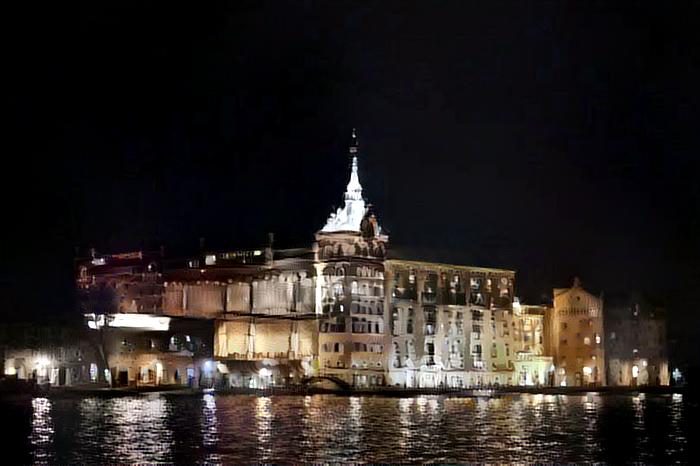} & \\
		{(a) Style }& { (b) Content } & { (c) Gatys \etal~\cite{GatysTransfer-CVPR2016} } \\
		\includegraphics[height = .18\linewidth, width = .32\linewidth]{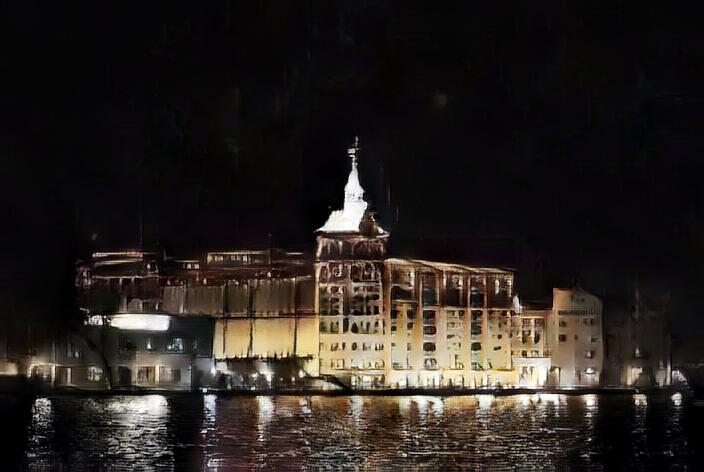} &
		\includegraphics[height = .18\linewidth, width = .32\linewidth]{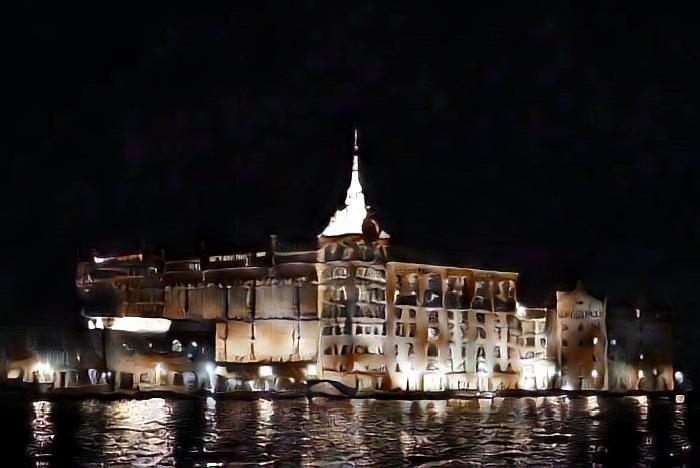} &
		\includegraphics[height = .18\linewidth, width = .32\linewidth]{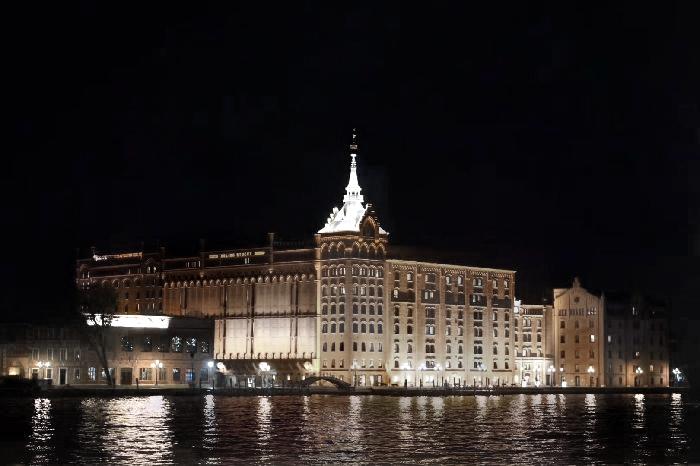} & \\
		{ (d) Huang \etal~\cite{Huang-2017-arbitrary}}& { (e) Li \etal~\cite{WCT-2017-NIPS}} & { (f) Ours }\\
		\includegraphics[height = .18\linewidth, width = .32\linewidth]{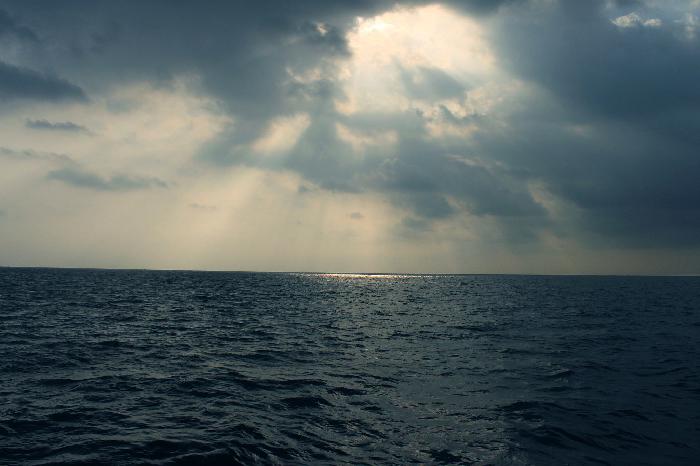} &
		\includegraphics[height = .18\linewidth, width = .32\linewidth]{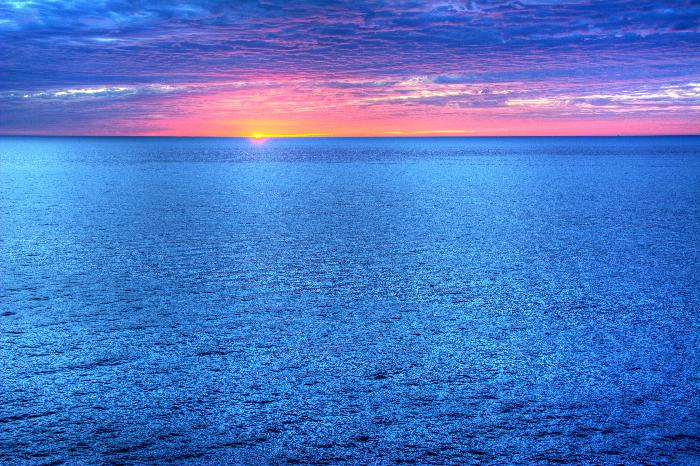} &
		\includegraphics[height = .18\linewidth, width = .32\linewidth]{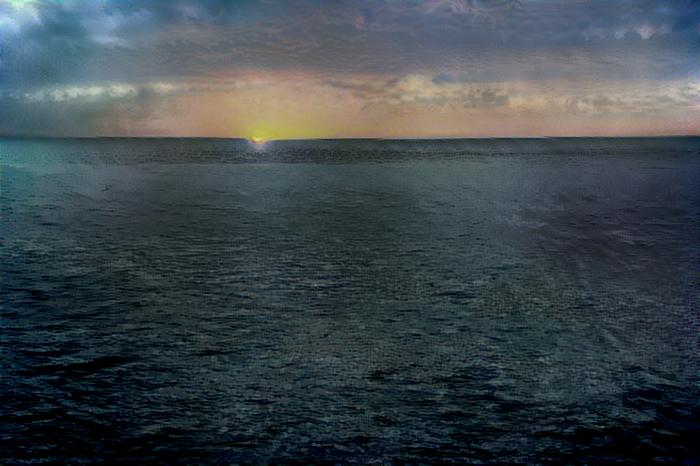} & \\
		{(a) Style }& { (b) Content } & { (c) Gatys \etal~\cite{GatysTransfer-CVPR2016} } \\
		\includegraphics[height = .18\linewidth, width = .32\linewidth]{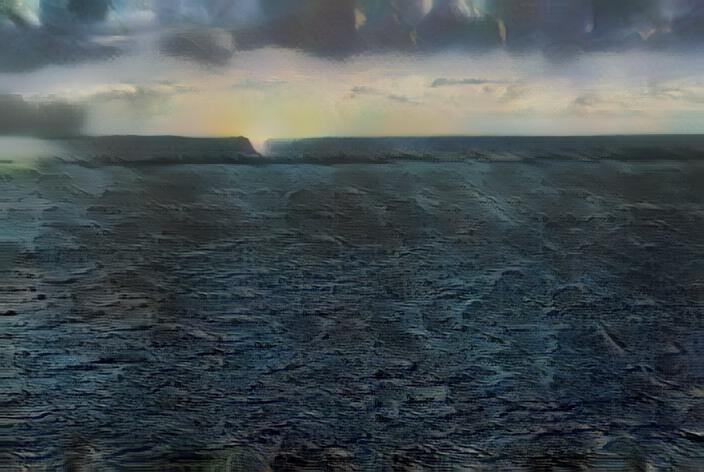} &
		\includegraphics[height = .18\linewidth, width = .32\linewidth]{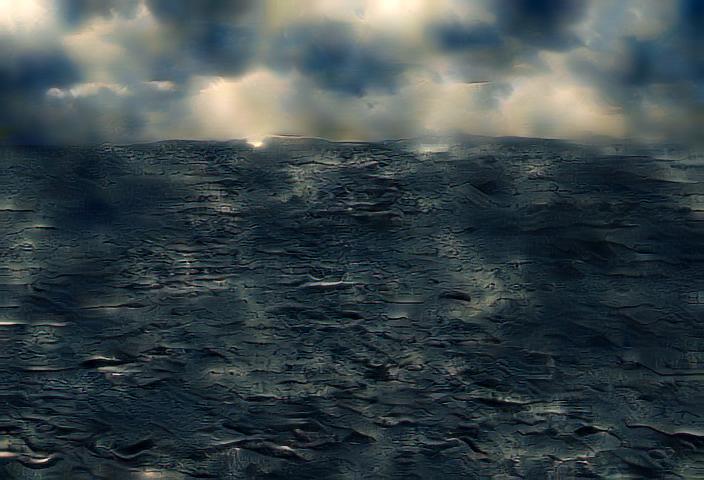} &
		\includegraphics[height = .18\linewidth, width = .32\linewidth]{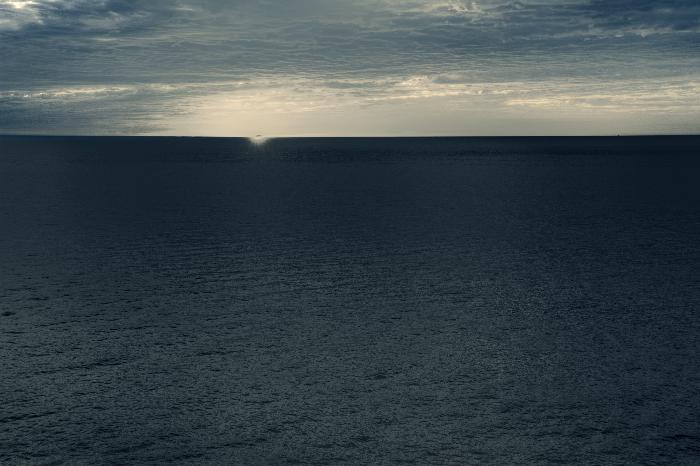} & \\
		{ (d) Huang \etal~\cite{Huang-2017-arbitrary}}& { (e) Li \etal~\cite{WCT-2017-NIPS}} & { (f) Ours }\\
	\end{tabular}
	\vspace{-0.5em}
	\caption{Visual comparison with artistic stylization algorithms. Note the structural distortions on object boundaries (e.g., building) and detailed edges (e.g., sea, cloud) generated by the competing stylization methods.}
	\label{fig:compare_artistic}
	\vspace{-1.5em}
\end{figure}

Figure~\ref{fig:compare_photorealistic} shows visual results of the evaluated photorealistic stylization algorithms. 
Overall, the images generated by the proposed algorithm exhibit better stylization effects. While both Reinhard \etal~\cite{reinhard-2001color} and Piti\'e \etal~\cite{Pitie-2005} change colors of the content photos, they fail to transfer the style. We argue that photorealistic stylization cannot be purely achieved via color transfer. It requires adding new patterns that represent the style photo to the content photo. For example, in the third example of Figure~\ref{fig:compare_photorealistic} (bottom), our algorithm not only changes the color of ground regions to white but also synthesizes the snow patterns as they appear in the style photo. The method of Luan \etal~\cite{Luan-2017-photorealism} achieves good stylization effects at first glance. However, a closer look reveals that the generated photos contain noticeable artifacts, e.g., the irregular brightness on buildings and trees. Several semantically similar regions are stylized inconsistently. 
%More visual comparison results \mbox{are given in the supplementary materials.}

Figure~\ref{fig:compare_artistic} shows the visual comparison between the proposed algorithm and artistic stylization algorithms. Although the other evaluated algorithms are able to transfer the style well, they render noticeable structural artifacts and inconsistent stylizations across the images. In contrast, our method produces more photorealistic results.

\paragraph{\bf User studies.} We resort to user studies for performance evaluation since photorealistic image stylization is a highly subjective task. Our benchmark dataset consists of a set of 25 content--style pairs provided by Luan \etal~\cite{Luan-2017-photorealism}\footnote{We note that the user studies reported in Luan \etal~\cite{Luan-2017-photorealism} are based on 8 different images in their dataset, which is about one third of our benchmark dataset size.}. We use the Amazon Mechanical Turk (AMT) platform for evaluation. % 
In each question, we show the AMT workers a content--style pair and the stylized results from the evaluated algorithms displayed in random order. The AMT workers\footnote{An AMT worker must have a lifetime Human Intelligent Task (HIT) approval rate greater than 98\% to qualify answering the questions.} are asked to select a stylized result based on the instructions. Each question is answered by 10 different workers. Hence, the performance score for each study is computed based on 250 questions. We compute the average number of times the images from an algorithm is selected, which is used as \mbox{the preference score of the algorithm.} 

We conduct two user studies. In one study, we ask the AMT workers to select which stylized photo better carries the target style. In the other study, we ask the workers to select which stylized photo looks more like a real photo (containing fewer artifacts). Through the studies, we would like to answer which algorithm better stylizes content images and which renders \mbox{better photorealistic outputs.}

In Table~\ref{tbl::photo}, we compare the proposed algorithm to Luan \etal~\cite{Luan-2017-photorealism}, which is the current state-of-the-art. The results show that 63.1\% of the users prefer the stylization results generated by our algorithm and 73.5\% regard our output photos as more photorealistic. We also compare our algorithm to the classical algorithm of Piti\'e \etal~\cite{Pitie-2005}. From Table~\ref{tbl::photo}, our results are as photorealistic as those computed by the classical algorithm (which simply performs color matching), and 55.2\% of the users consider our stylization results better. 

Table~\ref{tbl::art} compares our algorithm with the artistic stylization algorithms for user preference scores. We find our algorithm achieves a score of 56.4\% and 65.6\% for the stylization effect and photorealism, which are significantly better than the other algorithms. The artistic stylization algorithms do not perform well since \mbox{they are not designed for the photorealistic stylization task.}

\begin{table}[t]
	\caption{User preference: proposed vs. Luan \etal~and proposed vs. Piti\'e \etal}
	\label{tbl::photo}
	\centering
	\begin{tabular}{c|cc}
		\toprule
		& ~~Luan \etal~\cite{Luan-2017-photorealism} / proposed~~ & ~~Piti\'e \etal~\cite{Pitie-2005} / proposed~~ \\
		\midrule
		Better stylization & 36.9\% / \textbf{63.1\%}  & 44.8\% / \textbf{55.2\%}  \\
		Fewer artifacts     & 26.5\% / \textbf{73.5\%}  & 48.8\% / \textbf{51.2\%}  \\
		\bottomrule
	\end{tabular}
	\vspace{-0.5em}
	\caption{User preference: proposed versus \emph{artistic} stylization algorithms.}
	\label{tbl::art}
	\centering
	\begin{tabular}{c|cccc}
		\toprule
		& ~Gatys \etal~\cite{GatysTransfer-CVPR2016}~ & ~Huang \etal~\cite{Huang-2017-arbitrary}~  & ~Li \etal~\cite{WCT-2017-NIPS}~ & ~~proposed~~ \\
		\midrule
		Better stylization & 19.2\% & 8.4\% & 16.0\% & \textbf{56.4\%}\\
		Fewer artifacts     & 21.6\% & 6.0\% &  6.8\% & \textbf{65.6\%}\\
		\bottomrule
	\end{tabular}
	\vspace{-0.5em}
\end{table}

\begin{figure}[t]
	\centering
	\begin{tabular}{c@{\hspace{0.005\linewidth}}c@{\hspace{0.005\linewidth}}c@{\hspace{0.005\linewidth}}c@{\hspace{0.005\linewidth}}c}
		\includegraphics[height=.21\linewidth, width = .21\linewidth]{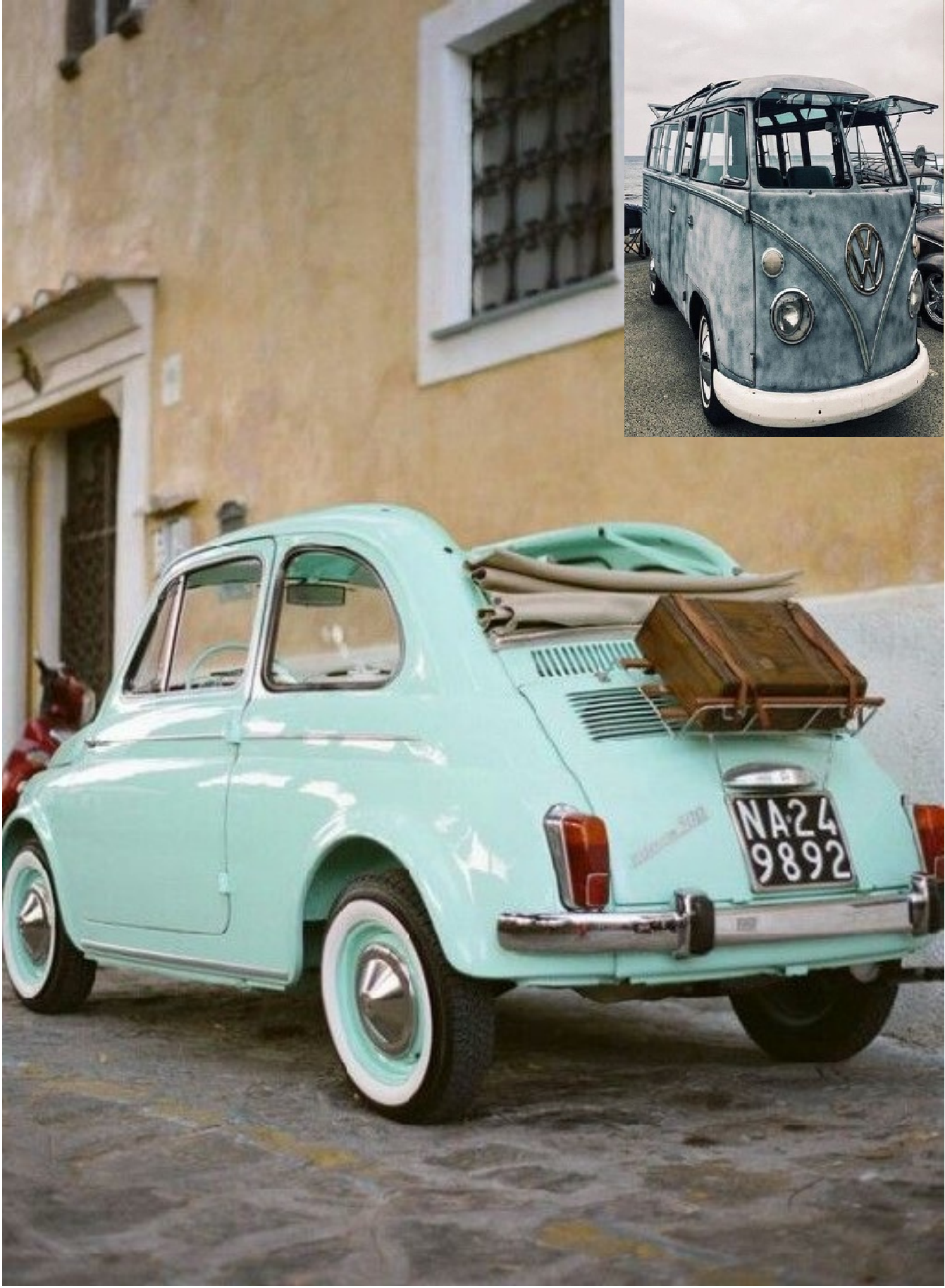} &
		\includegraphics[height=.21\linewidth, width = .21\linewidth]{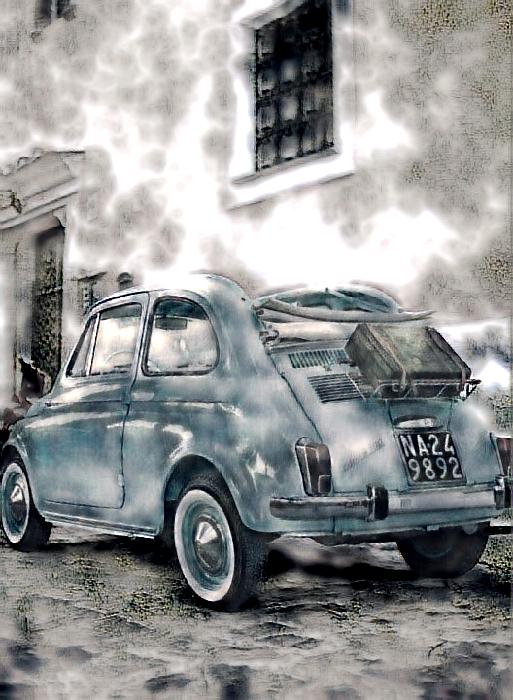} &
		\includegraphics[height=.21\linewidth, width = .21\linewidth]{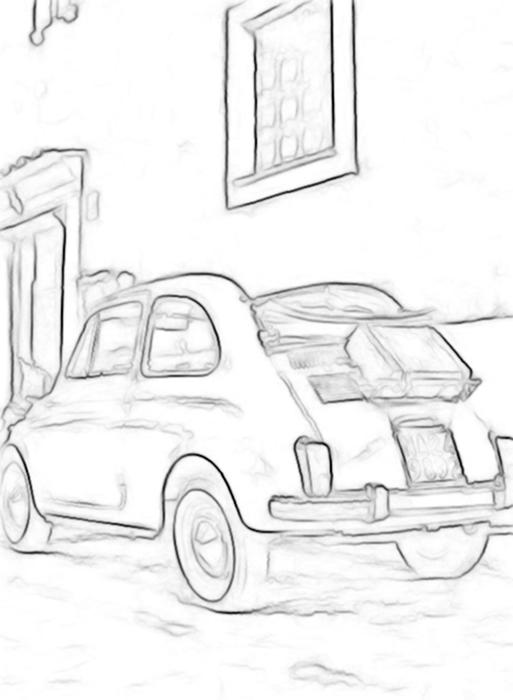} & 
		\includegraphics[width = .32\linewidth]{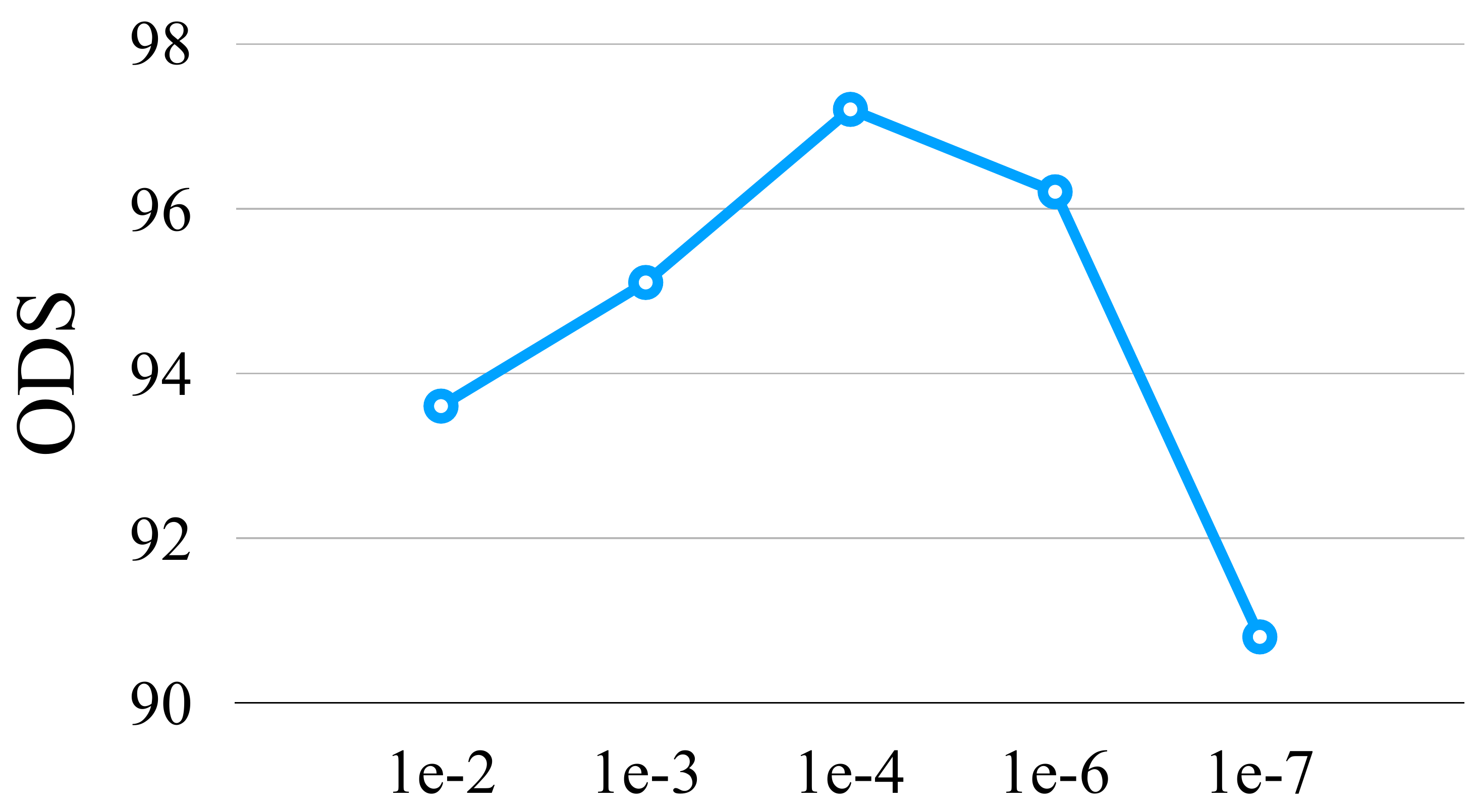} &\\[-2pt]
		{Content/Style} & {PhotoWCT} & {GT edges~\cite{HED-2015holistically}} & {$\lambda$}\\[4pt]
		\includegraphics[height=.21\linewidth, width = .21\linewidth]{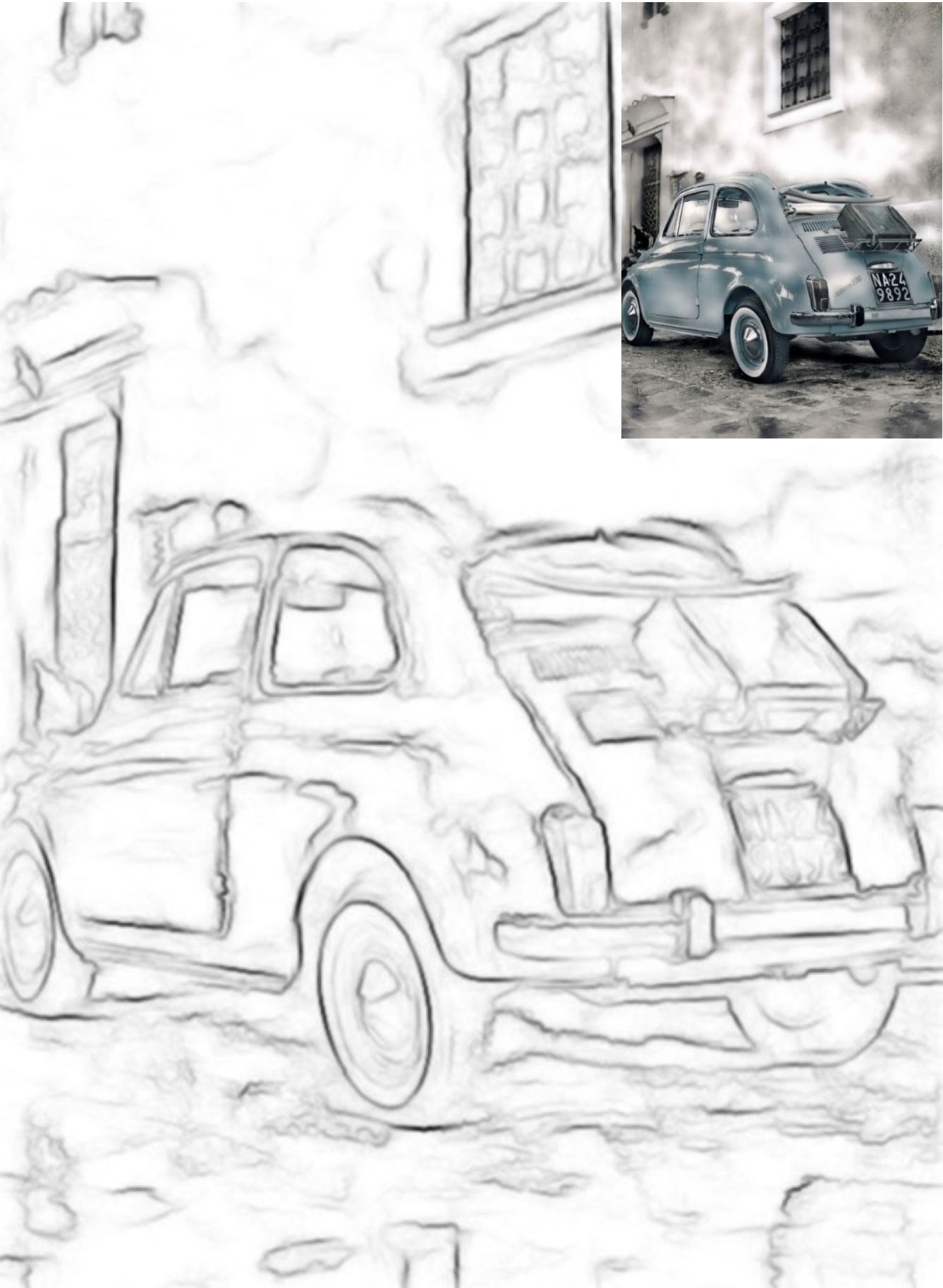} &
		\includegraphics[height=.21\linewidth, width = .21\linewidth]{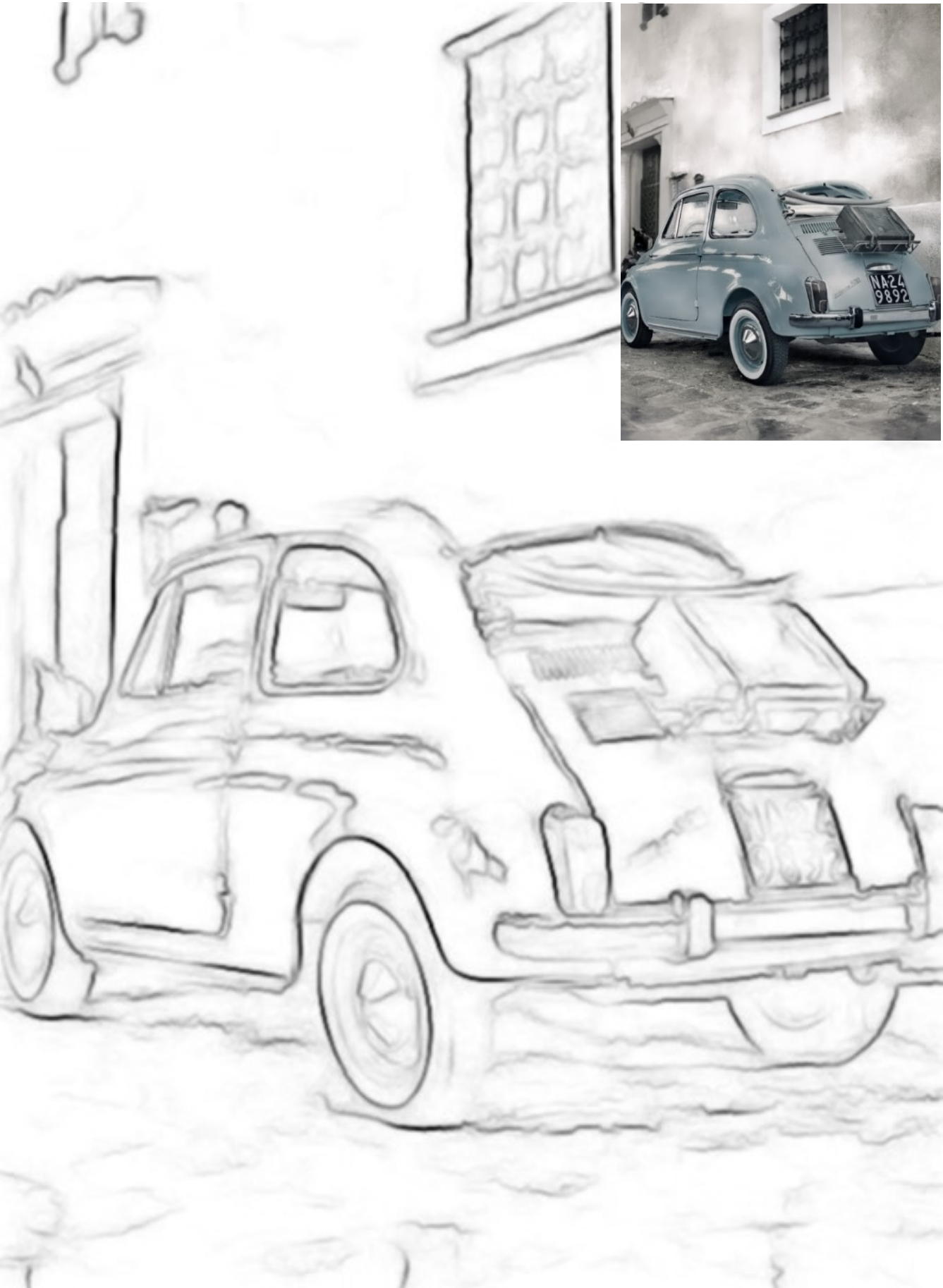} &
		\includegraphics[height=.21\linewidth, width = .21\linewidth]{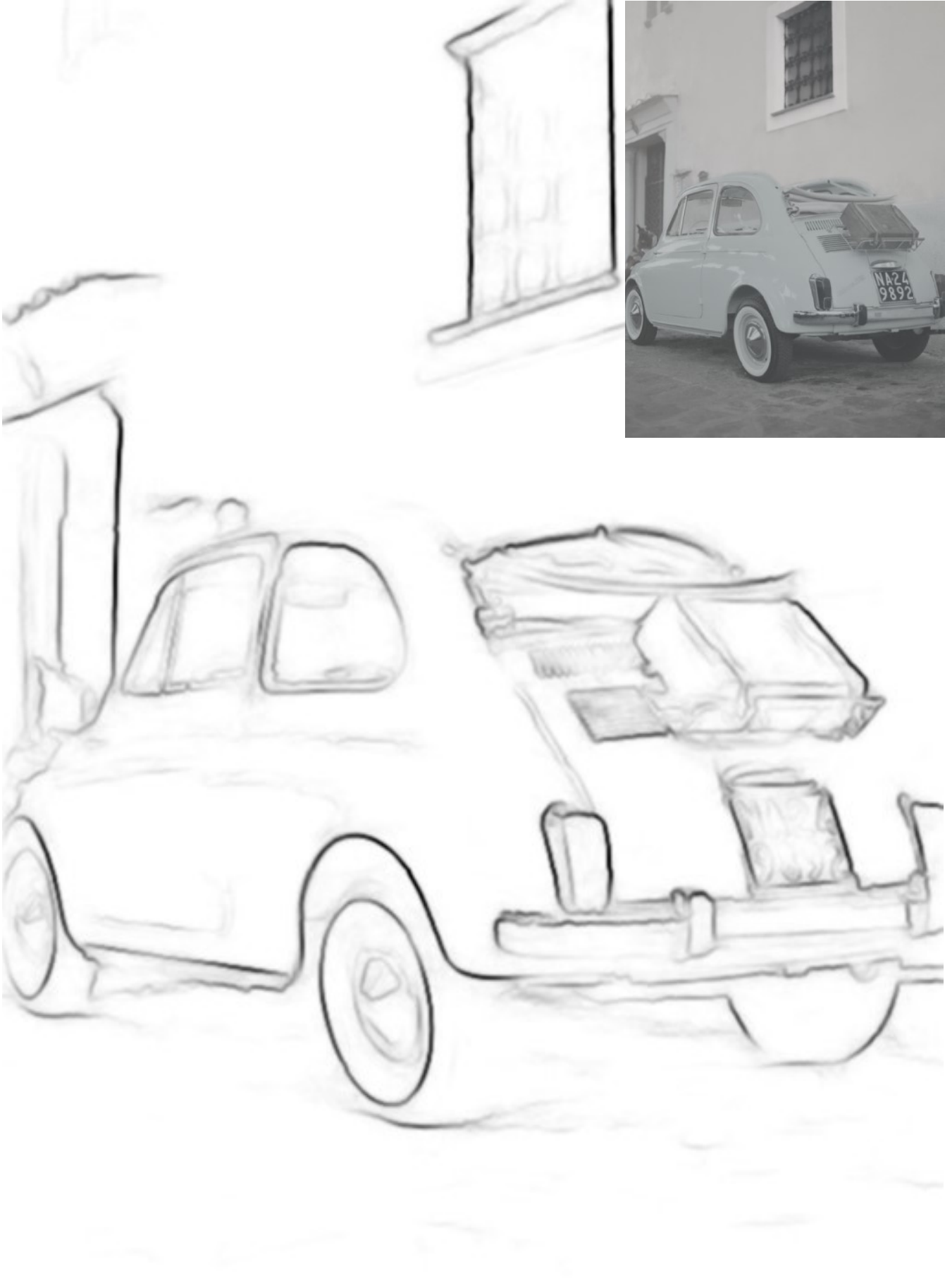} & 
		\includegraphics[width = .32\linewidth]{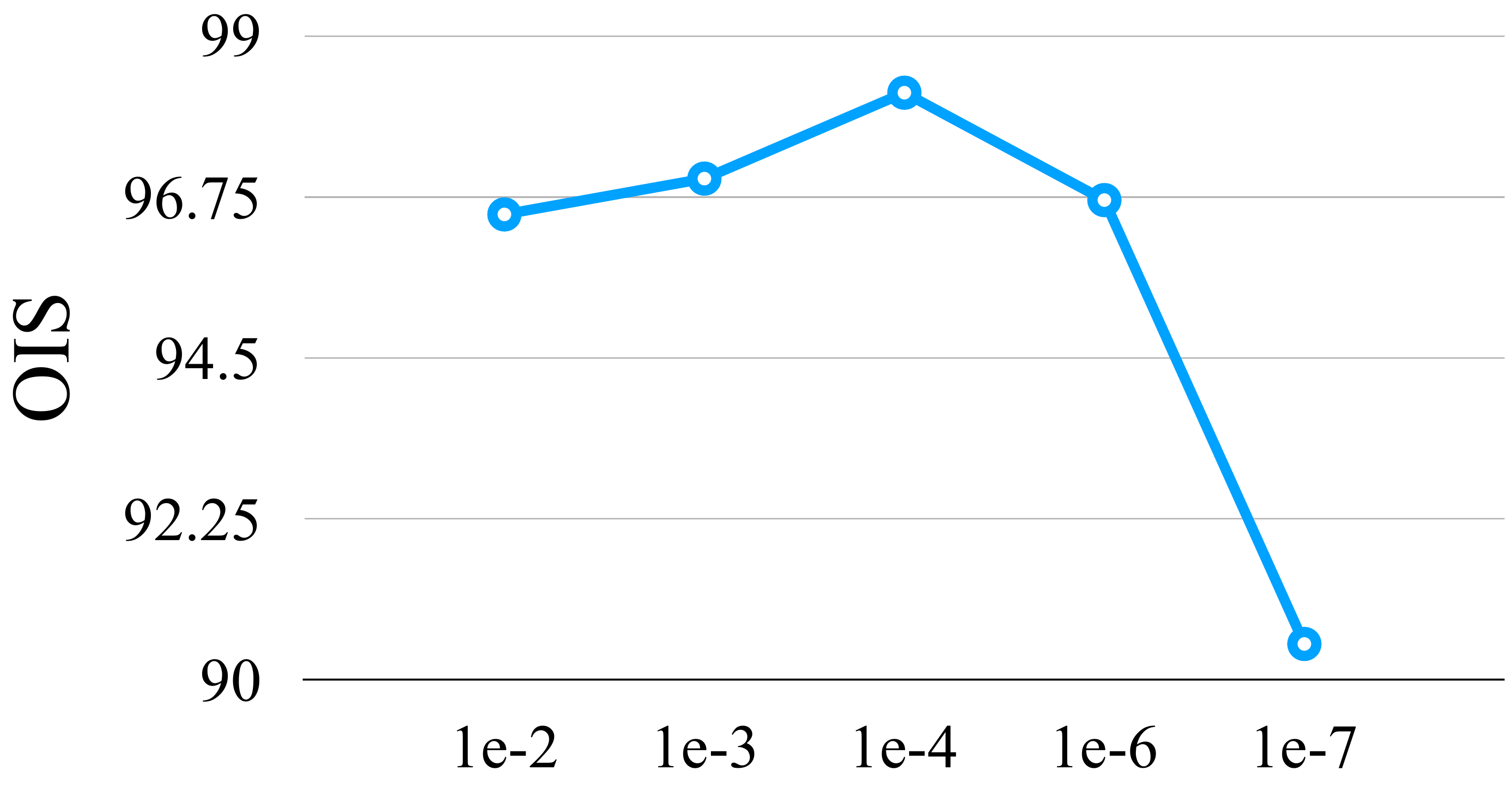} & \\[-2pt]
		{$\lambda=10^{-2}$} &  {$\lambda=10^{-4}$} & {$\lambda=10^{-6}$} &
		{$\lambda$} &\\
	\end{tabular}
	\vspace{-1.0em}
	\caption{Visualization of effects of using different $\lambda$ values in the photorealistic smoothing step. We show the edge maps of different stylization results (inset) at bottom and compare them with the edge map of the content in terms of the ODS and OIS metric (rightmost).
	}
	\label{fig:lambda}
	%\vspace{-0.5em}
	\centering
	\begin{tabular}{c@{\hspace{0.005\linewidth}}c@{\hspace{0.005\linewidth}}c@{\hspace{0.005\linewidth}}c@{\hspace{0.005\linewidth}}c@{\hspace{0.005\linewidth}}c@{\hspace{0.005\linewidth}}c}
		\includegraphics[width = .24\linewidth]{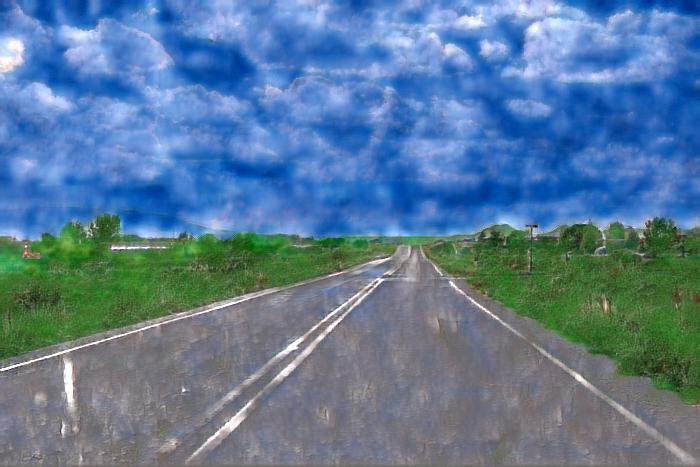} & 
		\hspace{0.5pt}\vrule\hspace{0.5pt}
		\includegraphics[width = .24\linewidth]{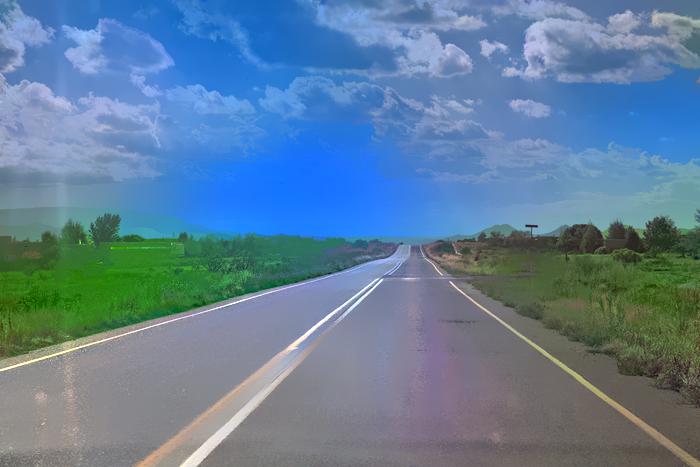} &
		\includegraphics[width = .24\linewidth]{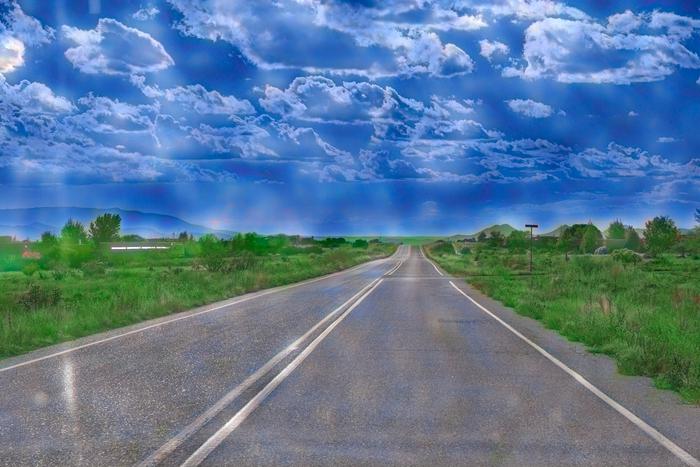} & 
		\includegraphics[width = .24\linewidth]{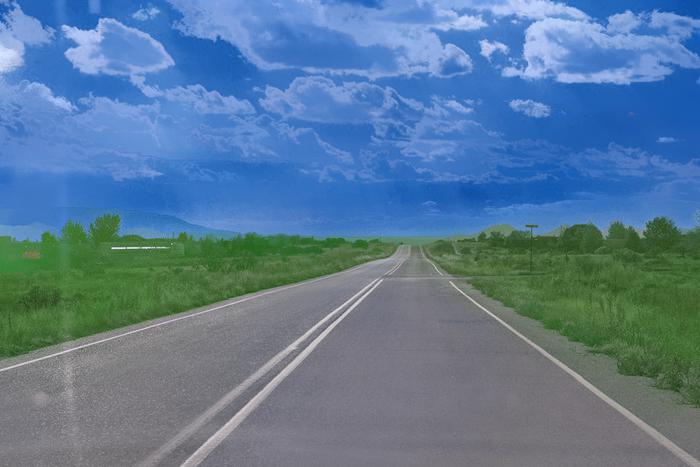} & \\
		{ (a) PhotoWCT } & {(b) Luan \etal~\cite{Luan-2017-photorealism}  } &  {(c) Mechrez \etal~\cite{MatchGradientSPE-2017-BMVC}} & {(d) proposed} & \\
	\end{tabular}
	\vspace{-1.0em}
	\caption{Comparison between using our photorealistic smoothing step and other refinement methods (b)-(d).}
	\label{fig:second_step}
	\vspace{-1.5em}
\end{figure}

\paragraph{\bf WCT versus PhotoWCT.} We compare the proposed algorithm with a variant where the PhotoWCT step is replaced by the WCT~\cite{WCT-2017-NIPS}. Again, we conduct two user studies on stylization effects and photorealism as described earlier. The result shows that the proposed algorithm is favored over its variant for better stylization 83.6\% of the \mbox{times and favored for better photorealism 83.2\% of the times.}

\paragraph{\bf Sensitivity analysis on $\lambda$.} In the photorealistic smoothing step, the $\lambda$ balances between the smoothness term and fitting term in (\ref{formula_propagation}). A smaller $\lambda$ renders smoother results, while a larger $\lambda$ renders results that are more faithful to the queries (the PhotoWCT result). Figure~\ref{fig:lambda} shows results of using different $\lambda$ values. In general, decreasing $\lambda$ helps remove artifacts and hence improves photorealism.  However, if $\lambda$ is too small, the output image tends to be over-smoothed. In order to find the optimal $\lambda$, we perform a grid search. We use the similarity between the boundary maps extracted from stylized and original content photos as the criteria since object boundaries should remain the same despite the stylization~\cite{cutzu-2003-estimating}. We employ the HED method~\cite{HED-2015holistically} for boundary detection and use two standard boundary detection metrics: ODS and OIS. A higher ODS or OIS score means a stylized photo better preserves the content in the original photo. The average scores over the benchmark dataset are shown on the rightmost of Figure~\ref{fig:lambda}. Based on the results, we use $\lambda=10^{-4}$ in all the experiments.

\paragraph{\bf Alternative smoothing techniques.} In Figure~\ref{fig:second_step}, we compare our photorealistic smoothing step with two alternative approaches. In the first approach, we use the PhotoWCT-stylized photo as the initial solution for solving the second optimization problem in the method of Luan \etal~\cite{Luan-2017-photorealism}. The result is shown in Figure~\ref{fig:second_step}(b). 
This approach leads to noticeable artifacts as the road color is distorted. In the second approach, we use the method of Mechrez \etal~\cite{MatchGradientSPE-2017-BMVC}, which refines stylized results by matching the gradients in the output photo to those in the content photo. As shown in Figure~\ref{fig:second_step}(c), we find this approach performs well for removing structural distortions on boundaries but does not remove visual artifacts. In contrast, our method (Figure~\ref{fig:second_step}(d)) generates more photorealistic results with an efficient closed-form solution.

\paragraph{\bf Run-time.} In Table~\ref{table:time}, we compare the run-time of the proposed algorithm to that of the state-of-the-art~\cite{Luan-2017-photorealism}. We note that while our algorithm has a closed-form solution, Luan \etal~\cite{Luan-2017-photorealism} rely on non-convex optimization. To stylize a photo, Luan \etal~\cite{Luan-2017-photorealism} solve two non-convex optimization problems sequentially where a solution\footnote{Note the solution is at most local optimal.} to the first optimization problem is used as an initial solution to solve the second optimization problem. 
% Jan suggested to remove
%Both of the problems stem from the neural style transfer algorithm~\cite{GatysTransfer-CVPR2016} and are solved by iterative optimization. 
%
We report the total run-time required for obtaining the final stylization results. We resize the content images in the benchmark dataset to different sizes and report the average run-time for each image size. The experiment is conducted on a PC with an NVIDIA Titan X Pascal GPU. 
To stylize images of 1024$\times$512 resolution, our algorithm takes 13.16 seconds, which is 49 times faster than 650.45 seconds achieved by Luan \etal~\cite{Luan-2017-photorealism}. 

\begin{table}[t]
	\caption{Run-time comparison. We compute the average run time (in seconds) of the evaluated algorithms across various image resolutions. }
	\label{table:time}
	\centering
	\begin{tabular}{c|c|ccc|c}
		\toprule
		Image resolution & ~Luan~\etal\cite{Luan-2017-photorealism}~ & ~proposed~ & ~PhotoWCT~ & ~smoothing~ & \texttt{approx} \\
		\midrule
		256$\times$128 &  ~79.61  &  ~0.96 & 0.40 & ~0.56 & 0.41\\
		512$\times$256 &  186.52 &  ~2.95 & 0.42 & ~2.53 & 0.47\\
		768$\times$384 &  380.82 &  ~7.05 & 0.53 & ~6.52 & 0.55\\
		1024$\times$512 & 650.45 &  13.16& 0.56 & 12.60& 0.64\\
		\bottomrule
	\end{tabular}
	\vspace{-0.5em}
	\caption{User preference score comparison: comparing \texttt{approx} (the fast approximation of the proposed algorithm) to the proposed algorithm as well as other photorealistic stylization algorithms.} 
	\label{tbl::approx}
	\centering
	\begin{tabular}{c|ccc}
		\toprule & 
		proposed/\texttt{approx} \  & 
		Luan \etal~\cite{Luan-2017-photorealism}/\texttt{approx} \  & 
		Piti\'e \etal~\cite{Pitie-2005}/\texttt{approx} \  \\
		\midrule
		Better stylization & 
		\textbf{59.6\%} / 40.4 & 
		36.4 / \textbf{63.6\%} & 
		46.0 / \textbf{54.0\%} \\
		Fewer artifacts     & \textbf{52.8\%} / 47.2 & 
		20.8 / \textbf{79.2\%} & 
		46.8 / \textbf{53.2\%}  \\
		\bottomrule
	\end{tabular}
	\vspace{-0.5em}
\end{table}

In Table~\ref{table:time}, we also report the run-time of each step in our algorithm. We find the smoothing step takes most of the computation time, since it involves inverting the sparse matrix $W$ in~(\ref{formula_closedform0}) using the LU decomposition. By employing efficient LU-decomposition algorithms developed for large sparse matrices, the complexity can be roughly determined by the number of non-zero entries in the matrices only. In our case, since each pixel is only connected to its neighbors (e.g., 3$\times$3 window), the number of non-zero values in $W$ grows \mbox{linearly with the image size.}

For further speed-up, we can approximate the smoothing step using guided image filtering~\cite{he2013guided}, which can smooth the PhotoWCT output based on the content photo. We will refer to this version of our algorithm \texttt{approx}. Although approximating the smoothing step with guided image filtering results in slightly degraded performance as comparing to the original algorithm, it leads to a large speed gain as shown in Table~\ref{table:time}. To stylize images of 1024$\times$512 resolution, \texttt{approx} only takes 0.64 seconds, which is 1,016 times faster than 650.45 seconds achieved by Luan \etal~\cite{Luan-2017-photorealism}. 
To quantify the performance degradation due to the approximation, we conduct additional user studies comparing the proposed algorithm and its approximation. We use the same evaluation protocol as described above. The results are shown in Table~\ref{tbl::approx}.
In general, the stylization results rendered by \texttt{approx} are less preferred by the users as compared to those generated by the full algorithm. 
However, the results from \texttt{approx} are still preferred over other methods
in terms of both stylization effects and photorealism.

\begin{figure}[t]
	\centering
	\begin{tabular}{c@{\hspace{0.005\linewidth}}c@{\hspace{0.005\linewidth}}c@{\hspace{0.005\linewidth}}c@{\hspace{0.005\linewidth}}c@{\hspace{0.005\linewidth}}c}
		\includegraphics[width = .192\linewidth]{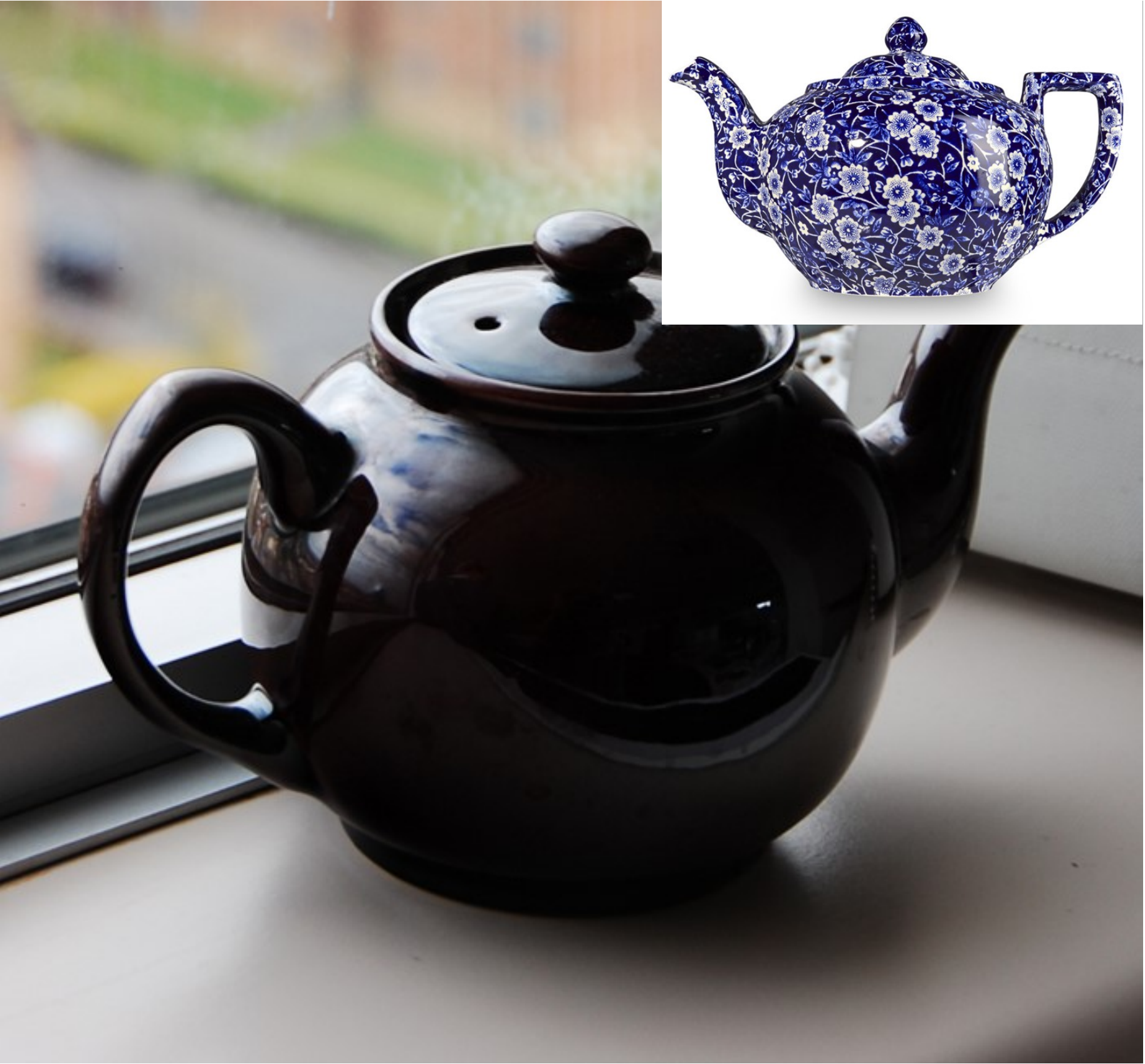} & 
		\includegraphics[width = .192\linewidth]{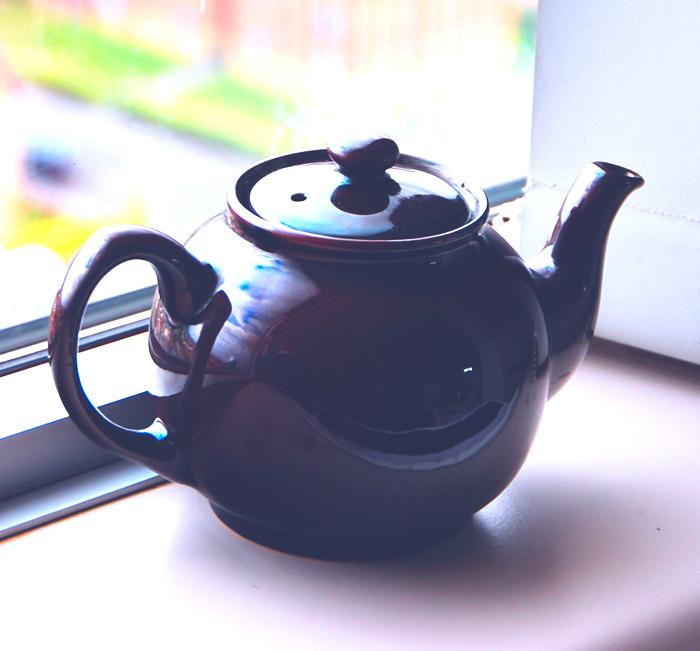} & 
		\includegraphics[width = .192\linewidth]{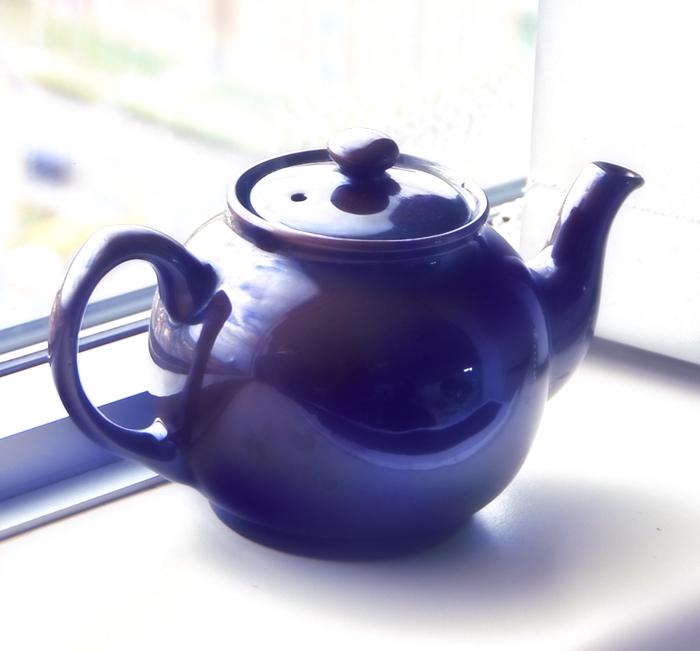} & 
		\includegraphics[width = .192\linewidth]{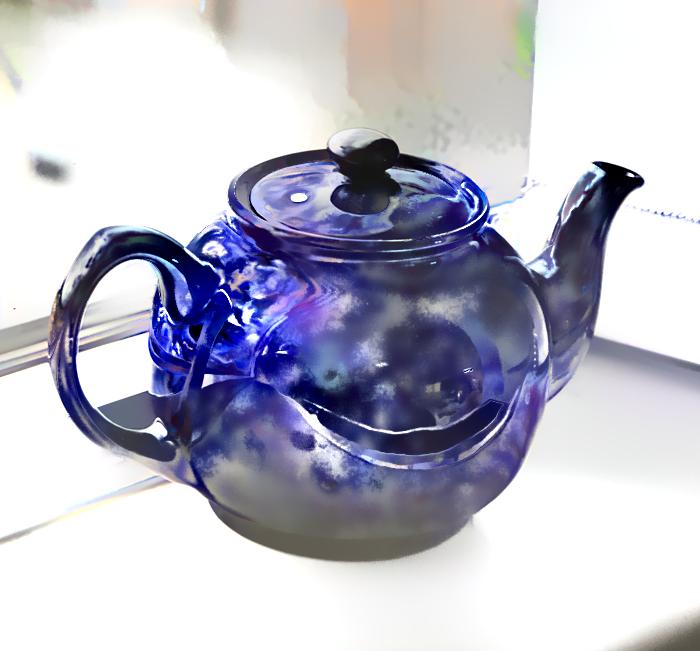} & 
		\includegraphics[width = .192\linewidth]{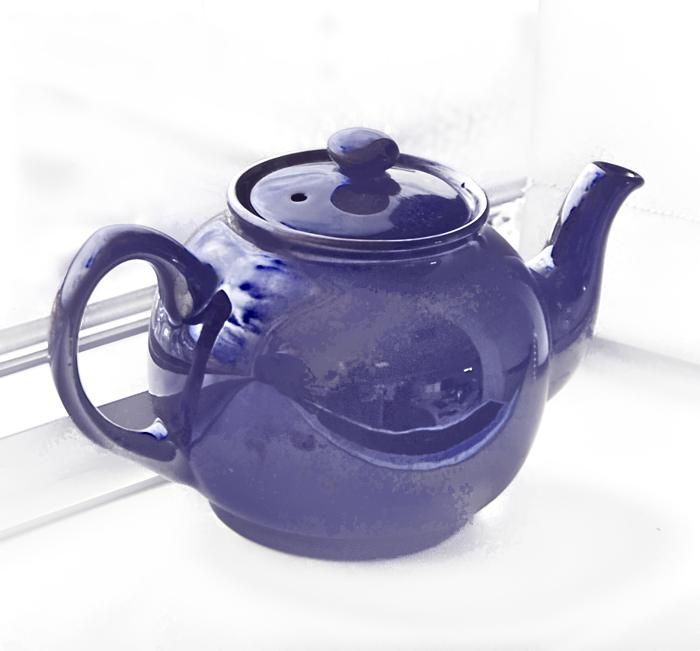} & \\
		{Content/Style } &  {Reinhard \etal~\cite{reinhard-2001color} } & {Piti\'e \etal~\cite{Pitie-2005}} & {Luan \etal~\cite{Luan-2017-photorealism} } & {\small{Ours }} & \\
	\end{tabular}
	\vspace{-1.0em}
	\caption{Failure case. Both the proposed and other photorealistic stylization algorithms fail to transfer the flower patterns to the pot.}
	\label{fig:failure_case}
	\vspace{-1.5em}
\end{figure}

\paragraph{\bf Failure case.}~Figure~\ref{fig:failure_case} shows a failure case where the proposed method fails to transfer the flower patterns in the style photo to the content photo. Similar limitations also apply to the other photorealistic stylization methods~\cite{Pitie-2005,reinhard-2001color,Luan-2017-photorealism}. Since the proposed method uses the pixel affinity of the content in the photorealistic smoothing step, it favors a stylization output with smooth color transition on the pot surface as in the input photo.

\section{Conclusions}

We presented a novel fast photorealistic image stylization method. It consists of a stylization step and a photorealistic smoothing step. Both steps have efficient closed-form solutions. Experimental results show that our algorithm generates stylization outputs that are much more preferred by human subject as compared to those by the state-of-the-art, while running much faster.

\bibliographystyle{splncs}
\bibliography{reference}

\begin{thebibliography}{10}

\bibitem{reinhard-2001color}
Reinhard, E., Ashikhmin, M., Gooch, B., Shirley, P.:
\newblock Color transfer between images.
\newblock IEEE Computer graphics and applications \textbf{21}(5) (2001)  34--41

\bibitem{Pitie-2005}
Piti\'e, F., Kokaram, A.C., Dahyot, R.:
\newblock N-dimensional probability density function transfer and its
  application to color transfer.
\newblock In: ICCV. (2005)

\bibitem{sunkavalli2010multi}
Sunkavalli, K., Johnson, M.K., Matusik, W., Pfister, H.:
\newblock Multi-scale image harmonization.
\newblock ACM Transactions on Graphics \textbf{29}(4) (2010)  125

\bibitem{bae-2006two}
Bae, S., Paris, S., Durand, F.:
\newblock Two-scale tone management for photographic look.
\newblock ACM Transactions on Graphics \textbf{25}(3) (2006)  637--645

\bibitem{laffont-2014-transient}
Laffont, P.Y., Ren, Z., Tao, X., Qian, C., Hays, J.:
\newblock Transient attributes for high-level understanding and editing of
  outdoor scenes.
\newblock ACM Transactions on Graphics \textbf{33}(4) (2014)  149

\bibitem{shih2-2014style}
Shih, Y., Paris, S., Barnes, C., Freeman, W.T., Durand, F.:
\newblock Style transfer for headshot portraits.
\newblock In: SIGGRAPH. (2014)

\bibitem{GatysTexture-NIPS2015}
Gatys, L.A., Ecker, A.S., Bethge, M.:
\newblock Texture synthesis using convolutional neural networks.
\newblock In: NIPS. (2015)

\bibitem{GatysTransfer-CVPR2016}
Gatys, L.A., Ecker, A.S., Bethge, M.:
\newblock Image style transfer using convolutional neural networks.
\newblock In: CVPR. (2016)

\bibitem{Luan-2017-photorealism}
Luan, F., Paris, S., Shechtman, E., Bala, K.:
\newblock Deep photo style transfer.
\newblock In: CVPR. (2017)

\bibitem{WCT-2017-NIPS}
Li, Y., Fang, C., Yang, J., Wang, Z., Lu, X., Yang, M.H.:
\newblock Universal style transfer via feature transforms.
\newblock In: NIPS. (2017)

\bibitem{freedman2010object}
Freedman, D., Kisilev, P.:
\newblock Object-to-object color transfer: Optimal flows and smsp
  transformations.
\newblock In: CVPR. (2010)

\bibitem{shih1-2013data}
Shih, Y., Paris, S., Durand, F., Freeman, W.T.:
\newblock Data-driven hallucination of different times of day from a single
  outdoor photo.
\newblock In: SIGGRAPH. (2013)

\bibitem{Wu-2013-content}
Wu, F., Dong, W., Kong, Y., Mei, X., Paul, J.C., Zhang, X.:
\newblock Content-based colour transfer.
\newblock Computer Graphics Forum \textbf{32}(1) (2013)  190--203

\bibitem{tsai2016sky}
Tsai, Y.H., Shen, X., Lin, Z., Sunkavalli, K., Yang, M.H.:
\newblock Sky is not the limit: Semantic-aware sky replacement.
\newblock ACM Transactions on Graphics \textbf{35}(4) (2016)  149

\bibitem{MrfTransfer-CVPR2016}
Li, C., Wand, M.:
\newblock Combining markov random fields and convolutional neural networks for
  image synthesis.
\newblock In: CVPR. (2016)

\bibitem{Texturenet-ICML2016}
Ulyanov, D., Lebedev, V., Vedaldi, A., Lempitsky, V.:
\newblock Texture networks: Feed-forward synthesis of textures and stylized
  images.
\newblock In: ICML. (2016)

\bibitem{Perceptual-ECCV2016}
Johnson, J., Alahi, A., Fei-Fei, L.:
\newblock Perceptual losses for real-time style transfer and super-resolution.
\newblock In: ECCV. (2016)

\bibitem{Me-2017-diversified}
Li, Y., Fang, C., Yang, J., Wang, Z., Lu, X., Yang, M.H.:
\newblock Diversified texture synthesis with feed-forward networks.
\newblock In: CVPR. (2017)

\bibitem{chen2017stylebank}
Chen, D., Yuan, L., Liao, J., Yu, N., Hua, G.:
\newblock Stylebank: An explicit representation for neural image style
  transfer.
\newblock In: CVPR. (2017)

\bibitem{GoogleMultiTexture-2016}
Dumoulin, V., Shlens, J., Kudlur, M.:
\newblock A learned representation for artistic style.
\newblock In: ICLR. (2017)

\bibitem{Ghiasi-2017-BMVC}
Ghiasi, G., Lee, H., Kudlur, M., Dumoulin, V., Shlens, J.:
\newblock Exploring the structure of a real-time, arbitrary neural artistic
  stylization network.
\newblock In: BMVC. (2017)

\bibitem{Huang-2017-arbitrary}
Huang, X., Belongie, S.:
\newblock Arbitrary style transfer in real-time with adaptive instance
  normalization.
\newblock In: ICCV. (2017)

\bibitem{MSRA-2017-visual}
Liao, J., Yao, Y., Yuan, L., Hua, G., Kang, S.B.:
\newblock Visual attribute transfer through deep image analogy.
\newblock arXiv preprint arXiv:1705.01088 (2017)

\bibitem{Lapstyle-ACMMM}
Li, S., Xu, X., Nie, L., Chua, T.S.:
\newblock Laplacian-steered neural style transfer.
\newblock In: ACM MM. (2017)

\bibitem{MatchGradientSPE-2017-BMVC}
Mechrez, R., Shechtman, E., Zelnik-Manor, L.:
\newblock Photorealistic style transfer with screened poisson equation.
\newblock In: BMVC. (2017)

\bibitem{isola2016image}
Isola, P., Zhu, J.Y., Zhou, T., Efros, A.A.:
\newblock Image-to-image translation with conditional adversarial networks.
\newblock In: CVPR. (2017)

\bibitem{wang2017high}
Wang, T.C., Liu, M.Y., Zhu, J.Y., Tao, A., Kautz, J., Catanzaro, B.:
\newblock High-resolution image synthesis and semantic manipulation with
  conditional gans.
\newblock In: CVPR. (2018)

\bibitem{liu2016coupled}
Liu, M.Y., Tuzel, O.:
\newblock Coupled generative adversarial networks.
\newblock In: NIPS. (2016)

\bibitem{taigman2016unsupervised}
Taigman, Y., Polyak, A., Wolf, L.:
\newblock Unsupervised cross-domain image generation.
\newblock In: ICLR. (2017)

\bibitem{shrivastava2016learning}
Shrivastava, A., Pfister, T., Tuzel, O., Susskind, J., Wang, W., Webb, R.:
\newblock Learning from simulated and unsupervised images through adversarial
  training.
\newblock In: CVPR. (2017)

\bibitem{liu2016unsupervised}
Liu, M.Y., Breuel, T., Kautz, J.:
\newblock Unsupervised image-to-image translation networks.
\newblock In: NIPS. (2017)

\bibitem{zhu2017unpaired}
Zhu, J.Y., Park, T., Isola, P., Efros, A.A.:
\newblock Unpaired image-to-image translation using cycle-consistent
  adversarial networks.
\newblock In: ICCV. (2017)

\bibitem{MUNIT}
Huang, X., Liu, M.Y., Belongie, S., Kautz, J.:
\newblock Multimodal unsupervised image-to-image translation.
\newblock In: ECCV. (2018)

\bibitem{VGG-2014}
Simonyan, K., Zisserman, A.:
\newblock Very deep convolutional networks for large-scale image recognition.
\newblock In: ICLR. (2015)

\bibitem{SWWAE-2016-ICLR}
Zhao, J., Mathieu, M., Goroshin, R., LeCun, Y.:
\newblock Stacked what-where auto-encoders.
\newblock In: ICLR Workshop. (2016)

\bibitem{zeiler-2014-visualizing}
Zeiler, M.D., Fergus, R.:
\newblock Visualizing and understanding convolutional networks.
\newblock In: ECCV. (2014)

\bibitem{noh-2015-learning}
Noh, H., Hong, S., Han, B.:
\newblock Learning deconvolution network for semantic segmentation.
\newblock In: ICCV. (2015)

\bibitem{zhou2004ranking}
Zhou, D., Weston, J., Gretton, A., Bousquet, O., Sch{\"o}lkopf, B.:
\newblock Ranking on data manifolds.
\newblock In: NIPS. (2004)

\bibitem{MR-2013-saliency}
Yang, C., Zhang, L., Lu, H., Ruan, X., Yang, M.H.:
\newblock Saliency detection via graph-based manifold ranking.
\newblock In: CVPR. (2013)

\bibitem{shi-2000-normalized}
Shi, J., Malik, J.:
\newblock Normalized cuts and image segmentation.
\newblock PAMI \textbf{22}(8) (2000)  888--905

\bibitem{Levin-2008-closed}
Levin, A., Lischinski, D., Weiss, Y.:
\newblock A closed-form solution to natural image matting.
\newblock PAMI \textbf{30}(2) (2008)  228--242

\bibitem{zelnik-2005-self}
Zelnik-Manor, L., Perona, P.:
\newblock Self-tuning spectral clustering.
\newblock In: NIPS. (2005)

\bibitem{COCO-lin2014-microsoft}
Lin, T.Y., Maire, M., Belongie, S., Hays, J., Perona, P., Ramanan, D.,
  Doll{\'a}r, P., Zitnick, C.L.:
\newblock Microsoft {COCO}: Common objects in context.
\newblock In: ECCV. (2014)

\bibitem{Gatys2016-control}
Gatys, L.A., Ecker, A.S., Bethge, M., Hertzmann, A., Shechtman, E.:
\newblock Controlling perceptual factors in neural style transfer.
\newblock In: CVPR. (2017)

\bibitem{HED-2015holistically}
Xie, S., Tu, Z.:
\newblock Holistically-nested edge detection.
\newblock In: ICCV. (2015)

\bibitem{cutzu-2003-estimating}
Cutzu, F., Hammoud, R., Leykin, A.:
\newblock Estimating the photorealism of images: Distinguishing paintings from
  photographs.
\newblock In: CVPR. (2003)

\bibitem{he2013guided}
He, K., Sun, J., Tang, X.:
\newblock Guided image filtering.
\newblock PAMI \textbf{35}(6) (2013)  1397--1409

\end{thebibliography}

\clearpage

\appendix

\section{Multi-level Stylization}

\begin{figure}[h]
	\centering
	\begin{tabular}{c@{\hspace{0.005\linewidth}}c@{\hspace{0.005\linewidth}}c}
		\includegraphics[width = .75\linewidth]{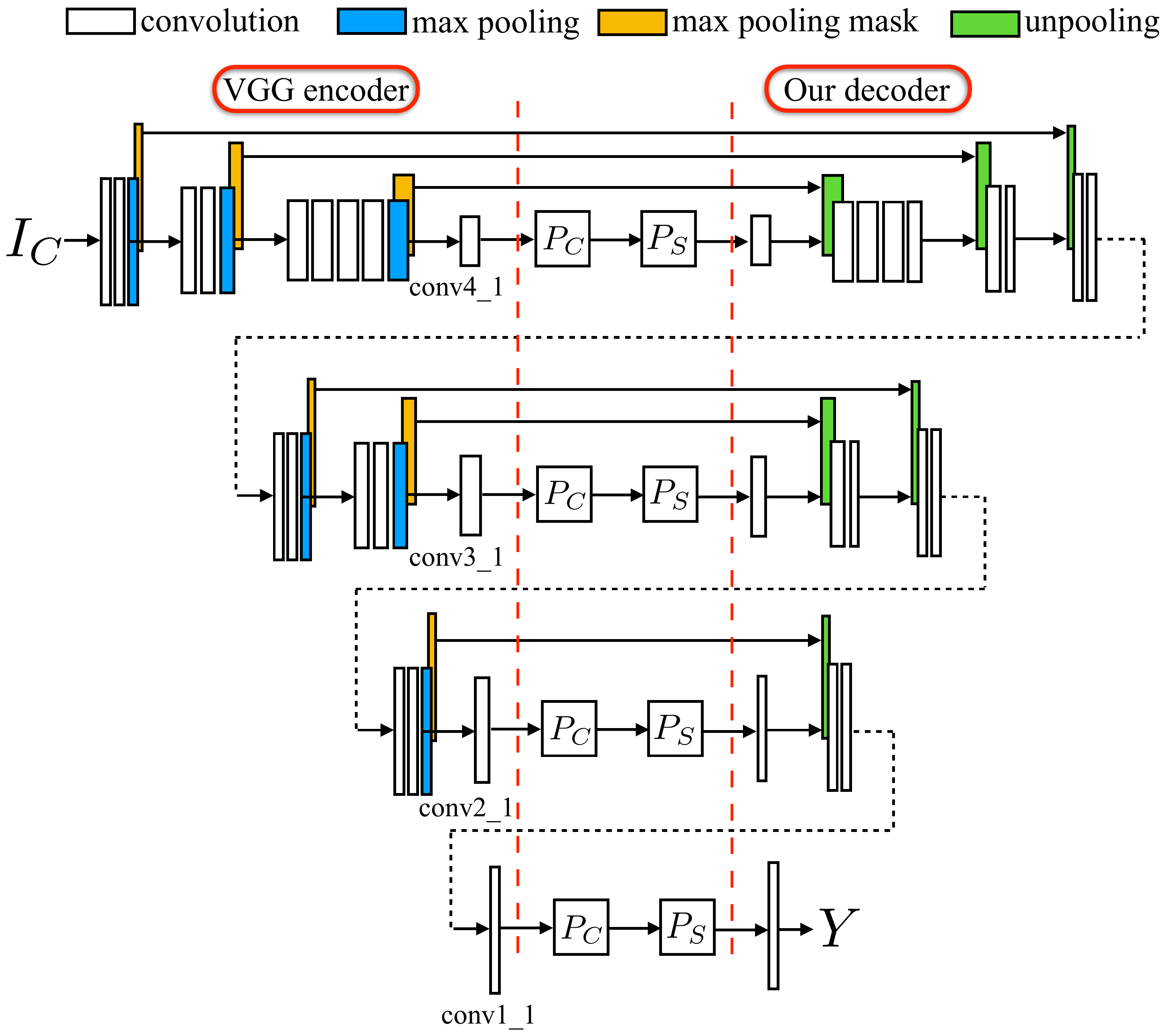} & \\
	\end{tabular}
	\caption{Illustration of the multi-level stylization scheme}
	\label{fig:multi_level}
\end{figure}

The PhotoWCT stylization step utilizes an auto-encoder with unpooling layers and a pair of feature transforms ($P_C$, $P_S$). The encoder is made of the first few layers of the VGG-19~\cite{VGG-2014} network. The feature transforms are applied to the features extracted by the encoder. As suggested in the WCT~\cite{WCT-2017-NIPS}, we match features across different levels in the VGG-19 encoder to fully capture the characteristics of the style. Specifically, we train four decoders for image reconstruction. They are responsible for inverting features extracted from $conv1\_1$, $conv2\_1$, $conv3\_1$, and $conv4\_1$ layer of VGG-19, respectively. With the four encoders, we have a set of 4 auto-encoder networks, which corresponds to a set of 4 PhotoWCT transforms. We first apply the transform that uses the deepest feature representation to stylize the content image. The stylized image is then passed to the transform that uses the second highest feature representation as shown in Figure~\ref{fig:multi_level}. Note that the decoders are trained separately and they do not share weights.

\section{Network Architecture}

Table~\ref{table:decoder} shows the detailed configurations of the decoders. We use the following abbreviation for ease of presentation: N=Filter number, K=Filter size, S=Stride.

\begin{table}[!htbp]
	\vspace{-2mm}
	\caption{Details of the decoders.}
	\label{table:decoder}
	\centering
	\begin{tabular}{llcccc}
		\toprule
		
		Layer Name~ & Specification & Decoder 1 & Decoder 2 & Decoder 3 & Decoder 4 \\
		\midrule
		$inv-conv4\_1$~ & Conv (N256, K3, S1), ReLU & v &  &  &  \\
		& MaxUnpooling (K2, S2)     & v &  &  &  \\
		$inv-conv3\_4$~ & Conv (N256, K3, S1), ReLU & v &  &  &  \\
		$inv-conv3\_3$~ & Conv (N256, K3, S1), ReLU & v &  &  &  \\
		$inv-conv3\_2$~ & Conv (N256, K3, S1), ReLU & v &  &  &  \\
		$inv-conv3\_1$~ & Conv (N128, K3, S1), ReLU & v & v &  &  \\
		& MaxUnpooling (K2, S2)     & v & v &  &  \\
		$inv-conv2\_2$~ & Conv (N128, K3, S1), ReLU & v & v &  &  \\
		$inv-conv2\_1$~ & Conv (N64, K3, S1), ReLU  & v & v & v &  \\
		& MaxUnpooling (K2, S2)     & v & v & v &  \\
		$inv-conv1\_2$~ & Conv (N64, K3, S1), ReLU  & v & v & v &  \\
		$inv-conv1\_1$~ & Conv (N3, K3, S1)         & v & v & v & v \\
		\bottomrule
	\end{tabular}
\end{table}

\section{Semantic Label Map}

\begin{figure}[!htbp]
\centering

\begin{tabular}{c@{\hspace{0.005\linewidth}}c@{\hspace{0.005\linewidth}}c@{\hspace{0.005\linewidth}}c@{\hspace{0.005\linewidth}}c@{\hspace{0.005\linewidth}}c@{\hspace{0.005\linewidth}}c}

\includegraphics[width = .16\linewidth]{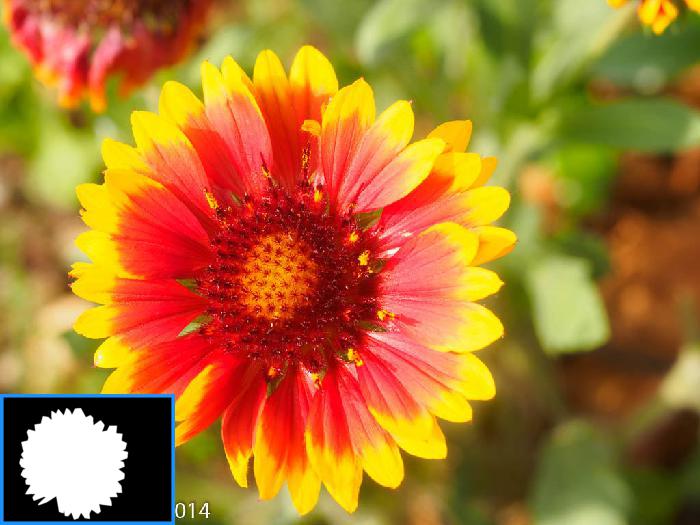} & 

\hspace{1pt}\vrule\hspace{1pt}

\includegraphics[width = .16\linewidth]{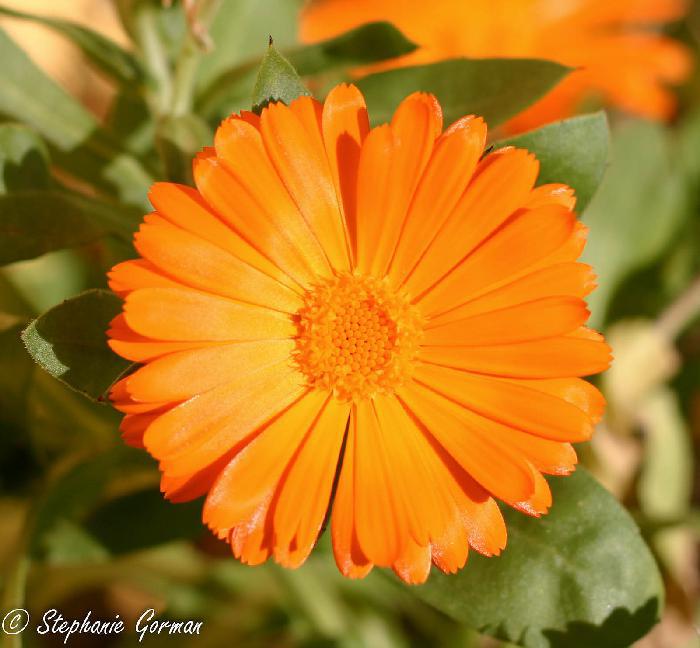} &
\includegraphics[width = .16\linewidth]{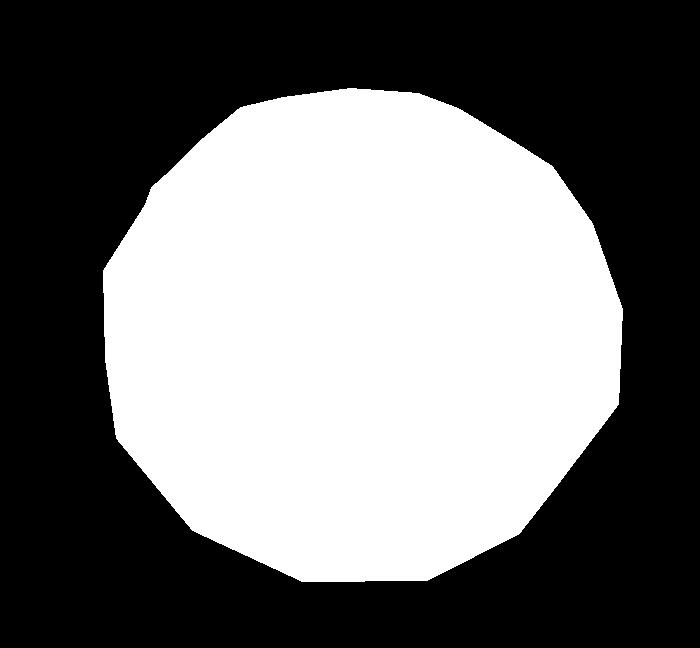} & 
\includegraphics[width = .16\linewidth]{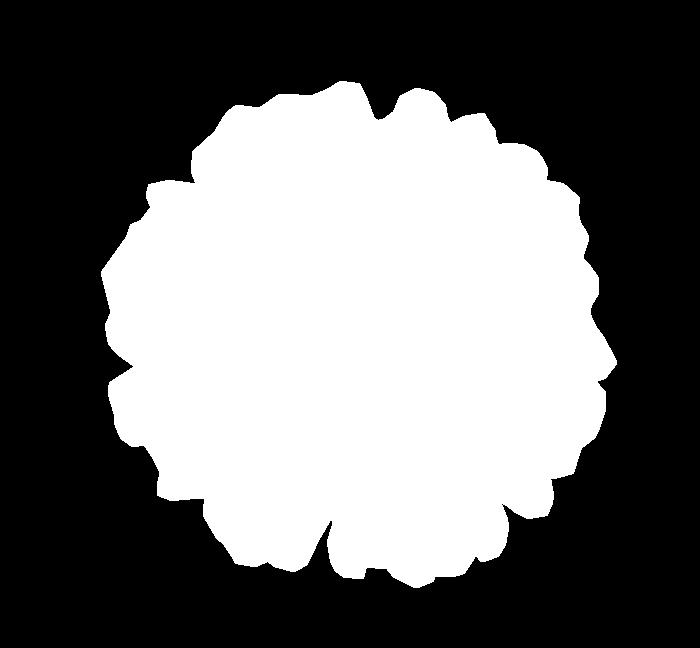} & 
\includegraphics[width = .16\linewidth]{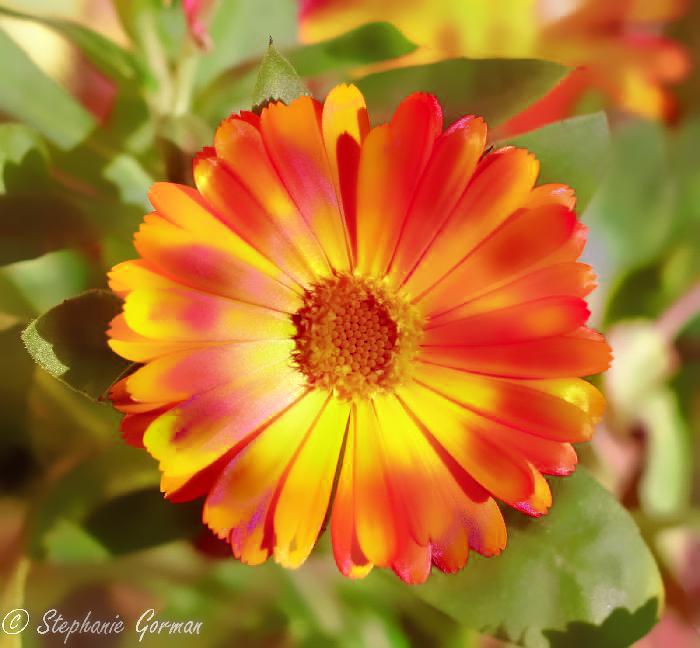} & 
\includegraphics[width = .16\linewidth]{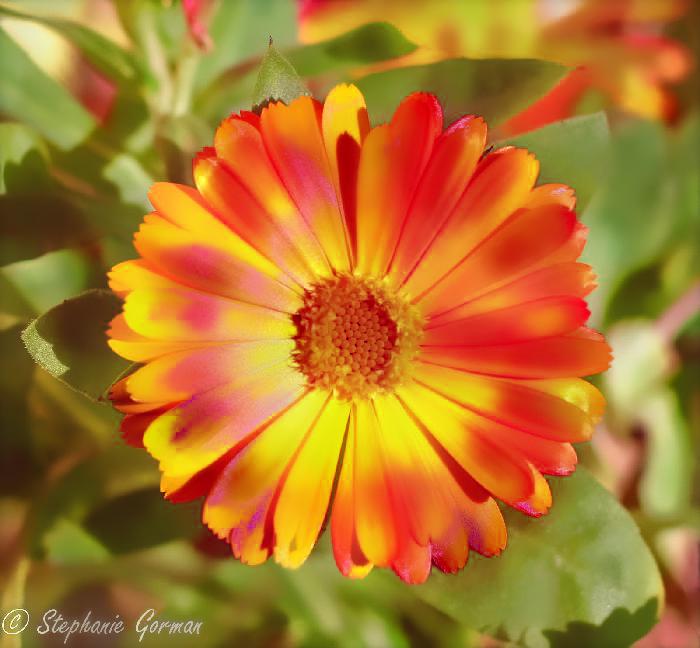} & 
\\

\includegraphics[height = .2\linewidth, width = .16\linewidth]{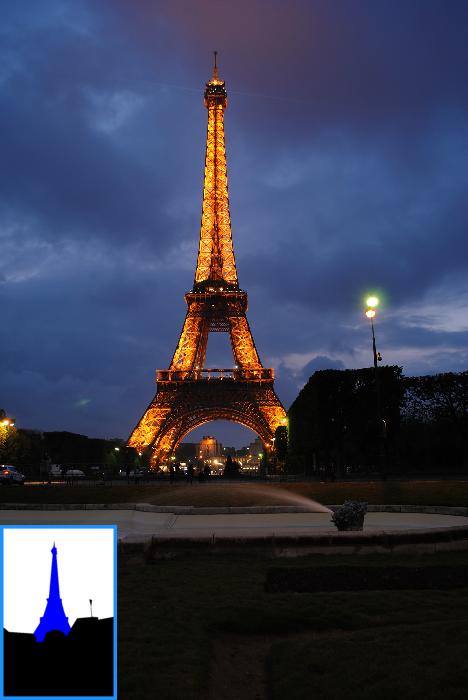} & 

\hspace{1pt}\vrule\hspace{1pt}

\includegraphics[height = .2\linewidth, width = .16\linewidth]{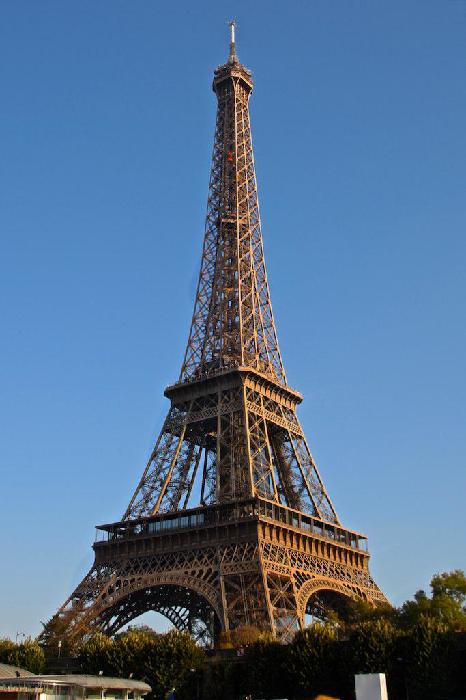} &
\includegraphics[height = .2\linewidth, width = .16\linewidth]{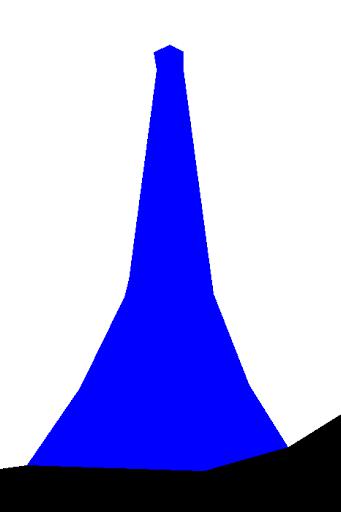} & 
\includegraphics[height = .2\linewidth, width = .16\linewidth]{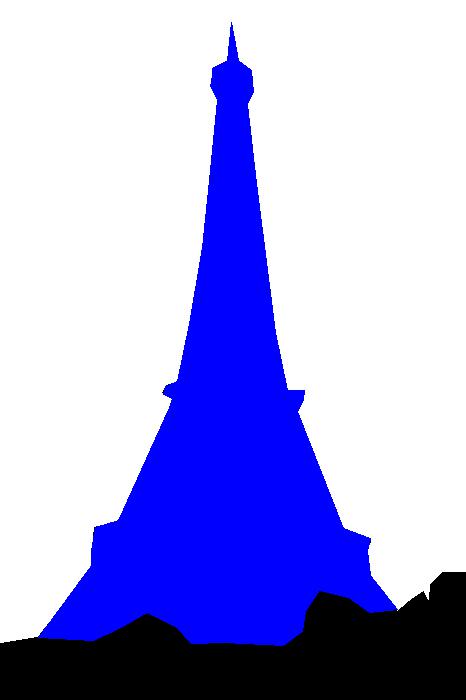} & 
\includegraphics[height = .2\linewidth, width = .16\linewidth]{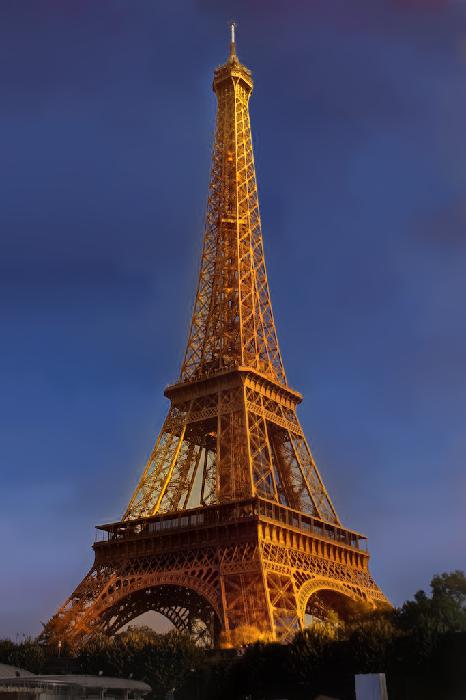} & 
\includegraphics[height = .2\linewidth, width = .16\linewidth]{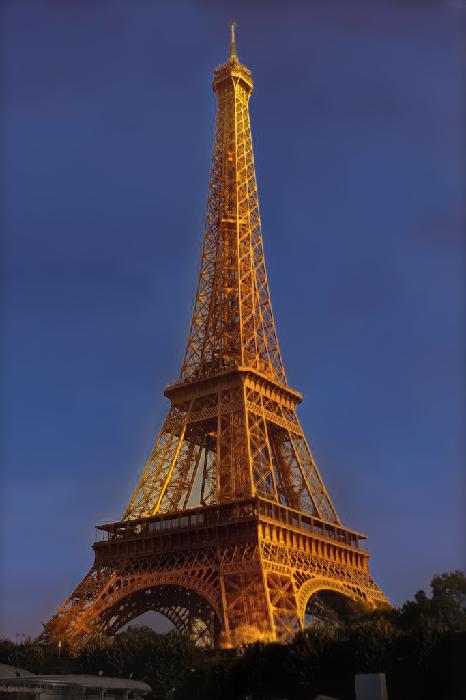} & 
\\

{(a) Style }& {(b) Content } & {(c) Coarse  }& {(d) Precise } & {(e) Stylization } & {(f) Stylization }\\
{}& {} & {map}& {map} & {w/ coarse map} & {w/ precise map}\\
%{}& {} & {}& {} & {map} & {map}\\

\end{tabular}

\caption{Comparisons of stylization results between drawing the coarse and precise label maps in the content.
}
\label{fig:mask}
\end{figure}

The proposed algorithm can leverage semantic label maps for better content--style matching when they are available, similar to the prior work~\cite{Gatys2016-control,Luan-2017-photorealism}. We only use the label map for finding matching areas between content and style images. The specific class information is not used. We further note that the proposed algorithm does not require the label map to be drawn precisely along object boundaries. The photorealistic smoothing step, which employs pixel affinities to encourage consistent stylization, could accommodate imprecise boundary annotations. This could greatly reduce the labeling burden for users. Figure~\ref{fig:mask} shows the comparisons between using coarse and precise label maps. The results in (e) and (f) show that using the coarse map can achieve nearly the same stylization performance as using the precise map.

\section{Additional Results}

\begin{figure}[!htbp]
\centering

\begin{tabular}{c@{\hspace{0.005\linewidth}}c@{\hspace{0.005\linewidth}}c@{\hspace{0.005\linewidth}}c@{\hspace{0.005\linewidth}}c@{\hspace{0.005\linewidth}}c}

\includegraphics[width = .32\linewidth]{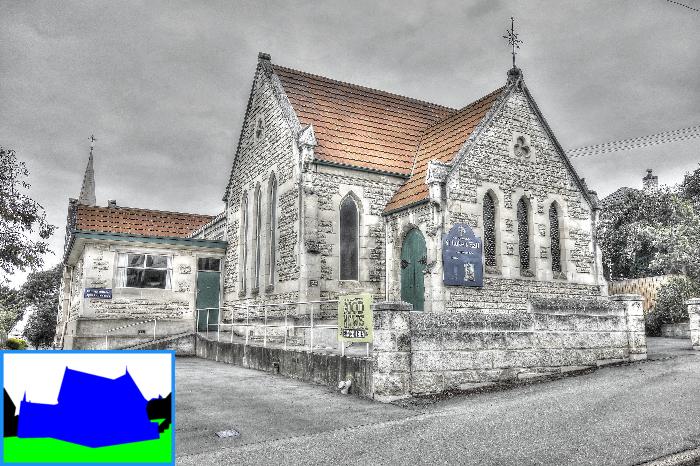} & 
\includegraphics[width = .32\linewidth]{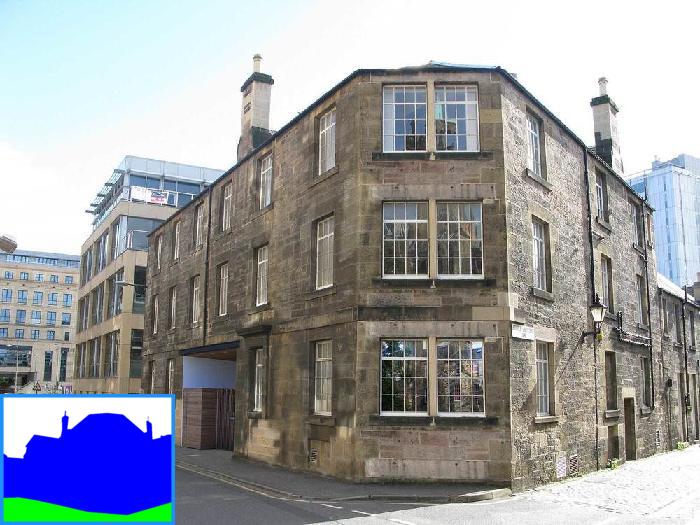} & 
\includegraphics[width = .32\linewidth]{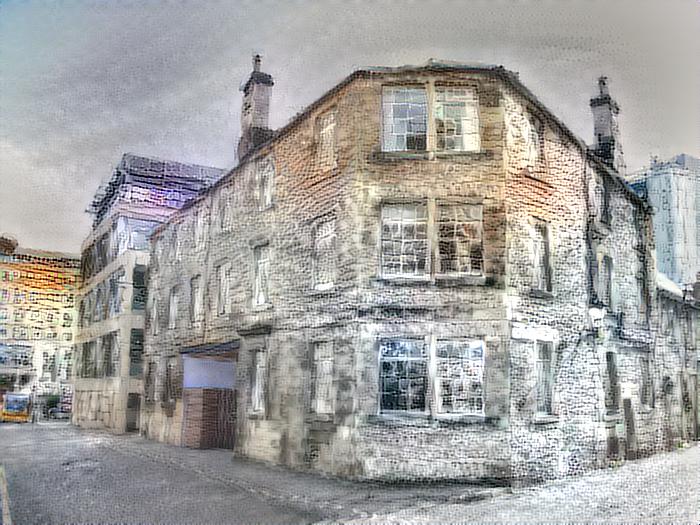} & \\

{(a) Style }& {(b) Content } & {(c) Gatys \etal~\cite{GatysTransfer-CVPR2016} }& \\

\includegraphics[width = .32\linewidth]{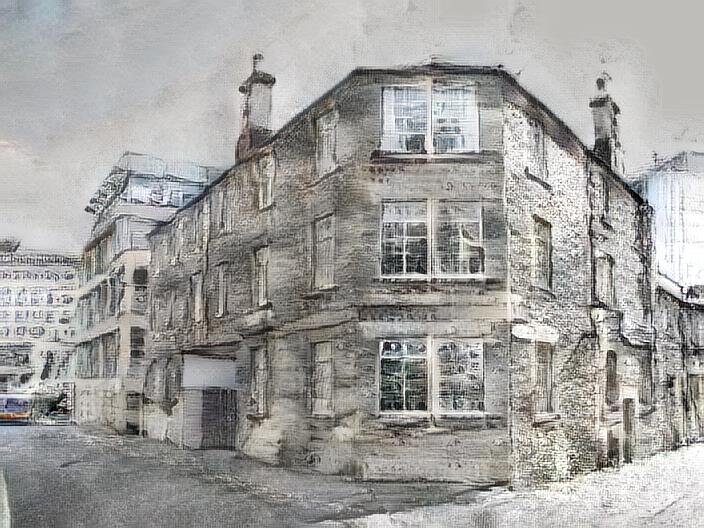} & 
\includegraphics[width = .32\linewidth]{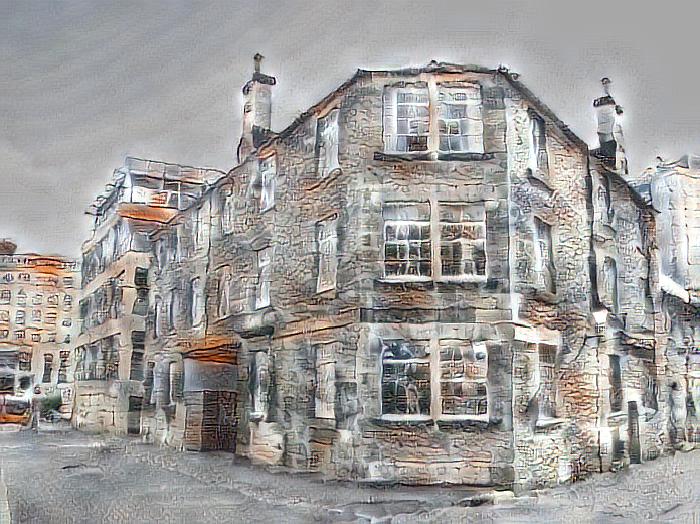} & 
\includegraphics[width = .32\linewidth]{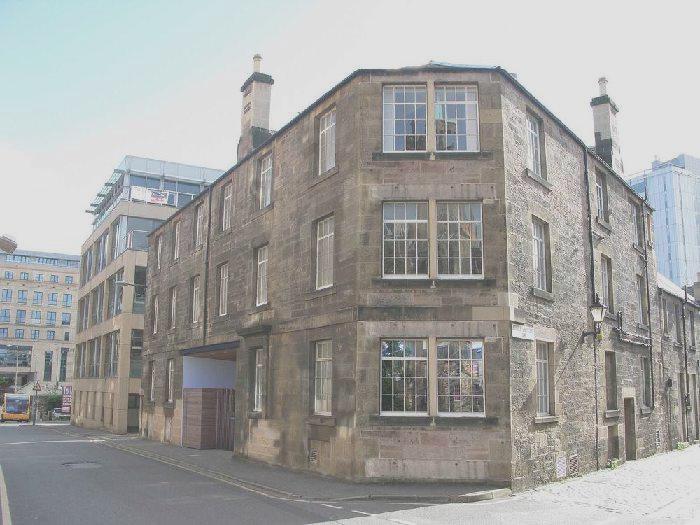} & \\

{(d) Huang \etal~\cite{Huang-2017-arbitrary}}& {(e) WCT~\cite{WCT-2017-NIPS} } & {(f) Reinhard \etal~\cite{reinhard-2001color} } &\\

\includegraphics[width = .32\linewidth]{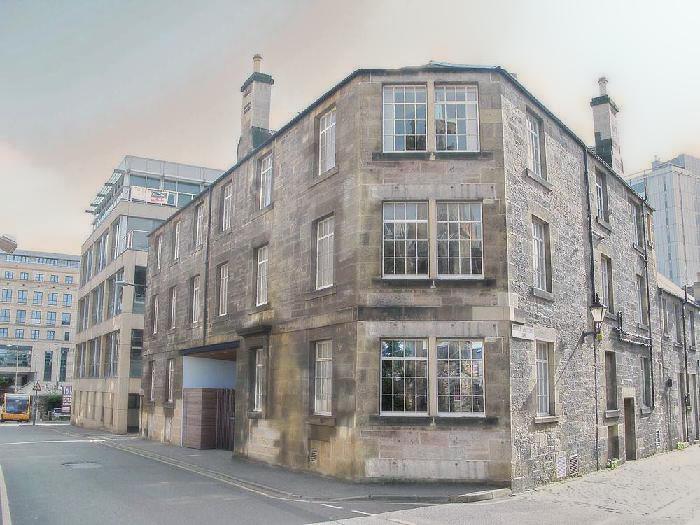} & 
\includegraphics[width = .32\linewidth]{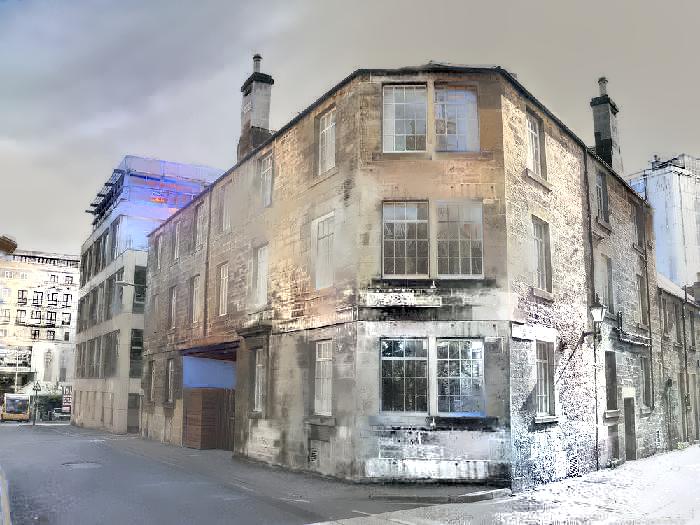} & 
\includegraphics[width = .32\linewidth]{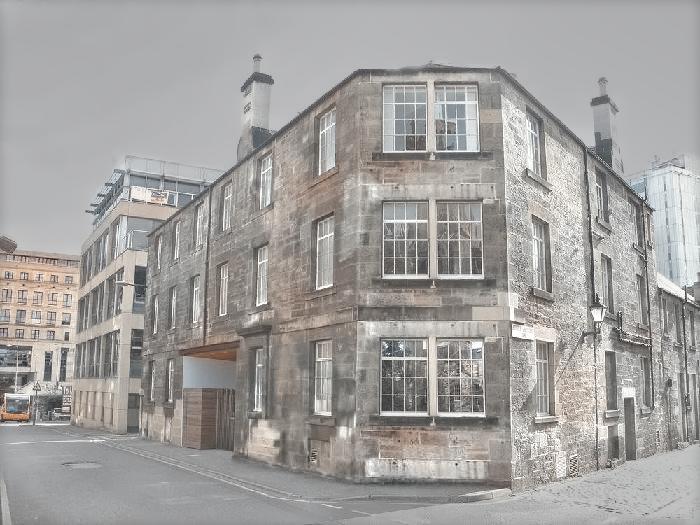} & \\

{(g) Piti\'e \etal~\cite{Pitie-2005} }& {(h) Luan \etal~\cite{Luan-2017-photorealism} } & {(i) Ours } &\\
\end{tabular}
\caption{Comparisons of different stylization methods. 
}
\label{fig:in63}
\end{figure}

\begin{figure}[!htbp]
\centering

\begin{tabular}{c@{\hspace{0.005\linewidth}}c@{\hspace{0.005\linewidth}}c@{\hspace{0.005\linewidth}}c@{\hspace{0.005\linewidth}}c@{\hspace{0.005\linewidth}}c}

\includegraphics[width = .32\linewidth]{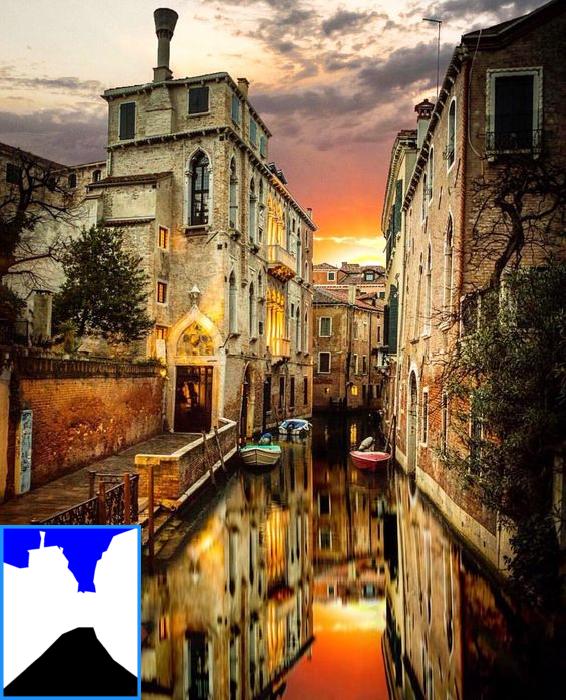} & 
\includegraphics[height= .36\linewidth, width = .32\linewidth]{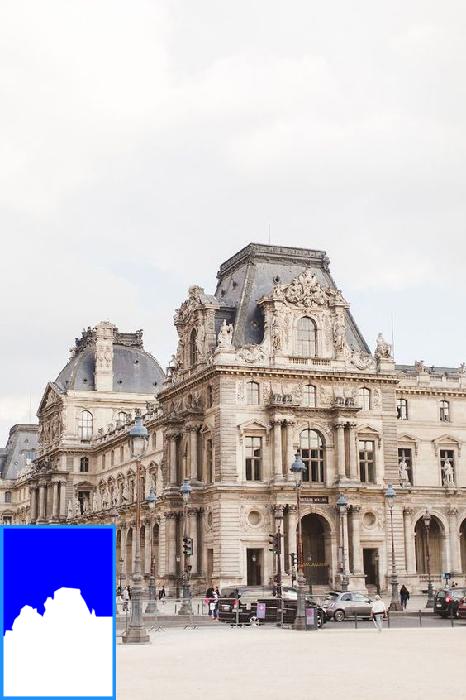} & 
\includegraphics[height= .36\linewidth, width = .32\linewidth]{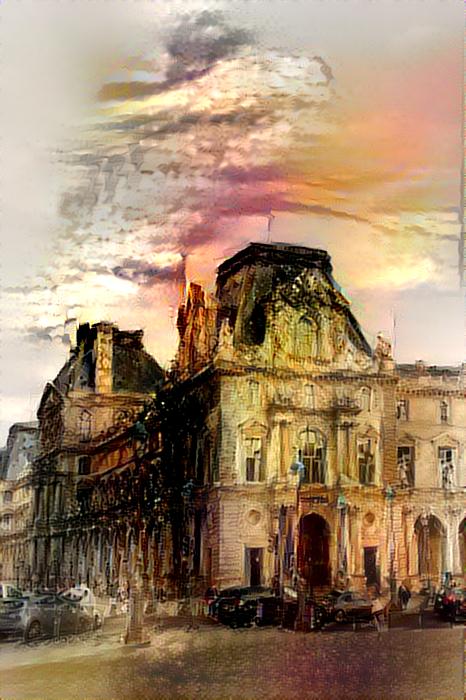} & \\

{(a) Style }& {(b) Content } & {(c) Gatys \etal~\cite{GatysTransfer-CVPR2016} }& \\

\includegraphics[height= .36\linewidth, width = .32\linewidth]{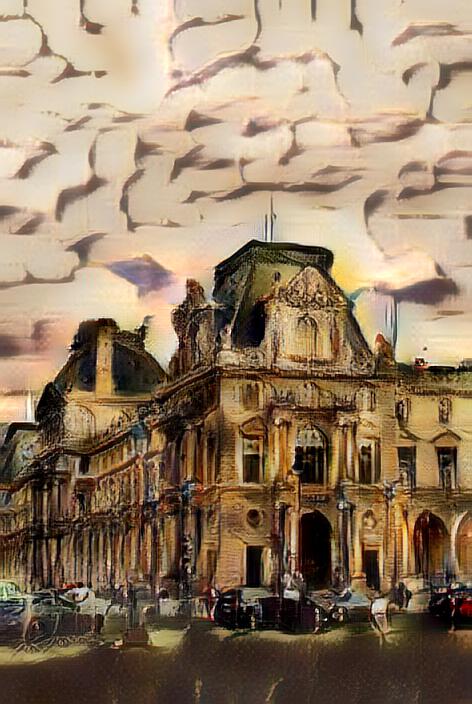} & 
\includegraphics[height= .36\linewidth, width = .32\linewidth]{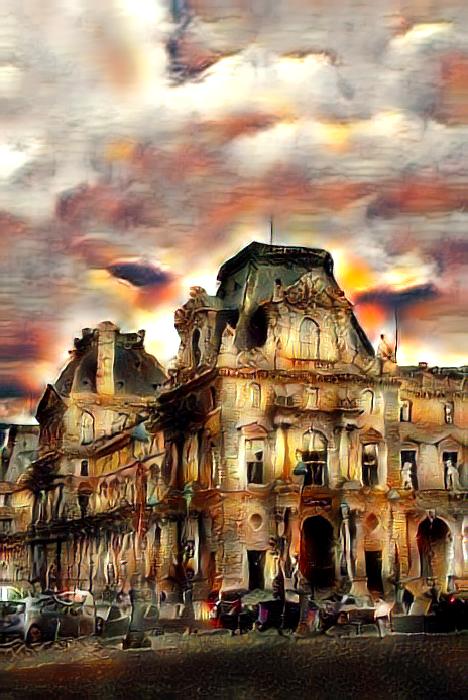} & 
\includegraphics[height= .36\linewidth, width = .32\linewidth]{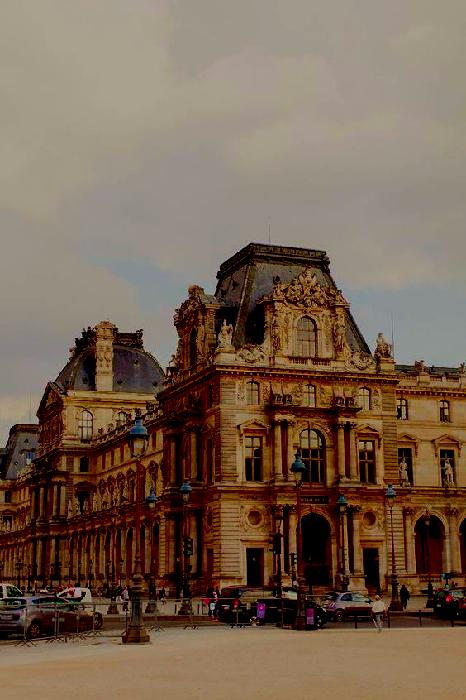} & \\

{(d) Huang \etal~\cite{Huang-2017-arbitrary}}& {(e) WCT~\cite{WCT-2017-NIPS} } & {(f) Reinhard \etal~\cite{reinhard-2001color} } &\\

\includegraphics[height= .36\linewidth, width = .32\linewidth]{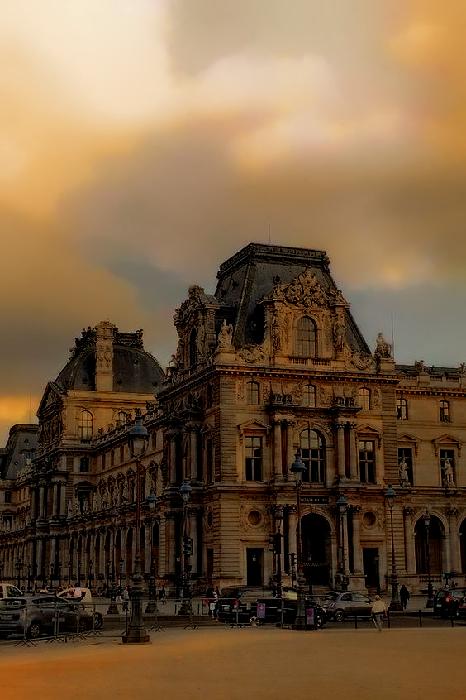} & 
\includegraphics[height= .36\linewidth, width = .32\linewidth]{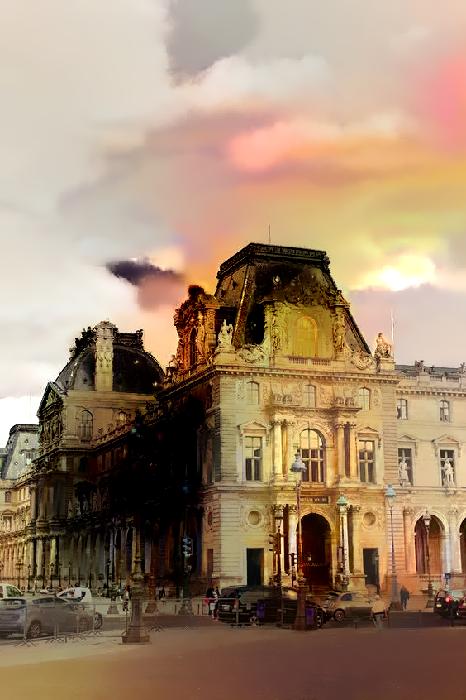} & 
\includegraphics[height= .36\linewidth, width = .32\linewidth]{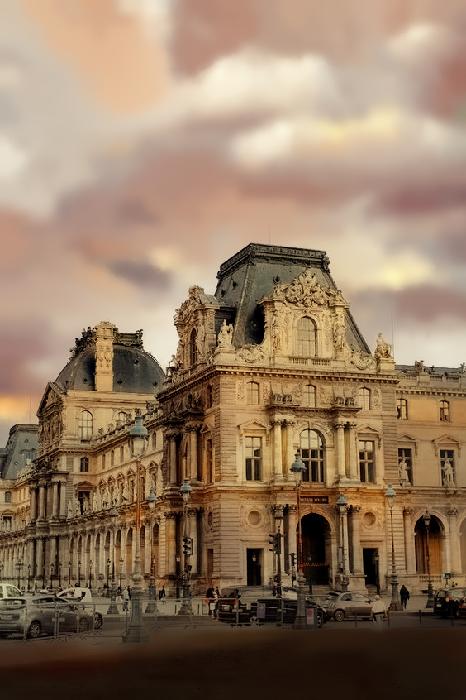} & \\

{(g) Piti\'e \etal~\cite{Pitie-2005} }& {(h) Luan \etal~\cite{Luan-2017-photorealism} } & {(i) Ours } &\\
\end{tabular}
\caption{Comparisons of different stylization methods. 
}
\label{fig:in84}
\end{figure}

\begin{figure}[!htbp]
\centering

\begin{tabular}{c@{\hspace{0.005\linewidth}}c@{\hspace{0.005\linewidth}}c@{\hspace{0.005\linewidth}}c@{\hspace{0.005\linewidth}}c@{\hspace{0.005\linewidth}}c}

\includegraphics[width = .32\linewidth]{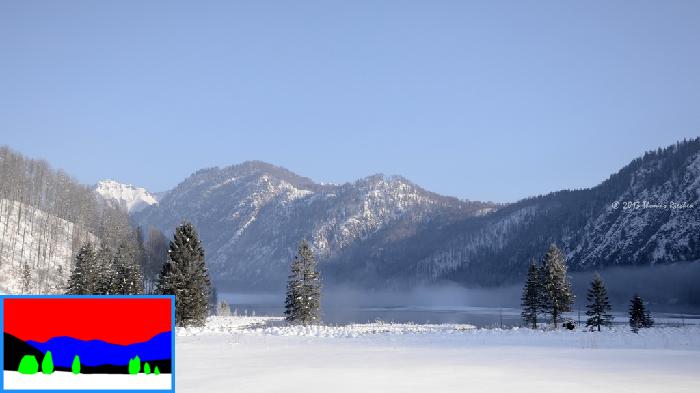} & 
\includegraphics[width = .32\linewidth]{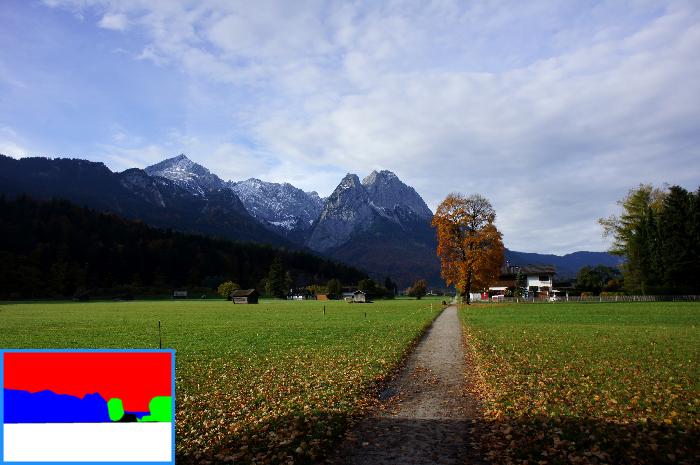} & 
\includegraphics[width = .32\linewidth]{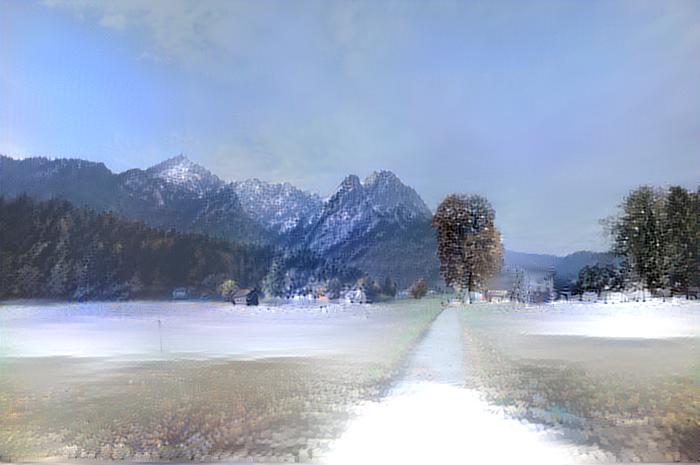} & \\

{(a) Style }& {(b) Content } & {(c) Gatys \etal~\cite{GatysTransfer-CVPR2016} }& \\

\includegraphics[width = .32\linewidth]{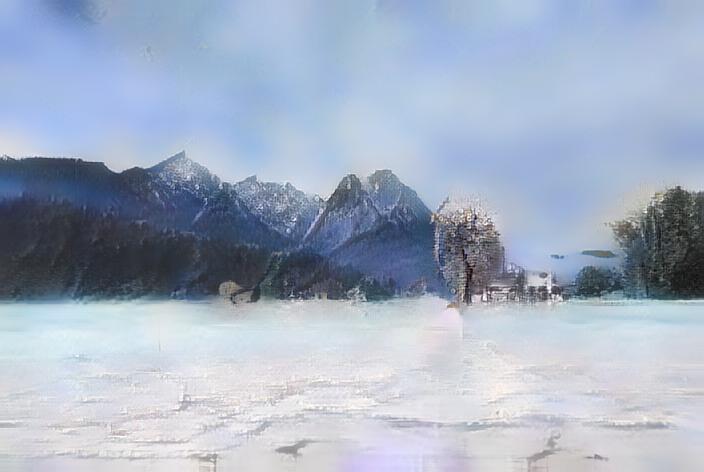} & 
\includegraphics[width = .32\linewidth]{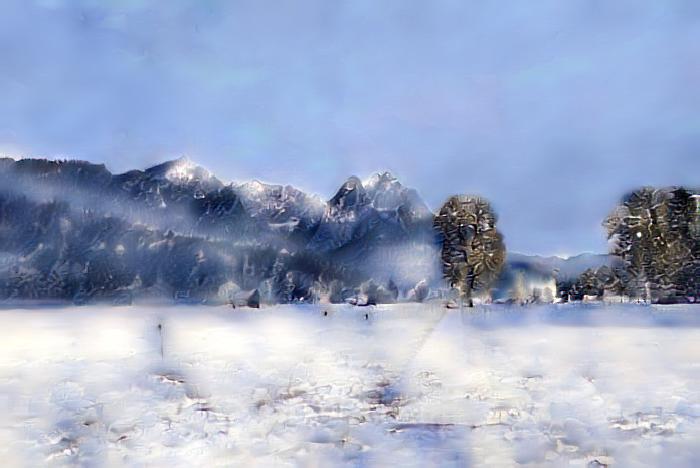} & 
\includegraphics[width = .32\linewidth]{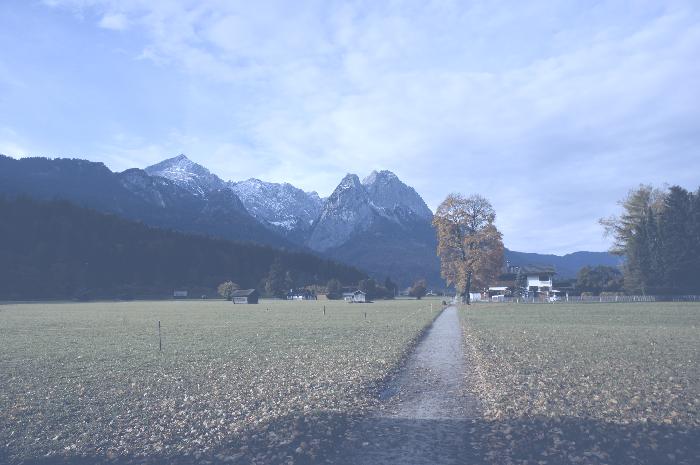} & \\

{(d) Huang \etal~\cite{Huang-2017-arbitrary}}& {(e) WCT~\cite{WCT-2017-NIPS} } & {(f) Reinhard \etal~\cite{reinhard-2001color} } &\\

\includegraphics[width = .32\linewidth]{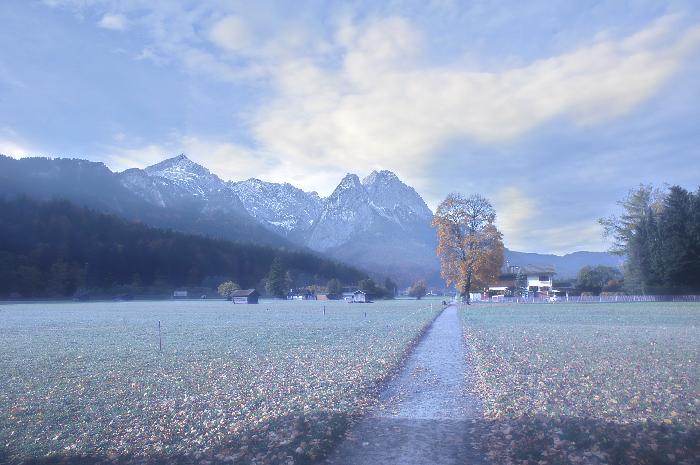} & 
\includegraphics[width = .32\linewidth]{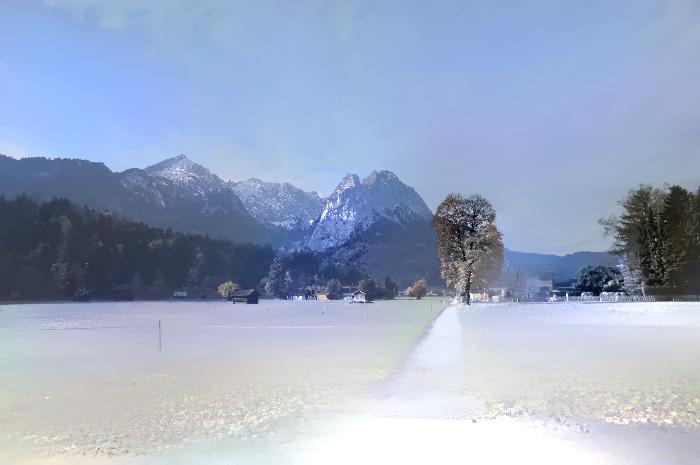} & 
\includegraphics[width = .32\linewidth]{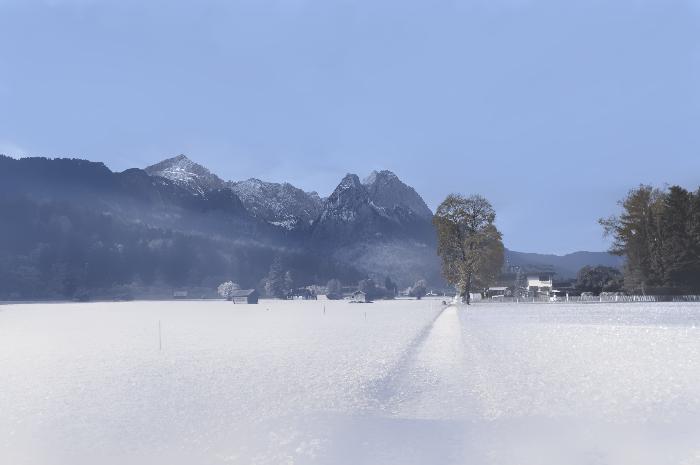} & \\

{(g) Piti\'e \etal~\cite{Pitie-2005} }& {(h) Luan \etal~\cite{Luan-2017-photorealism} } & {(i) Ours } &\\
\end{tabular}
\caption{Comparisons of different stylization methods. 
}
\label{fig:in81}
\end{figure}

\begin{figure}[!htbp]
\centering

\begin{tabular}{c@{\hspace{0.005\linewidth}}c@{\hspace{0.005\linewidth}}c@{\hspace{0.005\linewidth}}c@{\hspace{0.005\linewidth}}c@{\hspace{0.005\linewidth}}c}

\includegraphics[width = .32\linewidth]{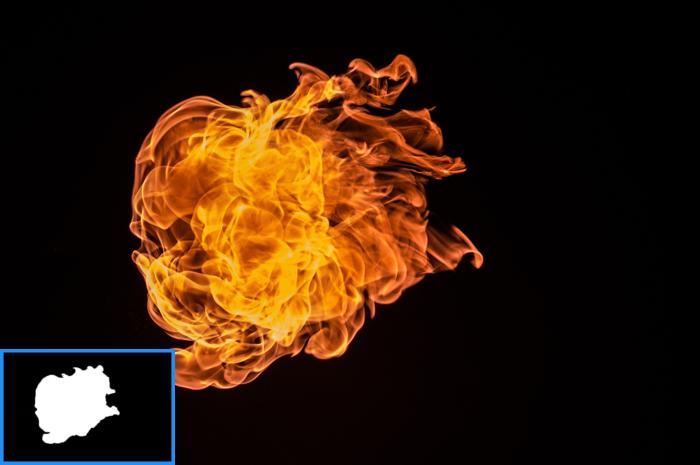} & 
\includegraphics[width = .32\linewidth]{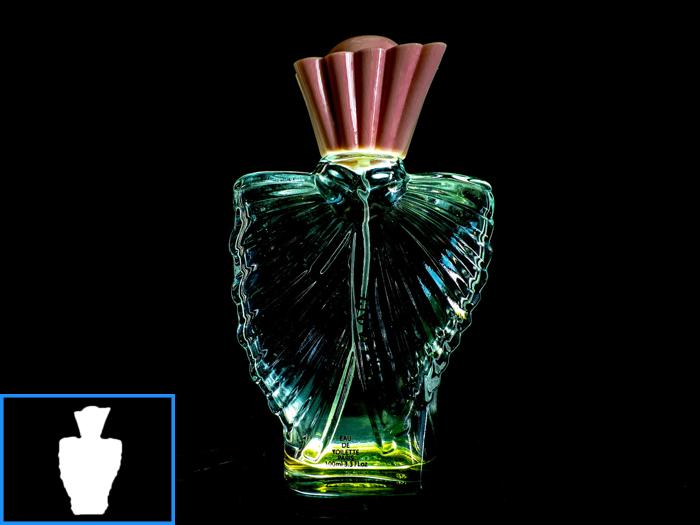} & 
\includegraphics[width = .32\linewidth]{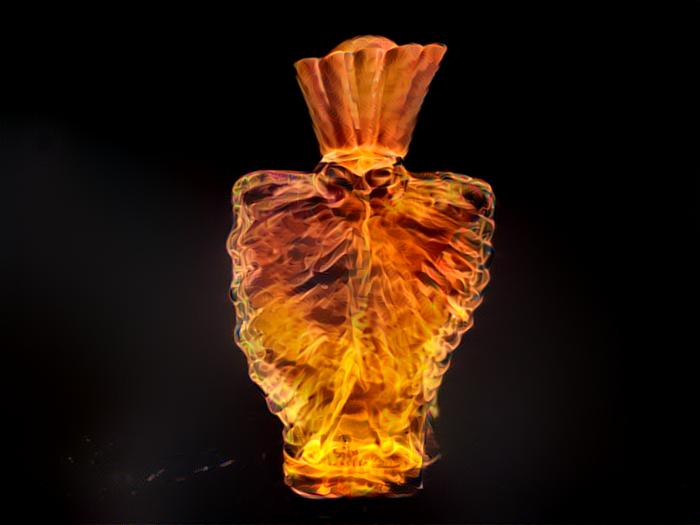} & \\

{(a) Style }& {(b) Content } & {(c) Gatys \etal~\cite{GatysTransfer-CVPR2016} }& \\

\includegraphics[width = .32\linewidth]{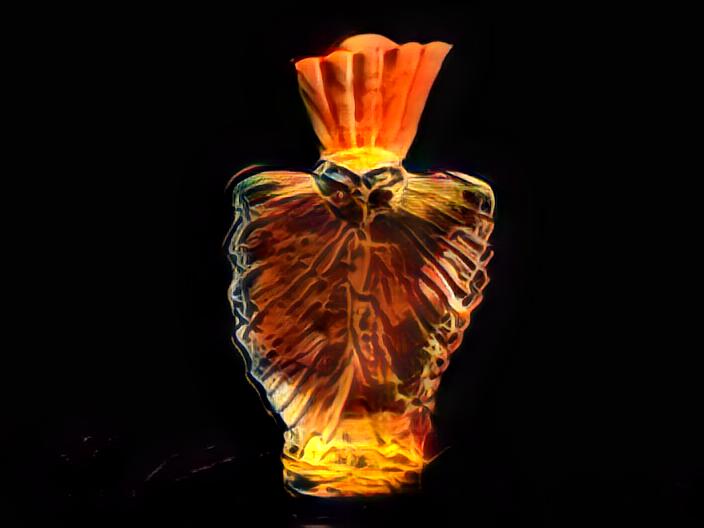} & 
\includegraphics[width = .32\linewidth]{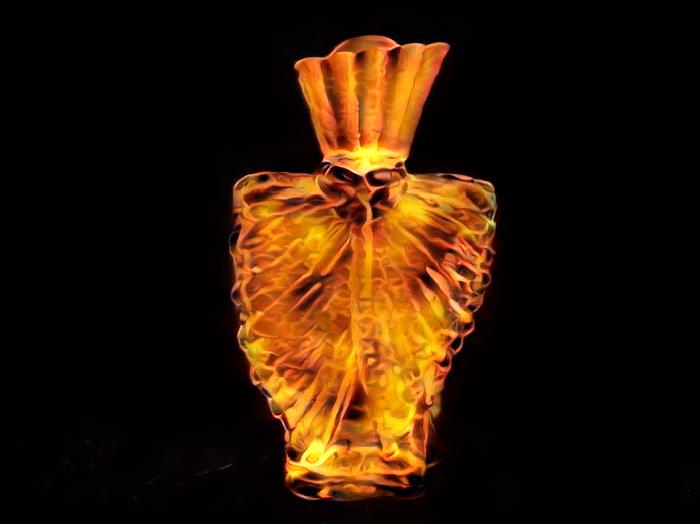} & 
\includegraphics[width = .32\linewidth]{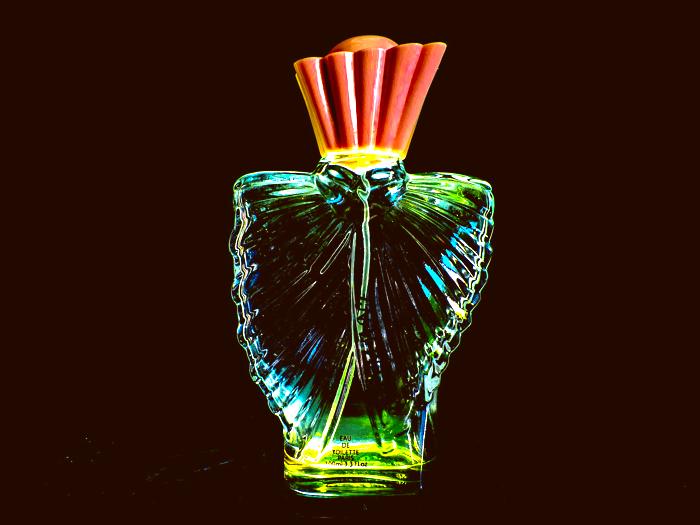} & \\

{(d) Huang \etal~\cite{Huang-2017-arbitrary}}& {(e) WCT~\cite{WCT-2017-NIPS} } & {(f) Reinhard \etal~\cite{reinhard-2001color} } &\\

\includegraphics[width = .32\linewidth]{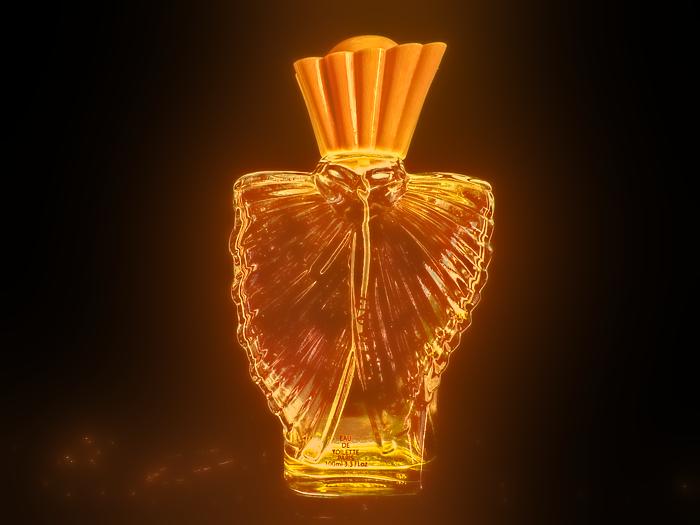} & 
\includegraphics[width = .32\linewidth]{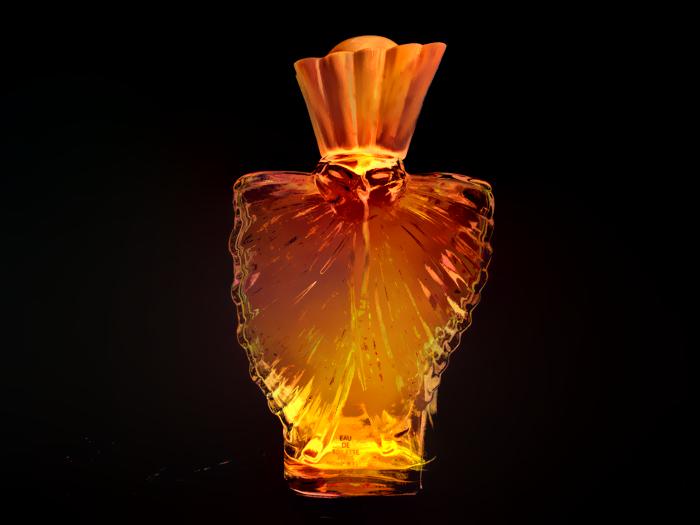} & 
\includegraphics[width = .32\linewidth]{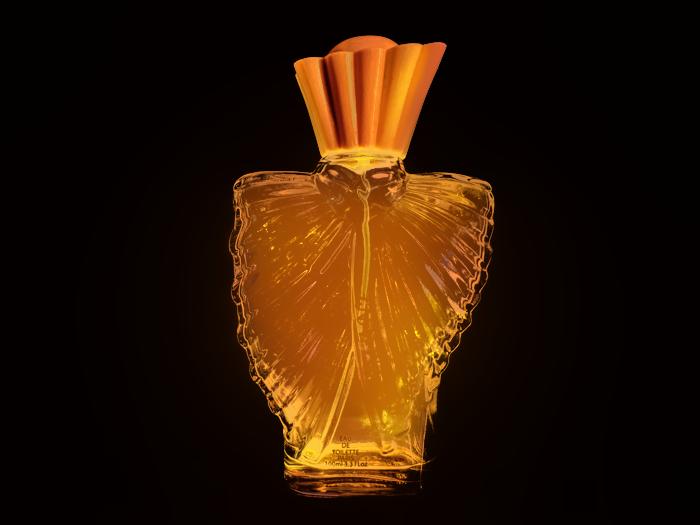} & \\

{(g) Piti\'e \etal~\cite{Pitie-2005} }& {(h) Luan \etal~\cite{Luan-2017-photorealism} } & {(i) Ours } &\\
\end{tabular}
\caption{Comparisons of different stylization methods. 
}
\label{fig:in16}
\end{figure}

\begin{figure}[!htbp]
\centering

\begin{tabular}{c@{\hspace{0.005\linewidth}}c@{\hspace{0.005\linewidth}}c@{\hspace{0.005\linewidth}}c@{\hspace{0.005\linewidth}}c@{\hspace{0.005\linewidth}}c}

\includegraphics[width = .32\linewidth]{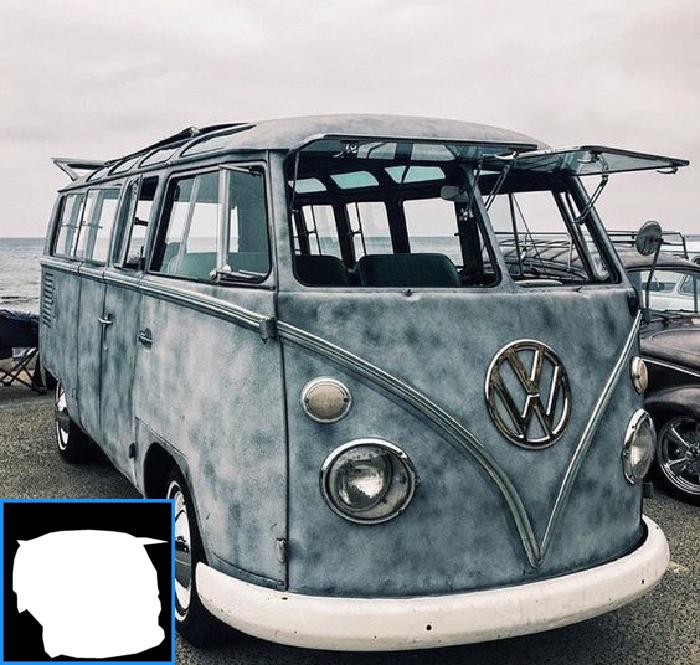} & 
\includegraphics[width = .32\linewidth]{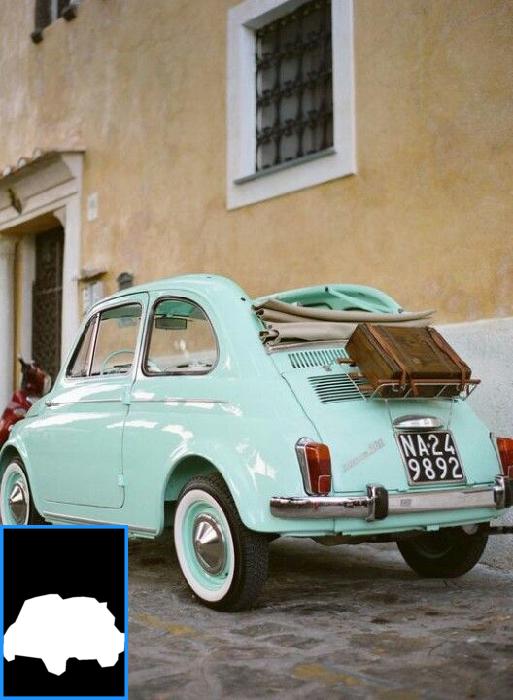} & 
\includegraphics[width = .32\linewidth]{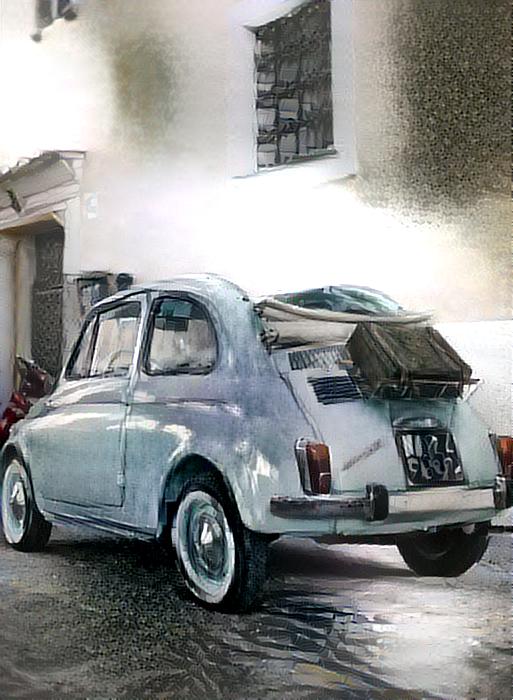} & \\

{(a) Style }& {(b) Content } & {(c) Gatys \etal~\cite{GatysTransfer-CVPR2016} }& \\

\includegraphics[width = .32\linewidth]{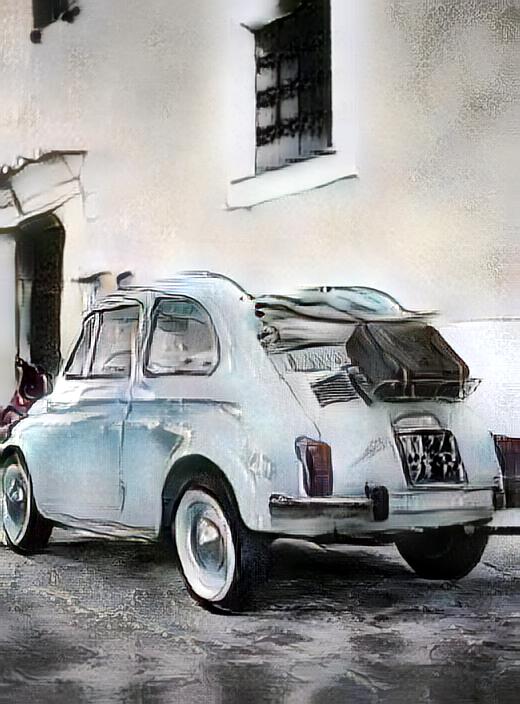} & 
\includegraphics[width = .32\linewidth]{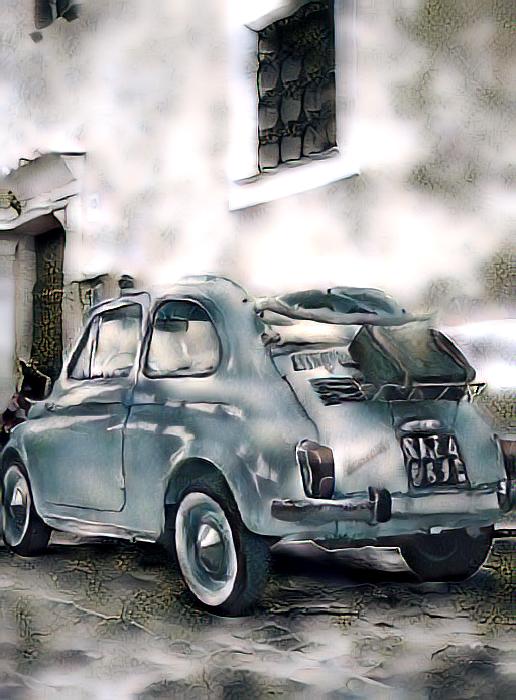} & 
\includegraphics[width = .32\linewidth]{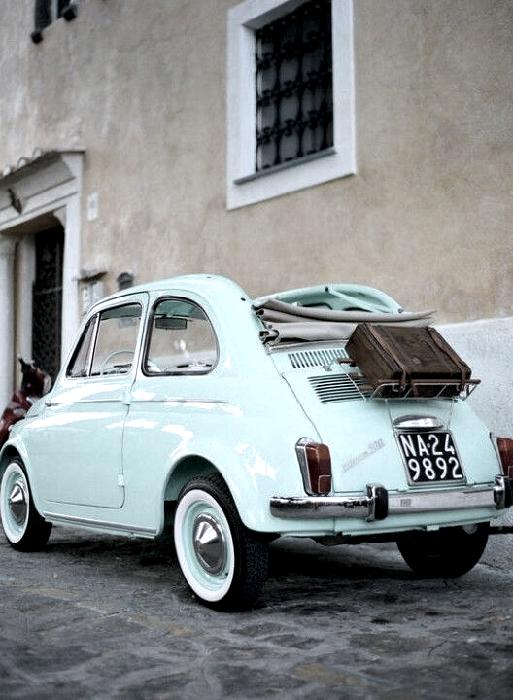} & \\

{(d) Huang \etal~\cite{Huang-2017-arbitrary}}& {(e) WCT~\cite{WCT-2017-NIPS} } & {(f) Reinhard \etal~\cite{reinhard-2001color} } &\\

\includegraphics[width = .32\linewidth]{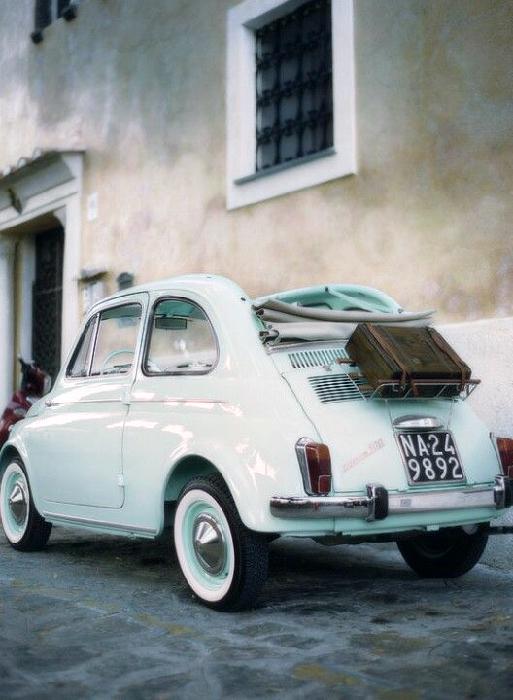} & 
\includegraphics[width = .32\linewidth]{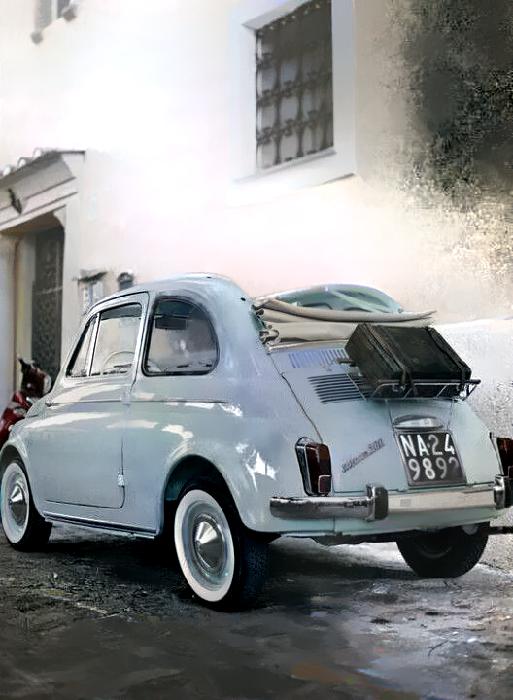} & 
\includegraphics[width = .32\linewidth]{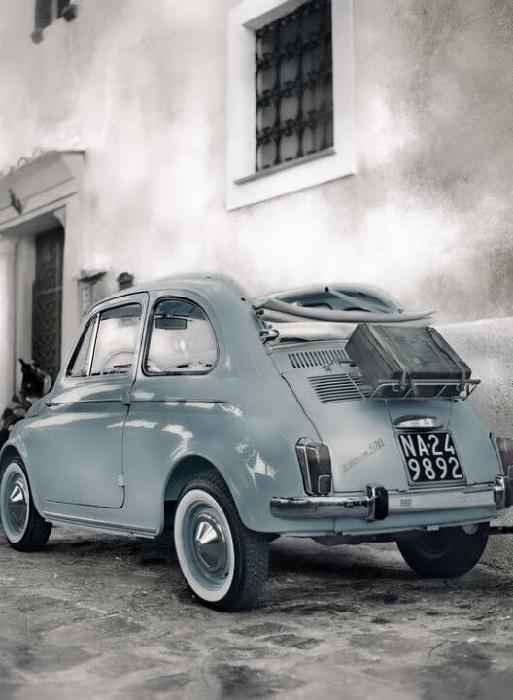} & \\

{(g) Piti\'e \etal~\cite{Pitie-2005} }& {(h) Luan \etal~\cite{Luan-2017-photorealism} } & {(i) Ours } &\\
\end{tabular}
\caption{Comparisons of different stylization methods. 
}
\label{fig:in87}
\end{figure}

We show more photorealistic stylization results in Figure~\ref{fig:in63} to Figure~\ref{fig:in87}. In each figure, we first present the content--style pair together with their corresponding label maps in (a) and (b). The maps are either from Luan \etal~\cite{Luan-2017-photorealism} or roughly drawn by human. Each color represents a different semantic label. 

We compare our method with three artistic stylization methods~\cite{GatysTransfer-CVPR2016,Huang-2017-arbitrary,WCT-2017-NIPS} in (c)--(e) and three photorealistic stylization methods~\cite{reinhard-2001color,Pitie-2005,Luan-2017-photorealism} in (f)--(h). The results show that our method generates more photorealistic results with much less structural artifacts and more consistent stylizations for a variety of examples.

\end{document}